\documentclass[lettersize,journal]{IEEEtran}
\ifCLASSINFOpdf
\else
\fi
\usepackage[stretch=10]{microtype}
\usepackage[utf8]{inputenc}
\usepackage[T1]{fontenc}
\PassOptionsToPackage{hyphens}{url}
\usepackage{times,multirow,float}
\usepackage{graphicx}
\usepackage{epsfig,xspace,layout}
\usepackage{color}
\usepackage{amsfonts}
\usepackage{times}
\usepackage{amssymb}
\usepackage{amsthm}
\usepackage{hyperref}
\usepackage{amsmath,bm}
\usepackage{rotating}
\usepackage{mathrsfs}
\usepackage{mathtools}
\usepackage{makeidx}
\usepackage[]{threeparttable}
\usepackage{dsfont}
\usepackage{cite}
\usepackage{xcolor}
\usepackage{cleveref}
\usepackage[ruled, lined, linesnumbered, commentsnumbered, longend]{algorithm2e}
\usepackage{tikz}
\usetikzlibrary{shapes,arrows,positioning,calc}
\usepackage{siunitx}
\usepackage{multicol,lipsum}
\usepackage{subcaption}
\usepackage[font=footnotesize]{caption}
\usepackage{booktabs}
\usepackage{lineno}
\usepackage{hhline}
\usepackage[Symbol]{upgreek}
\usepackage{float}
\newtheorem{lemma}{Lemma}
\newtheorem{remark}{Remark}
\SetKwComment{Comment}{$\triangleright$\ }{}
\SetKwInput{Return}{Return}
\makeatletter
\renewcommand{\Indentp}[1]{%
  \advance\leftskip by #1
  \advance\skiptext by -#1
  \advance\skiprule by #1}%
\renewcommand{\Indp}{\algocf@adjustskipindent\Indentp{\algoskipindent}}
\renewcommand{\Indm}{\algocf@adjustskipindent\Indentp{-\algoskipindent}}
\makeatother

\begin{document}
\captionsetup[table]{name=TABLE,labelsep=newline,textfont=sc}
%
\title{Neural Moving Horizon Estimation\\ for Robust Flight Control}
%
%

\author{Bingheng Wang, Zhengtian Ma, Shupeng Lai, and Lin Zhao

\thanks{
\setlength{\fboxrule}{0.4pt}%
  \setlength{\fboxsep}{6pt}%
  \fbox{\parbox{0.92\columnwidth}{%
    This article has been accepted for publication in IEEE
    Transactions on Robotics. \copyright~2023 IEEE. Personal use
    of this material is permitted. Permission from IEEE must be
    obtained for all other uses, in any current or future media,
    including reprinting/republishing this material for advertising
    or promotional purposes, creating new collective works, for
    resale or redistribution to servers or lists, or reuse of any
    copyrighted component of this work in other works. DOI:
    10.1109/TRO.2023.3331064}}

The authors are with the Department of Electrical and Computer Engineering,
        National University of Singapore, 117583 Singapore
        {\tt\small wangbingheng@u.nus.edu},
        {\tt\small shupenglai@gmail.com}
        {\tt\small $\left\{ {} \right.$zhengtian, elezhli$\left.\right\}$@nus.edu.sg}.}
}

%
%
\newcommand{\Lin}[1]{\textcolor{blue}{[#1]}}

\markboth{Accepted version (not the final version) for publication in IEEE Transactions on Robotics}%
{Shell \MakeLowercase{\textit{et al.}}: Bare Demo of IEEEtran.cls for IEEE Journals}
%



\maketitle

\begin{abstract}
Estimating and reacting to disturbances is crucial for robust flight control of quadrotors. Existing estimators typically require significant tuning for a specific flight scenario or training with extensive ground-truth disturbance data to achieve satisfactory performance. In this paper, we propose a neural moving horizon estimator (NeuroMHE) that can automatically tune the key parameters modeled by a neural network and adapt to different flight scenarios. We achieve this by deriving the analytical gradients of the MHE estimates with respect to the weighting matrices, which enables a seamless embedding of the MHE as a learnable layer into neural networks for highly effective learning. Interestingly, we show that the gradients can be computed efficiently using a Kalman filter in a recursive form. Moreover, we develop a model-based policy gradient algorithm to train NeuroMHE directly from the quadrotor trajectory tracking error without needing the ground-truth disturbance data. The effectiveness of NeuroMHE is verified extensively via both simulations and physical experiments on quadrotors in various challenging flights. Notably, NeuroMHE outperforms a state-of-the-art neural network-based estimator, reducing force estimation errors by up to {$76.7\%$}, while using a portable neural network that has only $7.7\%$ of the learnable parameters of the latter. The proposed method is general and can be applied to robust adaptive control of other robotic systems.
\end{abstract}


%
\IEEEpeerreviewmaketitle

\section*{Supplementary Material}
The videos and source code of this work are available on \url{https://github.com/RCL-NUS/NeuroMHE}.

\section{Introduction}
\label{sec:intro}
%
%
%
%

\IEEEPARstart{Q}{uadrotors} have been increasingly engaged in various challenging tasks, such as aerial swarming~\cite{honig2018trajectory}, racing~\cite{song2021autonomous}, aerial manipulation~\cite{brunner2022energy}, and cooperative transport~\cite{geng2020cooperative, jackson2020scalable}. They can suffer from strong disturbances caused by capricious wind conditions, unmodeled aerodynamics in extreme flights or tight formations~\cite{michael2010grasp, powers2013influence, naka2020coanda,bauersfeld2021neurobem,shi2020neural2}, force interaction with environments, time-varying cable tensions from suspended payloads~\cite{belkhale2021model}, etc. These disturbances must be compensated appropriately in control systems to avoid significant deterioration of flight performance and even crashing. However, it is generally intractable to have a portable model capable of capturing various disturbances over a wide range of complex flight scenarios. Therefore, online disturbance estimation that adapts to environments is imperative for robust flight control.
\begin{figure}[t!]
	\centering
	{\includegraphics[width=0.9\columnwidth]{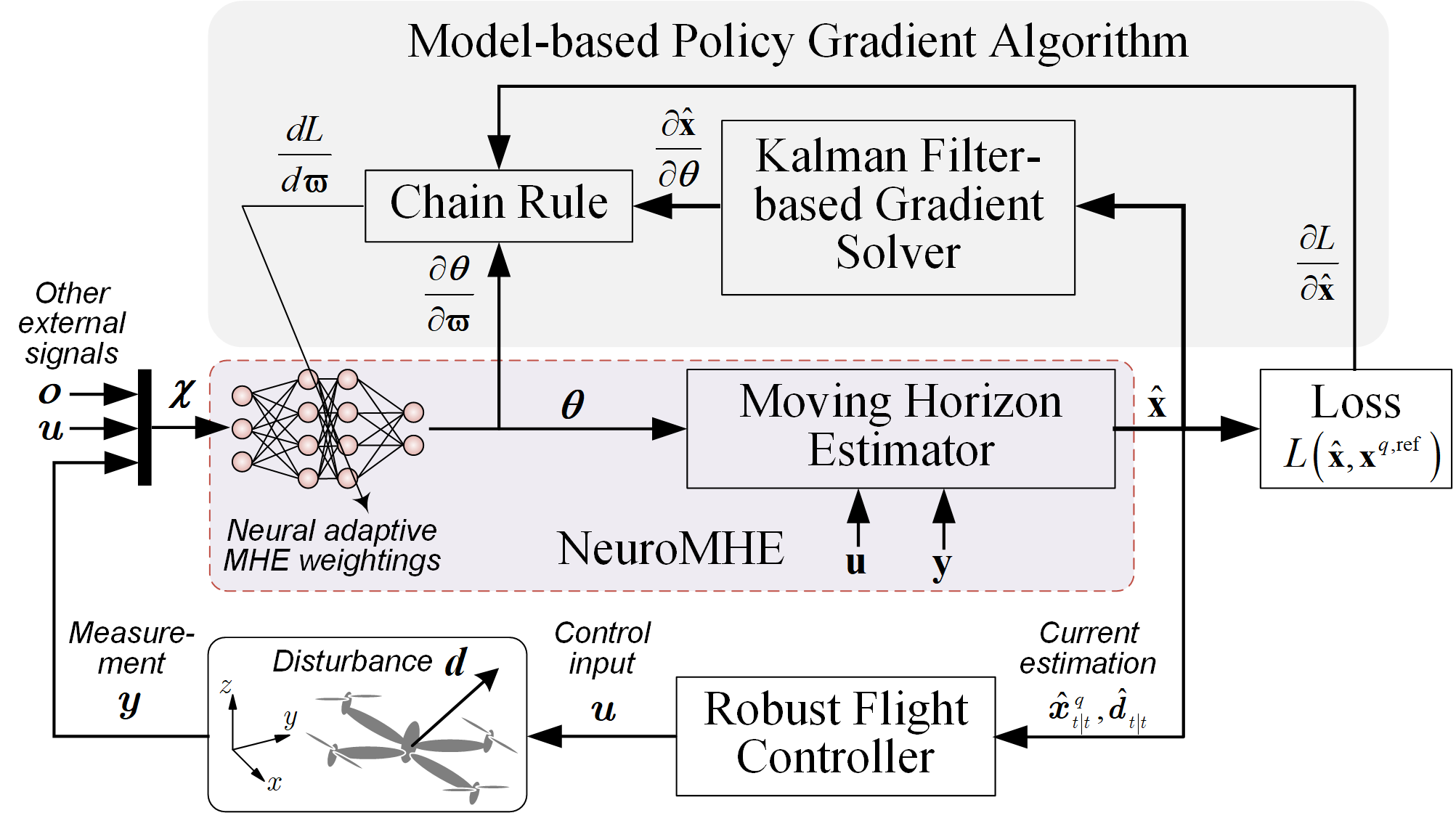}}
	\caption{\footnotesize A diagram of the NeuroMHE-based robust flight control system and its learning pipelines. We integrate a portable neural network with an MHE to obtain adaptive weightings $\bm \theta$ online. By defining the disturbance $\bm d$ as an augmented state, the NeuroMHE can generate accurate state estimates $\hat {\bm{\mathop{ {\rm x}}\nolimits}}$ based on a history of control inputs $\bm{\mathop{ {\rm u}}\nolimits}$ and a histroy of measurements $\bm{\mathop{ {\rm y}}\nolimits}$ (See \ref{section:preliminary}). The neural network parameters $\bm \varpi $ are efficiently learned from the trajectory tracking error. Central to our algorithm are the gradients $\frac{\partial \hat{\mathbf x}}{\partial \boldsymbol\theta}$, which are calculated recursively using a Kalman filter.}
\label{fig:neuromhe learning pipeline}	
\end{figure}

Estimating and reacting to disturbances has long been a focus of quadrotor research.
Some early works~\cite{6907146,yuksel2014nonlinear,tomic2014unified,9295362,brunner2022energy} proposed momentum-based estimators using a stable low pass filter.
These filters depend on static disturbance assumption, and thus have limited performance against fast-changing disturbances. McKinnon et al.~\cite{mckinnon2020estimating} modelled the dynamics of the disturbances as random walks and applied an unscented Kalman filter (UKF) to estimate the external disturbances. This method generally works well across different scenarios. However, its performance relies heavily on manually tuning tens of noise covariance
parameters that are hard to identify. In fact, the tuning process can be rather obscure and requires significant experimental efforts together with much expert knowledge of the overall hardware and software systems. Recently, there has been an increasing interest of utilizing deep neural networks (DNNs) for quadrotor disturbance estimation~\cite{punjani2015deep,mohajerin2018deep,belkhale2021model,shi2020neural2}. Shi et al.~\cite{shi2020neural2} trained DNNs to capture the aerodynamic interaction forces between multiple quadrotors in close-proximity flight. Bauersfeld et al. proposed a hybrid estimator NeuroBEM \cite{bauersfeld2021neurobem} to estimate aerodynamics for a single quadrotor at extreme flight. The latter combines the first-principle Blade Element Momentum model for single rotor aerodynamics modeling and a DNN for residual aerodynamics estimation. These DNN estimators generally employ a relatively large neural network to achieve satisfactory performance. Meanwhile, their training demands significant amounts of ground-truth disturbance data and can require complicated learning curricula.

In this paper, we propose a neural moving horizon estimator (NeuroMHE) that can accurately estimate the disturbances and adapt to different flight scenarios. Distinct from the aforementioned methods, NeuroMHE fuses a portable neural network with a control-theoretic MHE, which can be trained efficiently without needing the ground-truth disturbance data. MHE solves a nonlinear dynamic optimization online in a receding horizon manner. It is well known to have superior performance on nonlinear systems with uncertainties~\cite{diehl2009efficient, papadimitriou2022external}. The performance of MHE depends on a set of weighting matrices in the cost function. They are roughly inversely proportional to the covariances of the noises that enter the dynamic systems~\cite{wenz2019moving,osman2021generic,eltrabyly2021quadcopter,hu2022joint}. However, the tuning of these weightings is difficult in practice due to their large number and strong coupling, as well as the nonlinearity of the system dynamics. It becomes even more challenging when the optimal weightings are dynamic and highly nonlinear in reality. For example, this is the case when the noise covariances are functions of states or some external signals\cite{robertson1996moving,aravkin2014smoothing,nguyen2019state}. In NeuroMHE, we employ a portable neural network to generate adaptive weightings and develop a systematic way to tune these weightings online using machine learning techniques. Our approach leverages the advantages of both model-free and model-based methods--the expressive power of neural networks and the control-theoretic optimal estimation. Such a fusion provides NeuroMHE with high estimation accuracy for various external disturbances and fast online environment adaption.

Fig.~\ref{fig:neuromhe learning pipeline} outlines the robust flight control system using NeuroMHE and its learning pipelines. The estimated disturbance forces and torques (denoted by $\hat{\bm d}_{t\left| t \right.}$ in the figure) from the NeuroMHE are compensated in the flight controller. We develop a model-based policy gradient algorithm (the shaded blocks) to train the neural network parameters directly from the quadrotor trajectory tracking error. One of the key ingredients of the algorithm is the computation of the gradients of the MHE estimates with respect to the weightings. They are derived by implicitly differentiating through the Karush-Kuhn-Tucker (KKT) conditions of the corresponding MHE optimization problem. In particular, we derive a Kalman filter to compute these gradients very efficiently in a recursive form. These analytical gradients enable a seamless embedding of the MHE as a learnable layer into the neural network for highly effective learning. They allow for training NeuroMHE with powerful machine learning tools. 

There has been a growing interest in joining the force of control-theoretic policies and machine learning approaches, such as OptNet~\cite{amos2017optnet}, differentiable MPC~\cite{amos2018differentiable}, and Pontryagin differentiable programming~\cite{jin2020pontryagin,jin2021safe}. Our work adds to this collection another general policy for estimation, which is of interdisciplinary interest to both robotics and machine learning communities. Other recent works on optimally tuning MHE using gradient descent include~\cite{9483399,muntwiler2021learning}. There the gradients are computed via solving the inverse of a large KKT matrix, whose size is linear with respect to the MHE horizon. The computational complexity is at least quadratic with respect to the horizon. Besides, the method in~\cite{muntwiler2021learning} requires the system dynamics to be linear. In comparison, our proposed algorithm of computing the gradients explores a recursive form using a Kalman filter, which has a linear computational complexity with respect to the MHE horizon. Moreover, it directly handles general nonlinear dynamic systems, which has more applications in robotics.

We validate the effectiveness of NeuroMHE extensively via both simulations and physical experiments on quadrotors in various challenging flights. Using the real agile and extreme flight dataset collected in the world's largest indoor motion capture room~\cite{bauersfeld2021neurobem}, we show that compared with the state-of-the-art estimator NeuroBEM, our method: 1) requires much less training data; 2) uses a mere $7.7\%$ amount of the neural network parameters; and 3) achieves superior performance with force estimation error reductions of up to {$76.7\%$}. Further utilizing a trajectory tracking simulation with synthetic disturbances, we show that a stable NeuroMHE with a fast dynamic response can be efficiently trained in merely a few episodes from the trajectory tracking error. Beyond the impressive training efficiency, this simulation also demonstrates the robust control performance of using NeuroMHE under Gaussian noises with dynamic covariances. As compared to fixed-weighting MHE and $\mathcal{L}_1$ adaptive control, NeuroMHE reduces the average tracking errors by up to $86.3\%$.
Finally, we conduct experiments to show that NeuroMHE robustifies a baseline flight controller substantially against challenging disturbances, including state-dependent cable forces and the downwash flow.

This paper is based on our previous conference paper~\cite{wang2021differentiable}. The earlier version of this work proposed an auto-tuning MHE algorithm with analytical gradients and demonstrated training of fixed-weighting MHE to achieve quadrotor robust control in simulation. In this work, we additionally: 1) propose a more powerful auto-tuning NeuroMHE, to further improve the estimation accuracy and achieve fast online adaptation; 2) provide a theoretical justification of the computational efficiency of the Kalman filter-based gradient solver; 3) extend the theory and algorithms to consider constraints in the estimation, which helps meet various physical limits; 4) conduct extensive simulations using both real and synthetic data to demonstrate the training efficiency, the fast environment adaptation, and the improved estimation and trajectory tracking accuracy over the state-of-the-art estimator and adaptive controller; and 5) further test the proposed algorithm on a real quadrotor to show that NeuroMHE can significantly robustify a widely-used baseline flight controller against various challenging disturbances. 

The rest of this paper is organized as follows. Section~\ref{section:preliminary} briefly reviews the quadrotor dynamics and the MHE problem for the disturbance estimation. Section~\ref{section: neuromhe} formulates NeuroMHE. In Section~\ref{section: gradient}, we derive the analytical gradients via sensitivity analysis. Section~\ref{section:policy gradient} develops the model-based policy gradient algorithm for training NeuroMHE without the need for ground-truth data. Section~\ref{section: constraints} considers constrained NeuroMHE problems. Simulation and experiment results are reported in Section~\ref{section: simulation}. We discuss the advantages and disadvantages of our approach compared to the state-of-the-art methods in Section~\ref{section: discussion} and conclude this paper in Section~\ref{section: conclusion}.

\section{Preliminaries}\label{section:preliminary}

\subsection{Quadrotor and Disturbance Dynamics}
\label{subsec:quadrotor dynamics}
We model a quadrotor as a 6 degree-of-freedom (DoF) rigid body with mass $m$ and moment of inertia ${\bm J}\in {\mathbb R^{3 \times 3}}$. Let ${\bm p \in {\mathbb R^3} }$ denote the global position of Center-of-Mass (CoM) in world frame $\cal {\bm I} $, ${\bm v}\in {\mathbb R^3}$ the velocity of CoM in $\cal {\bm I} $, ${\bm R} \in SO(3)$ the rotation matrix from body frame ${\cal {\bm B}}$ to $\cal{\bm I}$, and $\bm \omega  \in {\mathbb R^3}$ the angular rate in ${\cal {\bm B}}$. The quadrotor model is given by:
\begin{subequations}
\begin{align}
\dot {\bm p} & = {\bm v}, & \dot {\bm v} & = {m^{ - 1}}\left( -{mg {\bm e}_3  + {\bm R}f {\bm e}_3  + {{\bm d}_f}} \right)
\label{eq:position dynamics},\\
\dot {\bm R} &= {\bm R}{\bm \omega ^ \times }, & \dot {\boldsymbol \omega}  & = {{\bm J}^{ - 1}}\left( { - {\boldsymbol \omega ^ \times } {\bm J}{\boldsymbol \omega}  + {{\boldsymbol \tau _m}} + {{{\bm d}_\tau }}} \right),  
\label{eq:rotational dynamics}
\end{align}
\label{eq:quadrotor model}%
\end{subequations}
where the disturbance forces ${{\bm d}_f} = \left[ {d_{fx};d_{fy};d_{fz}} \right]$ and torques ${{\bm d}_\tau } = \left[ {d_{\tau x};d_{\tau y};d_{\tau z} } \right]$ are expressed in $\cal{\bm I} $ and ${\cal {\bm B}}$, respectively, $g$ is the gravitational acceleration, ${\bm e}_3  = {\left[ {0;0;1} \right]}$, ${ \bm \omega ^ \times }$ denotes the skew-symmetric matrix form of $\bm \omega$ as an element of the Lie algebra $\mathfrak{so}(3)$, and $f$ and ${\bm \tau_m} = {\left[ {{\tau _{mx}};{\tau _{my}};{\tau _{mz}}} \right]}$ are the total thrust and control torques produced by the quadrotor's four motors, respectively. We define ${ {\bm x}^q} = {\left[ {{\bm p};{\bm v};vec\left( {\bm R} \right);{\bm \omega} } \right]}$ as the quadrotor state where $vec\left(  \cdot  \right)$ denotes the vectorization of a given matrix and ${{\bm u}} = {\left[ {f;{\tau _{mx}};{\tau _{my}};{\tau _{mz}}} \right]}$ as the control input. Hereafter, we denote by ${\bm a} = \left[ {a;b;\cdots } \right]$ a column vector and ${\bm b} = \left[ {a,b,\cdots } \right]$ a row vector. 

The disturbance can come from various sources (e.g., aerodynamic drag or tension force from cable-suspended payloads). A general way to model its dynamics is using random walks:
\begin{equation}
    {\dot {\bm d}_f} = {\bm w _f},\ {\dot {\bm d}_\tau } = {\bm w _\tau },
    \label{eq: random walk}
\end{equation}
where $\bm w_f$ and $\bm w_{\tau}$ are the process noises. This model has been proven very efficient in estimation of unknown and time-varying disturbances~\cite{mckinnon2020estimating}. Augmenting the quadrotor state $\bm x^q$ with the disturbance state $\bm d = {\left[ {\bm d_f;\bm d_\tau} \right]}$, we define the augmented state $\bm x = \left[ {\bm p;\bm v;{\bm d_f};vec\left( \bm R \right);\bm \omega ;{\bm d_\tau }} \right]\in \mathbb{R}^{n}$. The overall dynamics model can be written as:
\begin{subequations}
\begin{align}
    \dot {\bm x} &= {{\bm f}_{{\rm{dyn}}}}\left( {{\bm x},{\bm u},\bm w} \right), \\
    {\bm y} &= {\bm h}\left( {\bm x} \right) + {\bm \nu},
    \label{eq:output dynamics}
\end{align}
\label{eq:robotic model}%
\end{subequations}
where $\bm f_{\rm dyn}$ consists of both the quadrotor rigid body dynamics (\ref{eq:quadrotor model}) and the disturbance model (\ref{eq: random walk}), $\bm w  = \left[ {{\bm w _f};{\bm w _\tau }} \right]\in \mathbb{R}^{m}$ is the process noise vector, $\bm h$ is the measurement function, and ${\bm y}$ is the measurement subject to the noise ${\bm \nu}$. The model (\ref{eq:robotic model}) will be used in MHE for estimation.

\subsection{Moving Horizon Estimator}
MHE is a control-theoretic estimator which solves a nonlinear dynamic optimization online in a receding horizon manner. At each time step $t \geq N$, the MHE estimator optimizes over $N+1$ state vectors $\bm {\mathop{ {\rm x}}\nolimits}  = \left\{ {{\bm x_k}} \right\}_{k = t - N}^t$ and $N$ process noise vectors $\bm {\mathop{ {\rm w}}\nolimits}  = \left\{ {{\bm w_k}} \right\}_{k = t - N}^{t - 1}$ on the basis of $N+1$ measurements $\mathbf{y} = \left\{ {{\bm y_k}} \right\}_{k = t - N}^t$ and $N$ control inputs $\mathbf{u} =\left\{ {{\bm u_k}} \right\}_{k = t - N}^{t - 1}$ collected in a retrospective sliding window of length $N$. We denote by $\hat {\mathbf{x}} = \left\{ {{{\hat {\bm x}}_{k\left| t \right.}}} \right\}_{k = t - N}^t$ and $\hat {\mathbf{w}} = \left\{ {{{\hat {\bm w}}_{k\left| t \right.}}} \right\}_{k = t - N}^{t - 1}$ the MHE estimates of $\bm {\mathop{ {\rm x}}\nolimits}$ and $\bm {\mathop{ {\rm w}}\nolimits}$, respectively at time step $t$. They are the solutions to the following optimization problem:
\begin{subequations}
\begin{align}
\begin{split}
\mathop {\min }\limits_{\bm {\mathop{ {\rm x}}\nolimits}, \bm {\mathop{ {\rm w}}\nolimits} } J & = \underbrace{\frac{1}{2}\left\| {{{ {\bm x}}_{t - N}} - {{\hat{\bm x}}_{t - N}}} \right\|_{\bm P}^2}_{\rm arrival \ cost}\\
 &\quad + \underbrace{\frac{1}{2}\sum\limits_{k = t - N}^t {\left\| {{\bm y_k} - {\bm h}\left( {{{ {\bm x}}_{k}}} \right)} \right\|_{{\bm R_k}}^2}  + \frac{1}{2}\sum\limits_{k = t - N}^{t - 1} {\left\| {{{\bm w} _k}} \right\|_{{{\bm Q}_k}}^2}}_{\rm running \ cost} 
\end{split}
\label{eq:mhe cost}
\\
    {\rm s.t.}\ & { {\bm x}_{k + 1}} = {\bm f}\left( {{{ {\bm x}}_{k}},{{\bm u}_k},{{{\bm w}} _k},\Delta t} \right),
\label{eq:mhe equality constraint}
\end{align}
\label{eq:mhe}%
\end{subequations}
where all of the norms are weighted by the positive-definite matrices $\bm P$, ${\bm R}_k$, and ${\bm Q}_k$, e.g., $\left\| {{{\bm w} _k}} \right\|_{{\bm Q_k}}^2 = {\bm w} _k^T{\bm Q_k}{{\bm w} _k}$, $\Delta t$ is the sampling time, ${\bm f}\left( {{{{\bm x}}_{k}},{{\bm u}_k},{{\bm w} _k},\Delta t} \right)$ is the discrete-time model of $\bm f_{\rm dyn}$ for predicting the state, and ${{ \hat{\bm x}}_{t - N}}$ is the filter priori which is chosen as the MHE estimate ${{\hat {\bm x}}_{t - N\left| {t-1} \right.}}$ of $\bm x_{t-N}$ obtained at $t-1$~\cite{alessandri2010advances}. The first term of (\ref{eq:mhe cost}) is referred to as the arrival cost, which summarizes the historical running cost before the current estimation horizon~\cite{rao2003constrained}. The second and third terms of (\ref{eq:mhe cost}) are the running cost, which minimizes the predicted measurement error and state error, respectively.

\section{Formulation of NeuroMHE} \label{section: neuromhe}
\subsection{Problem Statement}
The MHE performance depends on the choice of the weighting matrices. 
Their tuning typically requires a priori knowledge of the covariances of the noises that enter the dynamic systems. For example, the higher the covariance is, the lower confidence we have in the polluted data. Then the weightings should be set smaller for minimizing the predicted estimation error in the running cost. Despite these rough intuitions, the tuning of the weighting matrices is still demanding due to their large number and complex coupling. It becomes even more challenging when the optimal weightings are non-stationary in reality.

For the ease of presentation, we represent all the tunable parameters in the weighting matrices $\bm P$, ${\bm R}_k$, and ${\bm Q}_k$ as $\boldsymbol\theta =\left [ vec\left ( {\bm P} \right ),\left \{ vec\left ( \bm {R}_k \right ) \right \}^{t}_{k=t-N},\left \{  vec\left ( \bm {Q}_k \right )\right \}^{t-1}_{k=t-N} \right ] \in \mathbb{R}^{p}$, parameterize Problem (\ref{eq:mhe}) as $\rm {MHE}\left (\boldsymbol \theta \right)$, and denote the corresponding MHE estimates by $\hat{\bm {\mathop{ {\rm x}}\nolimits}} \left( \boldsymbol \theta \right)$. If the ground-truth system state is available, we can directly evaluate the estimation quality using a differentiable scalar loss function $L\left( {\hat {\bm {\mathop{ {\rm x}}\nolimits}}\left( \boldsymbol \theta  \right)} \right)$ built upon the estimation error. Probably more meaningful in robust trajectory tracking control, $L\left( {\hat {\bm {\mathop{ {\rm x}}\nolimits}}\left( \bm \theta  \right)} \right)$ can be chosen to penalize the tracking error. Thus, the tuning problem is to find optimal weightings $\boldsymbol \theta^*$ such that the tracking error loss is minimized. It can be interpreted as the following optimization problem:
\begin{subequations} 
\begin{align}
    \mathop {\min }\limits_{\boldsymbol \theta}\   & L\left( {\hat {\mathbf{x}}\left(\boldsymbol \theta \right)} \right)
    \label{eq:high level}\\
    {\rm s.t.}\ &\hat {\mathbf{x}}\left(\boldsymbol \theta \right) \ {\rm generated\ by} \ {\rm{MHE}}\left( {\boldsymbol \theta}\right).
    \label{eq:low level}
\end{align}
\label{eq:bi-level mhe}%
\end{subequations}
Traditionally, a fixed $\bm \theta^*$ is tuned for the vanilla MHE, as in our previous conference paper~\cite{wang2021differentiable} by solving Problem (\ref{eq:bi-level mhe}). The fixed $\bm \theta^*$ may not be optimal and can lead to substantially degraded performance when the noise covariances are non-stationary. Therefore, we are interested in designing an adaptive $\bm \theta^*$ to improve the estimation or control performance. 

\subsection{Neural MHE Adaptive Weightings}
The adaptive weightings are generally difficult to model using the first principle. Instead, we employ a neural network (NN) to approximate them, denoted by
\begin{equation}
    \bm \theta  = {{\bm f}_{\boldsymbol \varpi} }\left( {\boldsymbol \chi}  \right),
    \label{eq:nn parameterization}
\end{equation}
where $\boldsymbol \varpi$ denotes the neural network parameters, and the input $\boldsymbol \chi$ can be various signals such as quadrotor state ${\bm x}^q$, control input $\bm u$, or other external signals, depending on the priori knowledge and applications.

\subsection{Tuning NeuroMHE via Gradient Descent}\label{subsection:joint optimization}

For NeuroMHE, the tuning becomes finding the optimal neural network parameters $\bm \varpi^*$ to minimize the loss function, that is,
\begin{subequations}
\begin{align}
\mathop {\min }\limits_{\boldsymbol \varpi}\   &L\left( {\hat {\mathbf{x}}\left(\boldsymbol \varpi \right)} \right)
\\
{\rm s.t.}\ & \hat {\mathbf{x}}\left(\boldsymbol \varpi \right)\ {\rm generated \ by} \ {\rm{NeuroMHE}}\left({\boldsymbol \theta\left( {\boldsymbol \varpi} \right)} \right).
\label{eq:low level neuromhe}
\end{align}
\label{eq:bi level neuromhe}%
\end{subequations}
Compared with the fixed $\bm \theta^*$ employed in (\ref{eq:bi-level mhe}), the neural network parameterized $\bm \theta^*\left( \bm \varpi^*  \right)$ is more powerful, which varies according to different states or external signals $\bm \chi$, thus allowing for fast online adaptation to different flight status and scenarios.
 
We aim to use gradient descent to train $\bm \varpi$. The gradient of $L\left( {\hat {\bm {\mathop{ {\rm x}}\nolimits}}\left(\bm \varpi \right)} \right)$ with respect to $\bm \varpi$ can be calculated using the chain rule
\begin{equation}
    \frac{{d L\left( {\hat {\mathbf{x}}}\left(\bm \varpi \right) \right)}}{{d {\bm \varpi} }} = \frac{{\partial L\left( \hat{\mathbf{x}} \right)}}{{\partial \hat {\mathbf{x}}}}\frac{{\partial \hat {\mathbf{x}}\left(\bm \theta \right)}}{{\partial \bm \theta }}\frac{{\partial \bm \theta\left(\bm \varpi \right) }}{{\partial {\bm \varpi} }}.
    \label{eq: gradient of loss}
\end{equation}
Fig. \ref{fig:neuromhe learning pipeline} depicts the learning pipelines for training NeuroMHE. Each update of $\bm \varpi$ consists of a \textit{forward pass} and a \textit{backward pass}. In the forward pass, given the current $\bm \varpi_t$, the weightings $\bm \theta_t$ are generated from the neural network for MHE to obtain $\hat{\bm {\mathop{ {\rm x}}\nolimits}}$, and thus $L\left( {\hat {\bm {\mathop{ {\rm x}}\nolimits}}\left( \bm \varpi_t  \right)} \right)$ is evaluated. In the backward pass, the gradients components $\frac{{\partial L\left( {\hat {\bm {\mathop{ {\rm x}}\nolimits}}} \right)}}{{\partial \hat {\bm {\mathop{ {\rm x}}\nolimits}}}}$, $\frac{{\partial \hat {\bm {\mathop{ {\rm x}}\nolimits}}\left(\bm \theta \right)}}{{\partial \bm \theta }}$, and $\frac{{\partial \bm \theta\left(\bm \varpi \right) }}{{\partial {\bm \varpi} }}$ are computed.

In solving the MHE, $\hat{\mathbf{x}}$ can be obtained by any numerical optimization solver. In the backward pass, the gradient $\frac{\partial L\left ( \hat {\mathbf{x}} \right )}{\partial \hat {\mathbf{x}}}$ is straightforward to compute as $ L\left( {\hat {\mathbf{x}}} \right)$ is generally an explicit function of $\hat {\mathbf{x}}$; computing $\frac{{\partial \bm \theta\left(\bm \varpi \right) }}{{\partial {\bm \varpi} }}$ for the neural network is standard and handy via many existing machine learning tools. The main challenge is how to solve for $\frac{{\partial \hat {\mathbf{x}}\left(\bm \theta \right)}}{{\partial \bm \theta }}$, the gradients of the MHE estimates with respect to $\bm \theta$. This requires differentiating through the MHE problem (\ref{eq:mhe}) which is a nonlinear optimization. Next, we will derive  $\frac{{\partial \hat {\mathbf{x}}\left(\bm \theta \right)}}{{\partial \bm \theta }}$ analytically and show that it can be computed recursively using a Kalman filter.

\section{Analytical Gradients} \label{section: gradient}
The idea to derive $\frac{{\partial \hat {\mathbf{x}}}}{{\partial {\bm \theta} }}$ is to implicitly differentiate through the KKT conditions associated with the MHE problem (\ref{eq:mhe}). The KKT conditions define a set of first-order optimality conditions which the locally optimal ${\hat{\mathbf{x}} }$ must satisfy. We associate the equality constraints~(\ref{eq:mhe equality constraint}) with the dual variables $\bm \lambda = \left\{ {{{ {\bm \lambda}}_{k}}} \right\}_{k = t - N}^{t-1}$ and denote their optimal values by $\bm \lambda^* \in \mathbb{R}^{n}$. Then, the corresponding Lagrangian can be written as
\begin{equation}
    {\cal L} = \frac{1}{2}\left\| {{{ {\bm x}}_{t - N}} - {{\hat{\bm x}}_{t - N}}} \right\|_{\bm P}^2 + \bar {\cal L},
    \label{eq: lagrangian}
\end{equation}
where 
\begin{equation}
    \begin{aligned}
\bar {\cal L} & = \frac{1}{2}\sum\limits_{k = t - N}^t {\left\| {{\bm y_k} - \bm h\left( {{{ {\bm x}}_{k}}} \right)} \right\|_{{\bm R_k}}^2}  + \frac{1}{2}\sum\limits_{k = t - N}^{t - 1} {\left\| {{{\bm w }_{k}}} \right\|_{{\bm Q_k}}^2} \\
&\quad + \sum\limits_{k = t - N}^{t - 1} {{\bm \lambda} _k^T\left( {{{ {\bm x}}_{k + 1}} - {\bm f}\left( {{{ {\bm x}}_{k}},{{\bm u}_k},{{\bm w} _{k}},\Delta t} \right)} \right)}. 
\end{aligned}
\label{eq:component in lagrangian}
\nonumber
\end{equation}
Then the KKT conditions of (\ref{eq:mhe}) at ${\hat{\bm {\mathop{ {\rm x}}\nolimits}} }$, ${\hat{\bm {\mathop{ {\rm w}}\nolimits}} }$, and $\bm \lambda^*$ are given by
\begin{subequations}
\begin{align}
\begin{split}
{\nabla _{{\hat x}_{t-N\left| t \right.}}}{{\cal L}} & = {\bm P}\left( {{{\hat {\bm x}}_{t - N\left| t \right.}}- {{\hat{\bm x}}_{t - N}}} \right)\\
&\quad - {\bm H}_{t - N}^T{{\bm R}_{t - N}}\left( {{{\bm y}_{t - N}} - {\bm h}\left( {{{\hat {\bm x}}_{t - N\left| t \right.}}} \right)} \right)\\
 &\quad  - {\bm F}_{t - N}^T{\bm \lambda^* _{t - N}}= \bm 0,
\end{split}
\label{eq:kkt boundary}\\
\begin{split}
{\nabla _{{\hat x}_{k\left| t \right.}}}{{\cal { L}}} & = {\bm \lambda^* _{k - 1}} - {\bm H}_k^T{\bm R_k}\left( {{{\bm y}_k} - {\bm h}\left( {{{\hat {\bm x}}_{k\left| t \right.}}} \right)} \right)\\
&\quad- {\bm F}_k^T{\bm \lambda_k^*} = \bm 0,\ k = t - N + 1, \cdots ,t,
\end{split}
\label{eq:kkt x}\\
\begin{split}
{\nabla _{{\hat w}_{k\left| t \right.}} }{{\cal { L}}} & = {\bm Q_k}{\hat{\bm w} _{k\left| t \right.}}- {\bm G}_k^T{\bm \lambda_k^*} = \bm 0,\\
&\quad \ k = t - N, \cdots ,t - 1,    
\end{split}
\label{eq:kkt noise}\\
\begin{split}
{\nabla _{\lambda^{*}_k} }{{\cal { L}}} & =  {{\hat {\bm x}}_{k + 1\left| t \right.}}- {\bm f}\left( {{{\hat {\bm x}}_{k\left| t \right.}},{\bm u_k},{\hat{\bm w} _{k\left| t \right.}},\Delta t} \right)  = \bm 0,\\
&\quad \ k = t - N, \cdots ,t - 1,
\end{split}
\label{eq:kkt lambda}
\end{align}
\label{eq:kkt}%
\end{subequations}
where
\begin{equation}
    {\bm F_k} = \frac{{\partial \bm f}}{{\partial {{\hat {\bm x}}_{k\left| t \right.}}}},\ {\bm G_k} = \frac{{\partial \bm f}}{{\partial {\hat{\bm w} _{k\left| t \right.}}}},\ {\bm H_k} = \frac{{\partial \bm h}}{{\partial {{\hat {\bm x}}_{k\left| t \right.}}}},
    \label{eq:definition of system matrices}
\end{equation}
and $\bm \lambda^*_t = \bm 0$ by definition. Equations (\ref{eq:kkt boundary}), (\ref{eq:kkt x}), and (\ref{eq:kkt noise}) are the Lagrangian stationarity conditions at ${\hat{\bm {\mathop{ {\rm x}}\nolimits}} }$, ${\hat{\bm {\mathop{ {\rm w}}\nolimits}} }$, and $\bm \lambda^*$. Eq. (\ref{eq:kkt lambda}) is the primal feasibility from the system dynamics.

\subsection{Differential KKT Conditions of MHE}
\label{subsec:diffkkt}
Recall that $\hat {\mathbf{x}} = \left\{ {{{\hat {\bm x}}_{k\left| t \right.}}} \right\}_{k = t - N}^t$ is the optimal estimates from $t-N$ to $t$. To calculate
\begin{equation}
    \frac{{\partial \hat {\mathbf{x}}}}{{\partial {\bm \theta} }}=\left\{ {\frac{{\partial {{\hat {\bm x}}_{k\left| t \right.}}}}{{\partial {\bm \theta} }}} \right\}_{k = t - N}^t,
    \label{eq:gradient trajectory}
\end{equation}
we are motivated to implicitly differentiate the KKT conditions (\ref{eq:kkt}) on both sides with respect to $\bm \theta$. This results in the following \textit{differential} KKT conditions.
\begin{subequations}
\begin{align}
    \begin{split}
\frac{{d{\nabla _{{\hat {x}_{t - N\left| t \right.}}}}{\cal L}}}{{d\bm \theta }} & =  { {\bm L}_{t - N}^{xx}} \frac{{\partial {{\hat {\bm x}}_{t - N\left| t \right.}}}}{{\partial {\bm \theta} }} - {\bm P}\frac{{\partial {{{ \hat{\bm x}}_{t - N}}}}}{{\partial {\bm \theta} }}+ {\bm L}_{t - N}^{x\theta } \\
& \quad  + {\bm L}_{t - N}^{xw}\frac{{\partial {\hat{\bm w} _{t-N\left| t \right.}}}}{{\partial {\bm \theta} }}- \mathrlap {{\bm F}_{t - N}^T\frac{{\partial {\bm \lambda^* _{t - N}}}}{{\partial {\bm \theta} }}= {\bm 0}},
    \end{split}
    \label{eq:boundary differential KKT}\\
    \begin{split}
\frac{{d{\nabla _{{\hat{x}_{k\left| t \right.}}}}{\cal { L}}}}{{d\bm \theta }} & = {\bm L}_k^{xx}\frac{{\partial {{\hat {\bm x}}_{k\left| t \right.}}}}{{\partial {\bm \theta} }} + {\bm L}_k^{xw }\frac{{\partial {\hat{\bm w} _{k\left| t \right.}}}}{{\partial {\bm \theta} }} - {\bm F}_k^T\frac{{\partial {\bm \lambda_k^*}}}{{\partial {\bm \theta} }}\\
&\quad + \frac{{\partial {\bm \lambda^* _{k - 1}}}}{{\partial {\bm \theta} }} + {\bm L}_k^{x\theta } = {\bm 0},\\
&\quad \ k = t - N + 1, \cdots ,t,
    \end{split}
    \label{eq:x differential KKT}\\
    \begin{split}
\frac{{d{\nabla _{{\hat{w} _{k\left| t \right.}}}}{\cal { L}}}}{{d\bm \theta }} & = {\bm L}_k^{w x}\frac{{\partial {{\hat {\bm x}}_{k\left| t \right.}}}}{{\partial {\bm \theta} }} + {\bm L}_k^{w w}\frac{{\partial {\hat{\bm w} _{k\left| t \right.}}}}{{\partial {\bm \theta} }} - {\bm G}_k^T\frac{{\partial {\bm \lambda_k^*}}}{{\partial {\bm \theta} }}\\
&\quad  + {\bm L}_k^{w \theta } = {\bm 0},\ k = t - N, \cdots ,t - 1,
    \end{split}
    \label{eq:w differential KKT}\\
    \begin{split}
\frac{{d{\nabla _{{\lambda^{*} _k}}}{\cal { L}}}}{{d\bm \theta }} & = \frac{{\partial {{\hat {\bm x}}_{k + 1\left| t \right.}}}}{{\partial {\bm \theta} }} - {\bm F_k}\frac{{\partial {{\hat {\bm x}}_{k\left| t \right.}}}}{{\partial {\bm \theta} }} - \mathrlap{{\bm G_k}\frac{{\partial {\hat{\bm w} _{k\left| t \right.}}}}{{\partial {\bm \theta} }} = {\bm 0}},\\
&\quad \ k = t - N, \cdots ,t - 1,
    \end{split}
    \label{eq:costate differential KKT}
\end{align}
\label{eq:differential KKT}%
\end{subequations}
where the coefficient matrices are defined as follows:
\begin{subequations}
\begin{align}
    {\bm L}_k^{xx} & = \frac{{{\partial ^2}{\cal L}}}{{\partial \hat {\bm x}_{k\left| t \right.}^2}},&  {\bm L}_k^{xw } & = \frac{{{\partial ^2} {\cal L}}}{{\partial {{\hat {\bm x}}_{k\left| t \right.}}\partial {\hat{\bm w} _{k\left| t \right.}}}}, & {\bm L}_k^{x\theta } &= \frac{{{\partial ^2}{\cal L}}}{{\partial {{\hat {\bm x}}_{k\left| t \right.}}\partial {\bm \theta} }},\\
    {\bm L}_k^{w w } & = \frac{{{\partial ^2} {\cal L}}}{{\partial {\hat{\bm w}} _{k\left| t \right.}^2}},& {\bm L}_k^{w x} & = \frac{{{\partial ^2} {\cal L}}}{{\partial {{\hat{\bm w}} _{k\left| t \right.}}\partial {{\hat {\bm x}}_{k\left| t \right.}}}},& {\bm L}_k^{w \theta } & = \frac{{{\partial ^2}{\cal L}}}{{\partial {\hat{\bm w} _{k\left| t \right.}}\partial {\bm \theta} }},
\end{align}
\label{eq:coefficient matrices}%
\end{subequations}
for $k = t - N, \cdots ,t$. Note that $\frac{{\partial {\bm y_{t - N}}}}{{\partial {\bm \theta} }}$ and $\frac{{\partial {\bm y_k}}}{{\partial {\bm \theta} }}$ are zeros as the measurements are independent of $\bm \theta$. The analytical expressions of all the matrices defined in (\ref{eq:definition of system matrices}) and (\ref{eq:coefficient matrices}) can be obtained via any software package that supports symbolic computation (e.g., CasADi~\cite{andersson2019casadi}), and their values are known once the estimated trajectories $\hat{\mathbf{x}}$, $\hat{\mathbf{w}}$, and $\bm \lambda^*$ are obtained in the forward pass. Note that $\frac{{\partial {{{\hat{\bm x}}_{t - N}}}}}{{\partial {\bm \theta} }}$ can be approximated by $\frac{{\partial {{{\hat{\bm x}}_{t - N\left| t-1 \right.}}}}}{{\partial {\boldsymbol \theta} }}$ at the previous time step $t-1$ (See Appendix-\ref{appendix:approximation}). We denote $\frac{{\partial {{{ \hat{\bm x}}_{t - N}}}}}{{\partial {\boldsymbol \theta} }}$ by $\hat{\bm X}_{t-N}$ to distinguish it from the unknown matrices $ {\frac{{\partial {{\hat {\bm x}}_{k\left| t \right.}}}}{{\partial {\boldsymbol \theta} }}} $, $ {\frac{{\partial {\hat{\bm w} _{k\left| t \right.}}}}{{\partial {\boldsymbol \theta} }}} $, and $ {\frac{{\partial {\boldsymbol \lambda_k^*}}}{{\partial {\boldsymbol \theta} }}} $. Next, we will demonstrate that these unknowns can be computed recursively by the Kalman filter-based solver proposed in the following subsection.

\subsection{Kalman Filter-based Gradient Solver}
From the definitions in \eqref{eq:coefficient matrices}, we can further calculate that ${\bm L}_{t - N}^{xx} = {\bm P} + \bar {\bm L}_{t - N}^{xx}$, where $\bar {\bm L}_{t - N}^{xx} = \frac{{{\partial ^2}\bar {\cal L}}}{{\partial \hat {\bm x}_{t - N\left| t \right.}^2}}$. Plugging it back to (\ref{eq:boundary differential KKT}), we can see that the differential KKT conditions (\ref{eq:differential KKT}) have a similar structure to the original KKT conditions (\ref{eq:kkt}). More importantly, it can be interpreted as the KKT conditions of an auxiliary linear MHE system whose optimal state estimates are exactly the gradients (\ref{eq:gradient trajectory}). To formalize it, we define ${ {\bm X}_{k}}   = \frac{{\partial {{ {\bm x}}_{k}}}}{{\partial {\bm \theta} }}\in \mathbb{R}^{n\times p}$ as the new state, ${{\bm W}_{k}}  = \frac{{\partial {{{\bm w}}_{k}}}}{{\partial {\bm \theta} }}\in \mathbb{R}^{m\times p}$ as the new process noise, and ${{\bm \Lambda_k} }   = \frac{{\partial {{\bm \lambda_k} }}}{{\partial {\bm \theta} }}\in \mathbb{R}^{n\times p}$ as the new dual variable of the following auxiliary MHE system:
\begin{subequations}
\begin{align}
\begin{split}
\mathop {\min }\limits_{\bm {\mathop{\rm X}\nolimits},\bm {\mathop{\rm W}\nolimits}} {J_2} & = \frac{1}{2}{\rm{Tr}}\left\| {{{ {\bm X}}_{t - N}} - \hat{\bm X}_{t-N}} \right\|_{\bm P}^2\\
&\quad + {\rm{Tr}}\sum\limits_{k = t - N}^t {\left( {\frac{1}{2} {\bm X}_{k}^T\bar {\bm L}_k^{xx}{{ {\bm X}}_{k}} + {{\bm W}_k}^T {\bm L}_k^{w x}  {\bm X}_{k} } \right)} \\
&\quad + {\rm{Tr}}\sum\limits_{k = t - N}^{t - 1} {\left( {\frac{1}{2}{\bm W}_k^T {\bm L}_k^{w w}{{\bm W}_k} +{{{\left( {{\bm L}_k^{w \theta }} \right)}^T}{{\bm W}_k}} }\right)}\\
&\quad + {\rm{Tr}}\sum\limits_{k = t - N}^{t} {\left( {{\left( {{\bm L}_k^{x\theta }} \right)}^T} {{ {\bm X}}_{k}}  \right)} 
\end{split}
\label{eq:auxiliary mhe cost}\\
{\rm s.t.}\ & {{ {\bm X}}_{k + 1}} = {\bm F_k}{{ {\bm X}}_{k}} + {\bm G_k}{{\bm W}_k},
\label{eq:auxiliary dynamics}
\end{align}
\label{eq: auxiliary MHE}%
\end{subequations}
where ${\mathbf{X}} = \left\{ {{{ {\bm X}}_{k}}} \right\}_{k = t - N}^t$, ${\mathbf{W}} = \left\{ {{{ {\bm W}}_{k}}} \right\}_{k=t - N}^{t - 1}$, $\bar {\bm L}_k^{xx} = {\bm L}_k^{xx}$ for $k = t - N + 1, \cdots ,t $, and ${\rm Tr}\left(  \cdot  \right)$ denotes the matrix trace. 

\begin{lemma}
Let $\hat {\mathbf{X}}$ and $\hat{\mathbf{W}}$ be the optimal estimates of ${\mathbf{X}}$ and ${\mathbf{W}}$, respectively, which are the stationary solutions to the auxiliary MHE system (\ref{eq: auxiliary MHE}). Then, they satisfy the KKT conditions of (\ref{eq: auxiliary MHE}) which are defined in (\ref{eq:differential KKT}), and
\begin{equation}
    \hat{\mathbf{X}}=\left \{ \frac{\partial \hat{\bm x}_{k\left| t \right.}}{\partial \boldsymbol \theta} \right \}_{k=t-N}^{t}, \quad \hat{\mathbf{W}}=\left \{ \frac{\partial \hat{\bm w}_{k\left| t \right.}}{\partial \boldsymbol \theta} \right \}_{k=t-N}^{t-1}.
    \label{eq: equivalent gradient}
\end{equation}
\label{lemma: auxiliary MHE}%
\end{lemma}

A proof of \Cref{lemma: auxiliary MHE} can be found in Appendix-\ref{appendix:lemma1}. This lemma shows that the KKT conditions of (\ref{eq: auxiliary MHE}) are the same as the differential KKT conditions (\ref{eq:differential KKT}) of the original MHE problem (\ref{eq:mhe}). In particular, the solution $\hat{\bm {\mathop{ {\rm X}}\nolimits}}$ of \eqref{eq: auxiliary MHE} are exactly the gradients of $\hat{\bm {\mathop{ {\rm x}}\nolimits}}$ with respect to $\bm \theta$. Hence, we can compute the desired gradients (\ref{eq:gradient trajectory}) by solving the auxiliary MHE system (\ref{eq: auxiliary MHE}). Notice that for this auxiliary MHE, its cost function (\ref{eq:auxiliary mhe cost}) is quadratic and the dynamics model (\ref{eq:auxiliary dynamics}) is linear. Therefore, it is possible to obtain ${\hat {\bm {\mathop{ {\rm X}}\nolimits}}}$ in closed form. Instead of solving the auxiliary MHE by some numerical optimization solver, we propose a computationally more efficient method to recursively solve for ${\hat {\bm {\mathop{ {\rm X}}\nolimits}}}$ using a Kalman filter. We now present the key result in the following lemma, which is obtained using forward dynamic programming~\cite{cox1963estimation}.

\begin{lemma}
The optimal estimates $\hat{\bm {\mathop{ {\rm X}}\nolimits}}$ of the auxiliary MHE (\ref{eq: auxiliary MHE}) can be obtained recursively through the following steps. 

Step 1: A Kalman filter (KF) is solved to generate the estimates $\left\{ {\hat {\bm X}_{k\left| k \right.}^{\rm KF}}\right\} _{k = t-N}^t$: at $k=t-N$ with $\hat {\bm X}_{t-N}$ obtained at $t-1$ (See \ref{subsec:diffkkt}), the initial conditions are given by
\begin{subequations}
\begin{align}
      {\bm P_{t-N}} & = {\bm P^{-1}},
      \label{eq: initial condition for P}\\
     {\bm C_{t - N}}& = {\left( {\bm I - {\bm P_{t - N}}{\bm S_{t - N}}} \right)^{ - 1}}{\bm P_{t - N}},
    \label{eq: initial condition for C}\\
    \begin{split}
    {\hat {\bm X}_{t - N\left| t \right. - N}^{\rm KF}} &=\left( {\bm I + {\bm C_{t - N}}{\bm S_{t - N}}} \right)\hat {\bm X}_{t-N}  + {\bm C_{t - N}}{\bm T_{t - N}}.
    \end{split}
    \label{eq: initial condition for x}
\end{align}
\label{eq: initial conditions lemma 2}%
\end{subequations}
Then, the remaining estimates $\left\{ {\hat {\bm X}_{k\left| k \right.}^{\rm KF}}\right\} _{k = t-N+1}^t$ are obtained by solving the following equations from $k=t-N+1$ to $t$:
\begin{subequations}
   \begin{align}
       {{\hat {\bm X}}_{k\left| {k - 1} \right.}} & = {{\bar {\bm F}}_{k - 1}}{{\hat {\bm X}}_{k - 1\left| {k - 1} \right.}^{\rm KF}} - {\bm G_{k - 1}}{\left( { {\bm L}_{k - 1}^{w w }} \right)^{ - 1}}{\bm L}_{k - 1}^{w \theta },
       \label{eq: kalman filter prediction}\\
       {\bm P_k} & = {{\bar {\bm F}}_{k - 1}}{\bm C_{k - 1}}\bar {\bm F}_{k - 1}^T + {\bm G_{k - 1}}{\left( { {\bm L}_{k - 1}^{w w }} \right)^{ - 1}}\bm G_{k - 1}^T,
       \label{eq: kalman filter covariance}\\
       {\bm C_k}& = {\left( {\bm I - {\bm P_k}{\bm S_k}} \right)^{ - 1}}{\bm P_k},
       \label{eq: kalman filter covariance update}\\
       {{\hat {\bm X}}_{k\left| k \right.}^{\rm KF}} & = \left( {\bm I + {\bm C_k}{\bm S_k}} \right){{\hat {\bm X}}_{k\left| {k - 1} \right.}} + {\bm C_k}{\bm T_k}.
       \label{eq: kalman filter estimate}
   \end{align}
    \label{eq: kalman}%
\end{subequations}
Here, $\bm I$ is an identity matrix with an appropriate dimension, the matrices ${\bm S_k}$, ${\bm T_k}$, and ${\bar {\bm F}_k}$ are defined as follows:
\begin{equation}
\begin{aligned}
    {\bm S_k}  & = {\bm L}_k^{xw }{\left( { {\bm L}_k^{w w }} \right)^{ - 1}}{\bm L}_k^{w x} - \bar {\bm L}_k^{xx}, \ k=t-N,\cdots,t-1,\\
    {\bm T_k}  & = {\bm L}_k^{xw }{\left( { {\bm L}_k^{w w }} \right)^{ - 1}}{\bm L}_k^{w \theta } - {\bm L}_k^{x\theta }, \ k=t-N,\cdots,t-1,\\
    {\bar {\bm F}_k}  & = {\bm F_k} - {\bm G_k}{\left( {{\bm L}_k^{w w }} \right)^{ - 1}} {\bm L}_k^{w x}, \ k=t-N,\cdots,t-1,\\
    {\bm S_t}  & = - \bar {\bm L}_t^{xx},\ {\bm T_t}  = - {\bm L}_t^{x\theta }.
\end{aligned}
\nonumber
\end{equation}

Step 2: The new dual variables $\bm \Lambda^*  = \left\{ {{\bm \Lambda^* _k}} \right\}_{k = t - N}^{t - 1}$ are computed iteratively by the following equation from $k=t$ to $t-N+1$, starting with $\bm \Lambda^*_t = \bm 0$:
\begin{equation}
    {\bm \Lambda^* _{k - 1}} = \left( {\bm I + {\bm S_k}{\bm C_k}} \right)\bar {\bm F}_k^T{\bm \Lambda^* _k} + {\bm S_k}{\hat {\bm X}_{k\left| k \right.}^{\rm KF}} + {\bm T_k}.
    \label{eq: backward lambda}
\end{equation}

Step 3: The optimal estimates $\hat{\bm {\mathbf{X}}}$ are computed by 
\begin{equation}
    {\hat {\bm X}_{k\left| t \right.}} = {\hat {\bm X}_{k\left| k \right.}^{\rm KF}} + {\bm C_k}\bar {\bm F}_k^T{\bm \Lambda^* _k},
    \label{eq: forward state}
\end{equation}
iteratively from $k=t-N$ to $t$.
\label{lemma:analytical gradient}%
\end{lemma}

Equations in (\ref{eq: initial conditions lemma 2}) and (\ref{eq: kalman}) represent a Kalman filter with the matrix state ${\hat {\bm X}_{k\left| k \right.}^{\rm KF}}$ and "zero measurement" (due to $\frac{{\partial {\bm y_k}}}{{\partial {\bm \theta} }} = {\bm 0}$). Among them, (\ref{eq: kalman filter prediction}) serves as the state predictor, (\ref{eq: initial condition for P}) and (\ref{eq: kalman filter covariance}) handle the error covariance prediction, (\ref{eq: initial condition for C}) and (\ref{eq: kalman filter covariance update}) address the error covariance correction, and finally (\ref{eq: initial condition for x}) and (\ref{eq: kalman filter estimate}) function the state correctors. Note that these equations are not expressed in the standard form of a Kalman filter mainly for the ease of presenting its proof in a nice inductive way (See Appendix-\ref{appendix:lemma2}). The more familiar form can be easily obtained. For example, the standard Kalman gain can be extracted from (\ref{eq: kalman filter estimate}) using the matrix inversion lemma~\cite{cox1963estimation}. The Kalman filter provides the analytical gradients in a recursive form, making their computation very efficient. This appealing recursive nature is theoretically justified by a proof of Lemma~\ref{lemma:analytical gradient} in Appendix-\ref{appendix:lemma2}. The gradient solver acts as a key component in the backward pass of training NeuroMHE, as shown in Fig. \ref{fig:neuromhe learning pipeline}. We summarize the procedure of solving for ${\hat {\bm {\mathop{ {\rm X}}\nolimits}}}$ using a Kalman filter in Algorithm \ref{alg: analytical solution}. 

\begin{algorithm}[!h]
\caption{Solving for $\hat {\mathbf{X}}$ using a Kalman filter}
\label{alg: analytical solution}
\SetKwInput{Input}{Input}
\SetKwInput{Initialization}{Initialization}
\SetKw{by}{by}
\SetKwProg{Pn}{def}{:}{\KwRet $\left\{ {{{\hat {\bm X}}_{k\left| t \right.}}} \right\}_{k = t - N}^t$}
\Input{The trajectories $\hat {\mathbf{x}}$, $\hat {\mathbf{w}}$, and $\boldsymbol{ \lambda}^*$ generated by solving Problem (\ref{eq:mhe}), the current weightings $\bm \theta$, the control inputs ${\mathbf u}$, the previous gradient $\hat{\bm X}_{t-N}$, and the matrices in (\ref{eq:definition of system matrices}) and (\ref{eq:coefficient matrices}) for constructing the auxiliary MHE system (\ref{eq: auxiliary MHE});}
\Indp \Pn{Kalman\_Filter\_based\_Gradient\_Solver}{\Comment{implementation of Lemma~\ref{lemma:analytical gradient}}
Set ${{\hat {\bm X}}_{{t-N}\left| {{t-N}} \right.}^{\rm KF}}$ using Eq.(\ref{eq: initial conditions lemma 2});\\
\For{$k \leftarrow t-N+1$ \KwTo $t$ \by $1$}{\Comment{Kalman filter}
Update ${{\hat {\bm X}}_{{k}\left| {{k}} \right.}^{\rm KF}}$ using Eq.(\ref{eq: kalman});
}
\For{$k \leftarrow t$ \KwTo $t-N+1$ \by $-1$}{
Update ${{\bm \Lambda^*_{k-1}} }$ using Eq.(\ref{eq: backward lambda}) with ${{\bm \Lambda^*_t}} = {\bm 0}$;
}
\For{$k \leftarrow t-N$ \KwTo $t$ \by $1$}{
Update ${\hat {\bm X}_{{k}\left| t \right.}}$ using Eq.(\ref{eq: forward state}); 
}
}
\Indm\Return{$\frac{{\partial {\hat {\bm {\mathop{ {\rm x}}\nolimits}} }}}{{\partial {\bm \theta} }}= {\hat {\bm {\mathop{ {\rm X}}\nolimits}}}$ \Comment{due to Lemma \ref{lemma: auxiliary MHE}}}
\end{algorithm}

\subsection{Sparse Parameterization of Weightings}
\label{subsec:parameterization}
We design an efficient sparse parameterization for the MHE weightings. First, we introduce two forgetting factors ${\gamma _{1,2}} \in \left( {0,1} \right)$ to parameterize the time-varying weighting matrices ${\bm R}_k$ and ${\bm Q}_k$ in the MHE running cost by ${{\bm R}_k} = \gamma _1^{t - k}{{\bm R}_t}$ for $k = t - N, \cdots ,t$ and ${{\bm Q}_k} = \gamma _2^{t - 1 - k}{{\bm Q}_{t - 1}}$ for $k=t - N, \cdots ,t-1$. That is, instead of training a neural network to generate all ${\bm R}_k$ and ${\bm Q}_k$, we only need to train it to generate adaptive forgetting factors together with (adaptive) $\bm P$, $\bm R_t$, and $\bm Q_{t-1}$. This effectively reduces the size of the neural network and keeps it portable, while enabling a flexible adjustment of the weightings over the horizon. Second, we set ${\bm P}$, ${\bm R}_t$, and ${\bm Q}_{t - 1}$ to be diagonal matrices to further reduce the size of the problem. With these simplifications, we parameterize the diagonal elements as ${P_{\cdot}} = \varsigma  + p_{\cdot}^2$, ${R_{\cdot}} = \varsigma  + r_{\cdot}^2$, and ${Q_{\cdot}} = \varsigma  + q_{\cdot}^2$, respectively, where $\varsigma  > 0$ is some pre-chosen small constant and the subscript "$\cdot$" denotes the appropriate index of the corresponding diagonal element. We further use two sigmoid functions $S\left ( \bar\gamma_{1,2} \right )={\left( {1 + \exp \left( { - {{\bar \gamma }_{1,2}}} \right)} \right)^{ - 1}}$ to constrain the values of $\gamma_1$ and $\gamma_2$ to be in $\left( {\gamma _{\min},1} \right)$, where $\gamma_{\min}> 0$ is some pre-chosen small constant, and have $\gamma_{1,2}=\gamma_{\min} + \left ( 1- \gamma_{\min}\right )S\left ( \bar\gamma_{1,2} \right )$. Therefore, the output of the neural network becomes $\boldsymbol\Theta = \left [ p_{:},\bar{\gamma}_1,r_{:},\bar{\gamma}_2,q_{:} \right ]$, where the subscript "$:$" denotes the appropriate dimension of a vector that collects all the corresponding diagonal elements such as $p_{:}=\left [ p_{1},p_{2},\cdots  \right ]$. To reflect the above sparse parameterization, the chain rule for the gradient calculation is further written as
\begin{equation}
    \frac{dL}{d{\boldsymbol{\varpi }}}=\frac{\partial L}{\partial \hat{\mathbf{x}}}\frac{\partial \hat{\mathbf{x}}}{\partial \boldsymbol{\theta}}\frac{\partial \boldsymbol{\theta}}{\partial\boldsymbol{\Theta}}\frac{\partial \boldsymbol{\Theta}}{\partial \boldsymbol{\varpi }}.
    \label{eq:new chain rule}
\end{equation}

\section{Model-based Policy Gradient Algorithm}\label{section:policy gradient}
In this section, we propose a model-based policy gradient reinforcement learning (RL)  algorithm to train NeuroMHE. This algorithm enables the neural network parameters $\bm \varpi$ to be learned directly from the trajectory tracking error without the ground-truth disturbance data.

Denote the robust tracking controller by
\begin{equation}
    \bm u_t = {\bm u}\left( {{\bm x}_t^{q,{\rm{ref}}},\hat {\bm x}_{t\left| t \right.}} \right),
    \label{eq:general controller}
\end{equation}
where the estimate $\hat {\bm x}_{t\left| t \right.}$ comprises the current quadrotor state estimate $ \hat{\bm x}_{t\left| t \right.}^q$ and disturbance estimate $\hat{\bm d}_{t\left| t \right.}$, and it is computed online by solving Problem (\ref{eq:mhe}). Such a controller can be simply designed as directly compensating $\hat{\bm d}_{t\left| t \right.}$ in a nominal geometric flight controller~\cite{lee2010geometric}. Supervised training of NeuroMHE forms the estimation error $\left\| {{{\hat {\bm d}}_{t\left| t \right.}} - {\bm d_t}} \right\|$ as the loss function, which requires the ground truth disturbance. In practice, the ground truth $\bm d_t$ is often difficult to obtain. It is of great convenience to train the NeuroMHE directly by minimizing the trajectory tracking errors. For example, the loss  can be chosen as

\begin{equation}
    L\left( {\hat {\mathbf{x}}} \right) = \sum\limits_{k = t-N}^t {\left\| {{{\hat {\bm x}}_{k\left| t \right.}^q} - {\bm x}_k^{{q,\rm{ref}}}} \right\|_{{{\bm W}_e}}^2},
    \label{eq:loss function}
\end{equation}
where ${\bm W}_e$ is a positive-definite weighting matrix. It penalizes the difference between the estimated quadrotor states by MHE and the reference states.

In training, we perform gradient descent to update $\bm \varpi$. We first obtain the analytical gradients $\frac{{\partial {\hat{\mathbf{x}} }}}{{\partial {\bm \theta} }}$ using Algorithm~\ref{alg: analytical solution}, then compute $ \frac{{\partial L\left( {\hat {\mathbf{x}}} \right)}}{{\partial \hat {\mathbf{x}}}}$, $\frac{\partial \boldsymbol{\theta}}{\partial \boldsymbol{\Theta}}$, and $\frac{{\partial \boldsymbol \Theta\left(\boldsymbol \varpi \right) }}{{\partial {\bm \varpi} }}$ from their analytical expressions, and finally apply the chain rule \eqref{eq:new chain rule} to obtain the gradient $\frac{{d L\left( {\hat {\bm {\mathop{ {\rm x}}\nolimits}}} \right)}}{{d {\bm \varpi} }}$. We summarize the procedures of training NeuroMHE using the proposed model-based policy gradient RL in Algorithm~\ref{alg: online} where $L_{\rm mean}$ is the mean value of the loss (\ref{eq:loss function}) over one training episode with the duration $T_{\rm episode}$.

\begin{algorithm}
\caption{Model-based Policy Gradient RL}
\label{alg: online}
\SetKwInput{Input}{Input}
\SetKwInput{Initialization}{Initialization}
\SetKwInput{Forward}{Forward Pass}
\SetKwInput{Backward}{Backward Pass}
\Input{The quadrotor reference trajectory ${\bm x}^{q,\rm ref}$ and the learning rate $\varepsilon$} 
\Initialization{$\bm \varpi_0 $}
\While {$L_{\rm mean}$ not converged}{
\For{$t \leftarrow 0$ \KwTo $T_{\rm episode}$}{
\Forward{}
Compute the adaptive weightings $\bm \theta_t$ using $\bm \Theta_t =  \bm f_{\bm \varpi_t}$ and the sparse parameterization;\\
Solve Problem (\ref{eq:mhe}) to obtain $\hat{\mathbf{x}}$; \Comment{using any numerical optimization solver}
Compute the loss function $L\left( \hat {\mathbf{x}} \right)$ using Eq.(\ref{eq:loss function});\\
Compute $\bm u_t$ from the control law (\ref{eq:general controller});\\
Apply $\bm u_t$ to update the quadrotor state ${\bm x}_{t}^q$;\\
\Backward{}
Compute $\frac{{\partial {\hat {\mathbf{x}} }}}{{\partial {\bm \theta} }}$ using Algorithm~\ref{alg: analytical solution} given $\hat{\mathbf{x}}$ and $\bm \theta_t$;\\
Compute $ \frac{{\partial L\left( {\hat {\mathbf{x}}} \right)}}{{\partial \hat {\mathbf{x}}}}$ from the loss $L\left( {\hat {\mathbf{x}}} \right)$;\\
{Compute $\frac{\partial \boldsymbol{\theta}}{\partial \boldsymbol{\Theta}}$ from the sparse parameterization;}\\
Compute $\frac{{\partial {\boldsymbol{\Theta}}\left(\boldsymbol \varpi \right) }}{{\partial {\boldsymbol \varpi} }}$ of the NN; \Comment{using any machine learning tool}
Apply the chain rule {(\ref{eq:new chain rule})} to obtain $\frac{{d L\left( \hat {\mathbf{x}} \right)}}{{d {\boldsymbol \varpi} }}$;\\
Update $\bm \varpi_t$ using gradient-based optimization;
}
{Calculate $L_{\rm mean}$ for the next episode}
\Comment{ one training episode}
}
\end{algorithm}

\section{Constrained NeuroMHE} \label{section: constraints}
Problem (\ref{eq:mhe}), as in our earlier conference work~\cite{wang2021differentiable}, does not consider inequality constraints imposed on the system state and process noise. To address this limitation, we extend our method to the case where inequality constraints need to be respected in the MHE. This can enhance flight safety by preventing unrealistic estimations. For example, when a quadrotor is carrying an unknown payload, the estimated payload disturbance should not exceed the maximum collective force produced by the propellers. Without loss of generality, we consider the following constraints defined in one horizon, i.e., at the time step $k \in \left[ {t - N,t} \right]$:
\begin{equation}
    {g_{k,i}}\left( {{{{\bm x}}_{k}},{{\bm w} _k}} \right) \le 0,i = 1, \cdots ,n.
    \label{eq:general constraint}
\end{equation}
Although the inequalities (\ref{eq:general constraint}) can have a general form, they must satisfy the linear independence constraint qualification (LICQ) to ensure the KKT conditions hold at the optimal solutions\footnote{It is straightforward to show that the equality constraint (\ref{eq:mhe equality constraint}) satisfies the LICQ when the system is both controllable and observable.}. In other words, $\nabla g_{k,i},i=1,\cdots ,n$ are linearly independent. It is worth noting that this requirement is not overly conservative and can be satisfied by various common constraints in practice, such as box constraints.

By enforcing (\ref{eq:general constraint}) as hard constraints in the MHE optimization problem, we have:
\begin{subequations}
    \begin{align}
&\mathop {\min }\limits_{{{\bm {\mathop{ {\rm x}}\nolimits}},{{{\bm {\mathop{ {\rm w}}\nolimits}} }}}} J
\label{eq:cost of constrained mhe}\\
{\rm s.t.}&\ {{{\bm x}}_{k + 1}} = {\bm f}\left( {{{{\bm x}}_{k}},{\bm u_k},{{\bm w} _k},\Delta t} \right),
\label{eq:dynamics of constrained mhe}\\
&\ {g_{k,i}}\left( {{{{\bm x}}_{k}},{{\bm w} _k}} \right) \le 0, i = 1, \cdots ,n,
\label{eq:inequality constraint of constrained mhe}
\end{align}
\label{eq:constrained mhe}%
\end{subequations}
where $J$ is the same as defined in \eqref{eq:mhe}. We denote the optimal constrained estimates of (\ref{eq:constrained mhe}) by $\hat{\mathbf{x}}_c$ and $\hat{\mathbf{w}}_c$. Similar to (\ref{eq:mhe}), $\hat{\mathbf{x}}_c$ is parameterized as $\hat {\mathbf{x}}_{\rm c}\left( \boldsymbol \theta  \right)$ by the weighting parameters. This method, however, has several implementation difficulties for computing the gradients $\frac{{\partial \hat {\mathbf{x}}_c}}{{\partial {\boldsymbol \theta} }}$. When differentiating the KKT conditions of (\ref{eq:constrained mhe}) with respect to $\boldsymbol \theta$, one needs to identify all the active constraints ${g_{k,i}}\left( {{{{\bm x}}_{k}},{{\bm w} _k}} \right) = 0$, which can be numerically inefficient. Moreover, the discontinuous switch between inactive and active inequality constraints may incur numerical instability in learning. Instead of treating (\ref{eq:general constraint}) as hard constraints, we use interior-point methods to softly penalize (\ref{eq:general constraint}) in the cost function of the MHE optimization problem. In particular, using the logarithm barrier functions, we have
\begin{subequations}
\begin{align}
    \mathop {\min }\limits_{ {\bm {\mathop{ {\rm x}}\nolimits}}, {\bm {\mathop{ {\rm w}}\nolimits}} } & \ J - \delta \sum\limits_{k = t - N}^t {\sum\limits_{i = 1}^n {\ln \left( { - {g_{k,i}}\left( {{{{\bm x}}_{k}},{{\bm w} _k}} \right)} \right)} }
\label{eq:soft penalty cost function}\\
{\rm s.t.}&\ {{{\bm x}}_{k + 1}} = {\bm f}\left( {{{ {\bm x}}_{k}},{\bm u_k},{{\bm w} _k},\Delta t} \right),
\label{eq:dynamics soft constraint}
\end{align}
\label{eq:mhe with soft constraint}%
\end{subequations}
where $\delta $ is a positive barrier parameter. 

Compared with (\ref{eq:constrained mhe}), the optimal estimates $\hat{\bm {\mathop{ {\rm x}}\nolimits}}$ to the soft-constrained MHE problem (\ref{eq:mhe with soft constraint}) is now determined by both $\boldsymbol \theta$ and $\delta$, denoted by $\hat {\mathbf{x}}\left( {\bm \theta ,\delta } \right)$. If $\delta  \to 0$, then $\hat {\mathbf{x}}\left( {\boldsymbol \theta ,\delta } \right) \to \hat {\mathbf{x}}_{\rm c}\left( \boldsymbol \theta  \right)$ and $\frac{\partial \hat{\mathbf{x}}\left ( \boldsymbol \theta,\delta  \right )}{\partial \boldsymbol \theta}\rightarrow \frac{\partial \hat{\mathbf{x}}_c\left ( \boldsymbol \theta  \right )}{\partial \boldsymbol \theta}$, which is a well-known property of interior-point methods~\cite{forsgren2002interior}. Hence, by setting $\delta > 0$ to be sufficiently small, we can utilize $\frac{{\partial {{\hat {\mathbf{x}}}}\left( {\boldsymbol \theta ,\delta } \right)}}{{\partial \boldsymbol \theta }}$ to approximate $\frac{{\partial {{\hat {\mathbf{x}}}_c}\left( \boldsymbol \theta  \right)}}{{\partial \boldsymbol \theta }}$ with arbitrary accuracy. This allows for applying Algorithm~\ref{alg: analytical solution} to calculate $\frac{{\partial {{\hat {\mathbf{x}}}}\left( {\boldsymbol \theta ,\delta } \right)}}{{\partial \boldsymbol \theta }}$. Similarly, by approximating the adaptive $\boldsymbol \theta$ in (\ref{eq:mhe with soft constraint}) with the neural network (\ref{eq:nn parameterization}), we can train the soft-constrained NeuroMHE (\ref{eq:mhe with soft constraint}) using Algorithm~\ref{alg: online}.

\section{Experiments} \label{section: simulation}
We validate the effectiveness of NeuroMHE in robust flight control through both numerical and physical experiments on quadrotors. In particular, we will show the following advantages of NeuroMHE. First, it enjoys computationally efficient training and significantly improves the force estimation performance over a state-of-the-art estimator (See \ref{subsec:expA}). Second, a stable NeuroMHE with a fast dynamic response can be trained directly from the trajectory tracking error using Algorithm~\ref{alg: online} (See \ref{subsubsection:training setup}). Third, NeuroMHE exhibits superior estimation and robust control performance than a fixed-weighting MHE and a state-of-the-art adaptive controller for handling dynamic noise covariances (See \ref{subsubsection: evaluation unseen trajectory}). Finally, NeuroMHE is efficiently transferable to different challenging flight scenarios on a real quadrotor without extra parameter tuning, including counteracting state-dependent cable forces and flying under the downwash flow (See \ref{subsec:expC}).

\begin{figure}[h!]
	\centering
	{\includegraphics[width=0.6\linewidth]{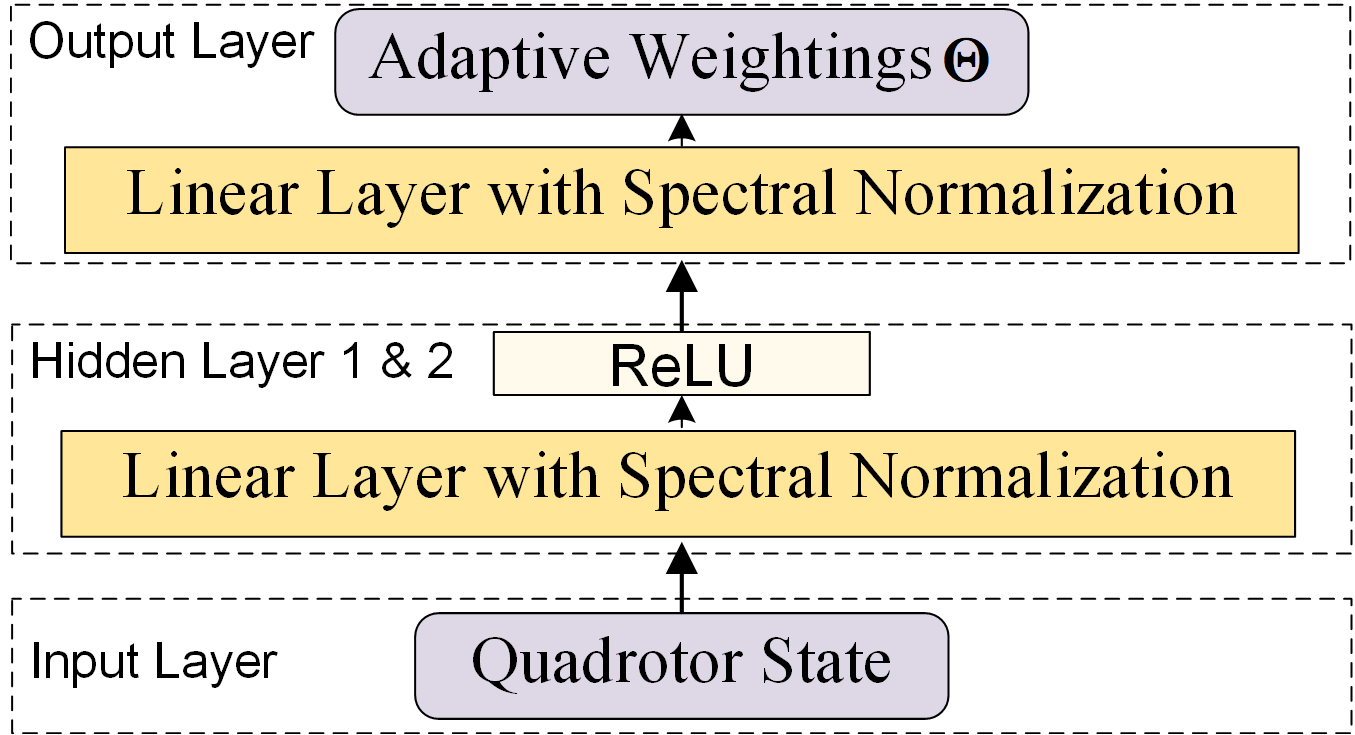}}
	\caption{\footnotesize Architecture of the neural network for generating the adaptive MHE weighting matrices online. }
	\label{fig:neural network structure}
\end{figure}

In our experiments, we design the neural network to take the current partial or full quadrotor state ${{\bm x}^q_t}$ as input, which we assume is available through onboard sensors or motion capture systems. Correspondingly, we set the measurement of the augmented system (\ref{eq:robotic model}) to be ${\bm y_t} = {{\bm x}^q_t} + {\bm \nu} $. A multi-layer perceptron (MLP) network is adopted, and its architecture is depicted in Fig.~\ref{fig:neural network structure}. 
The MLP has two hidden layers with the rectified linear unit (ReLU) $\phi\left( x \right) = \max \left( {x,0} \right)$ as the activation function. 
The linear layer undergoes spectral normalization, which involves rescaling the weight matrix with its spectral norm. This technique constrains the Lipschitz constant of the layer, thereby enhancing the network's robustness and generalizability~\cite{bartlett2017spectrally}. The adopted MLP is therefore expressed mathematically as
\begin{equation}
  {\bm f_{\bm \varpi} }\left( {{{\bm y}_t}} \right) = {{\bm A}_{o}}\phi \left( {{{\bm A}_{2}}\phi \left( {{{\bm A}_{1}}{{\bm y}_t} + {{\bm b}_{1}}} \right) + {{\bm b}_{2}}} \right) + {{\bm b}_o},
    \label{eq:neural network expression}
\end{equation}
where ${\bm A_o}\in \mathbb{R}^{D_{\rm out}\times D_{\rm h}}$, ${\bm A_{{1}}}\in \mathbb{R}^{D_{\rm h}\times D_{\rm in}}$, and ${\bm A_{{2}}}\in \mathbb{R}^{D_{\rm h}\times D_{\rm h}}$ are the weight matrices, ${\bm b_o}\in \mathbb{R}^{D_{\rm out}}$, ${\bm b_{{1}}}\in \mathbb{R}^{D_{\rm h}}$, and ${\bm b_{{2}}} \in \mathbb{R}^{D_{\rm h}}$ are the bias vectors, they are the neural network parameters ${\bm \varpi}$ to be learned, $D_{\rm in}$, $D_{\rm h}$, and $D_{\rm out}$ denote the numbers of the neurons in the input, hidden, and output layers, respectively.

We implement our algorithm in Python and use \texttt{ipopt} with CasADi~\cite{andersson2019casadi} to solve the nonlinear MHE optimization problem (\ref{eq:mhe}). The MLP~(\ref{eq:neural network expression}) is built using PyTorch~\cite{paszke2019pytorch} and trained using \texttt{Adam}~\cite{kingma2014adam}. In the implementation, we customize the loss function (\ref{eq:loss function}) to suit the typical training procedure in PyTorch. Specifically, the customized loss function for training the neural network (\ref{eq:neural network expression}) is defined by $L_{\rm{pytorch}}=\left.\begin{matrix}
\frac{dL}{d\boldsymbol\Theta}
\end{matrix}\right|_{\boldsymbol\Theta_{t}}{\boldsymbol\Theta}$ where $\left.\begin{matrix}
\frac{dL}{d\boldsymbol\Theta}
\end{matrix}\right|_{\boldsymbol\Theta_{t}}=\left.\begin{matrix}
\frac{\partial L}{\partial \hat{\mathbf{x}}}
\end{matrix}\right|_{\hat{\mathbf{x}}_{t}}\left.\begin{matrix}
\frac{\partial \hat{\mathbf{x}}}{\partial \boldsymbol\theta}
\end{matrix}\right|_{\boldsymbol\theta_{t}}\left.\begin{matrix}
\frac{\partial \boldsymbol\theta}{\partial \boldsymbol\Theta}
\end{matrix}\right|_{\boldsymbol\Theta_{t}}$ is the gradient of the loss (\ref{eq:loss function}) with respect to $\boldsymbol\Theta$ evaluated at $\boldsymbol\Theta_{t}$, such that $\frac{{d{L_{{\rm{pytorch}}}}}}{{d\bm \varpi }} = \frac{{dL}}{{d\bm \varpi }}$.

In training with gradient descent, overlarge gradients can lead to unstable training. This could occur when the lower bounds $\varsigma$ and $\gamma_{\min}$ of the parameterized weightings are very small, since the calculation of $\frac{\partial \hat{\mathbf{x}}}{\partial \boldsymbol\theta}$ requires the inverse of the weightings. Thus, we additionally set a threshold $\rho $ for the 2-norm of the gradient $\left \| \frac{dL}{d\boldsymbol\Theta} \right \|_{2}$ to improve training. Whenever $\left \| \frac{dL}{d\boldsymbol\Theta} \right \|_{2}\geqslant \rho $, we bound the gradient $\frac{dL}{d\boldsymbol\Theta}$ via multiplying it by $\rho {\left \| \frac{dL}{d\boldsymbol\Theta} \right \|^{-1}_{2}}$ and increase the lower bounds slightly.

\subsection{Efficient Learning for Accurate Estimation}
\label{subsec:expA}

In this numerical experiment, we compare NeuroMHE with the state-of-the-art estimator NeuroBEM~\cite{bauersfeld2021neurobem} 
using the same flight dataset presented in~\cite{bauersfeld2021neurobem}, which was collected in the world's largest indoor motion capture room from various agile and extreme flights. The ground-truth data (quadrotor states and disturbance) can be obtained from the dataset and the quadrotor dynamics model. To be consistent with NeuroBEM, we train NeuroMHE from the estimation error using supervised learning. Specifically, we use Algorithm~\ref{alg: analytical solution} to compute the gradients $\frac{{\partial {\hat {\mathbf{x}} }}}{{\partial {\bm \theta} }}$, build the loss function $L$ using the estimation error, and apply gradient descent to update the neural network parameters ${\bm \varpi}$. The ground-truth disturbance forces and torques $\bm {d} = \left [ {\bm F};{\boldsymbol {\tau} } \right ]$ are computed by
\begin{equation}
    {\bm F} = m\left( {{{\bm a}_{\bm v}} + g{\bm e}_3 } \right),\ {\boldsymbol {\tau} } = {\bm J}{{\bm a}_{\boldsymbol \omega} } + {{\boldsymbol \omega} ^ \times }{\bm J}{\boldsymbol \omega},
\label{eq:ground truth}
\end{equation}
where ${\bm{J}} = {\rm{diag}}\left( {2.5,2.1,4.3} \right) \times {\rm{1}}{{\rm{0}}^{ - 3}}\ {\rm{kg}} {{\rm{m}}^{\rm{2}}}$ and $m = 0.772\ {\rm kg}$ as used in~\cite{bauersfeld2021neurobem}\footnote{The quadrotor's mass was initially reported as $0.752\ {\rm kg}$ in \cite{bauersfeld2021neurobem}. However, a subsequent update on the NeuroBEM website (\url{https://rpg.ifi.uzh.ch/neuro_bem/Readme.html}) revised this to $0.772\ {\rm kg}$. We have confirmed with the authors that the revised mass was used in both the training of NeuroBEM and the computation of RMSE values.}, ${\bm a}_{\bm v}$ and ${\bm a}_{\bm \omega}$ are the ground-truth linear and angular accelerations given in the dataset. 
Here $\bm F$ and $\boldsymbol{\tau}$ include the control forces and torques generated by the quadrotor's motors, following the convention in~\cite{bauersfeld2021neurobem}. 

As introduced in~\ref{sec:intro}, NeuroBEM fuses first principles with a DNN that models the residual aerodynamics caused by  interactions between the propellers. The DNN takes as inputs both the linear and angular velocities, along with the motor speeds. In practice, measuring the motor speeds typically requires specialized sensors and autopilot firmware. In comparison, we only need the linear and angular velocities (i.e., a subset of the quadrotor state-space) as inputs to our network. They are also the only measurements utilized by MHE. Moreover, we use a reduced quadrotor model in MHE, which contains only the velocity dynamics. We augment these velocities with the unmeasurable disturbances to define the augmented state ${\bm x}_{\rm r} = \left [ {\bm v};{\bm F};{\boldsymbol {\omega }};{\boldsymbol{\tau}} \right ]$, and the corresponding dynamics model with the random walks (\ref{eq: random walk}) for predicting the state in MHE is given by
\begin{subequations}
\begin{align}
        \dot {\bm v} & = {m^{ - 1}}\left( -mg {\bm e}_3  + {\bm F} \right),& {\dot {\bm F}} & = {\bm w _f}
        \label{eq:position reduced dynamics},\\
        \dot {\boldsymbol \omega}  & = {{\bm J}^{ - 1}}\left(  - {\bm \omega ^ \times } {\bm J}{\boldsymbol \omega}  + {{\boldsymbol \tau}} \right),& {\dot {\boldsymbol{\tau}}} & = {\bm w _{\tau}}.
        \label{eq:attitude reduced dynamics}
\end{align}
\label{eq:quadrotor reduced model}%
\end{subequations}
The above configuration results in the weightings of NeuroMHE having the following dimensions: ${\bm P}\in\mathbb{R}^{12\times 12}$, ${\bm R}_k\in\mathbb{R}^{6\times 6}$, and ${\bm Q}_k\in\mathbb{R}^{6\times 6}$. The weightings are subject to the sparse parameterization, which is elaborated in~\ref{subsec:parameterization}. Additionally, the first diagonal element in ${\bm R}_t$ is fixed at $100$. Then, we train the MLP (\ref{eq:neural network expression}) to tune the remaining $23$ diagonal elements and the two forgetting factors, which are collected in the vector $\boldsymbol{\Theta}=\left [ p_{1:12}, \bar{\gamma}_1,r_{1:5},\bar{\gamma}_2,q_{1:6}\right ]$. Fixing the numerical scale helps improve the training efficiency and optimality, as it reduces the number of local optima for the gradient descent. For the MLP (\ref{eq:neural network expression}), we set $D_{\rm in}=6$, $D_{\rm h}=30$, and $D_{\rm out}=25$, leading to total $1915$ network parameters. The dynamics model (\ref{eq:quadrotor reduced model}) is discretized using the 4th-order Runge-Kutta method with the same time step of $2.5\ {\rm ms}$ as in the dataset. 


\begin{figure*}[h!]
\centering
\begin{subfigure}[b]{0.49\textwidth}
\centering
\includegraphics[width=0.875\textwidth]{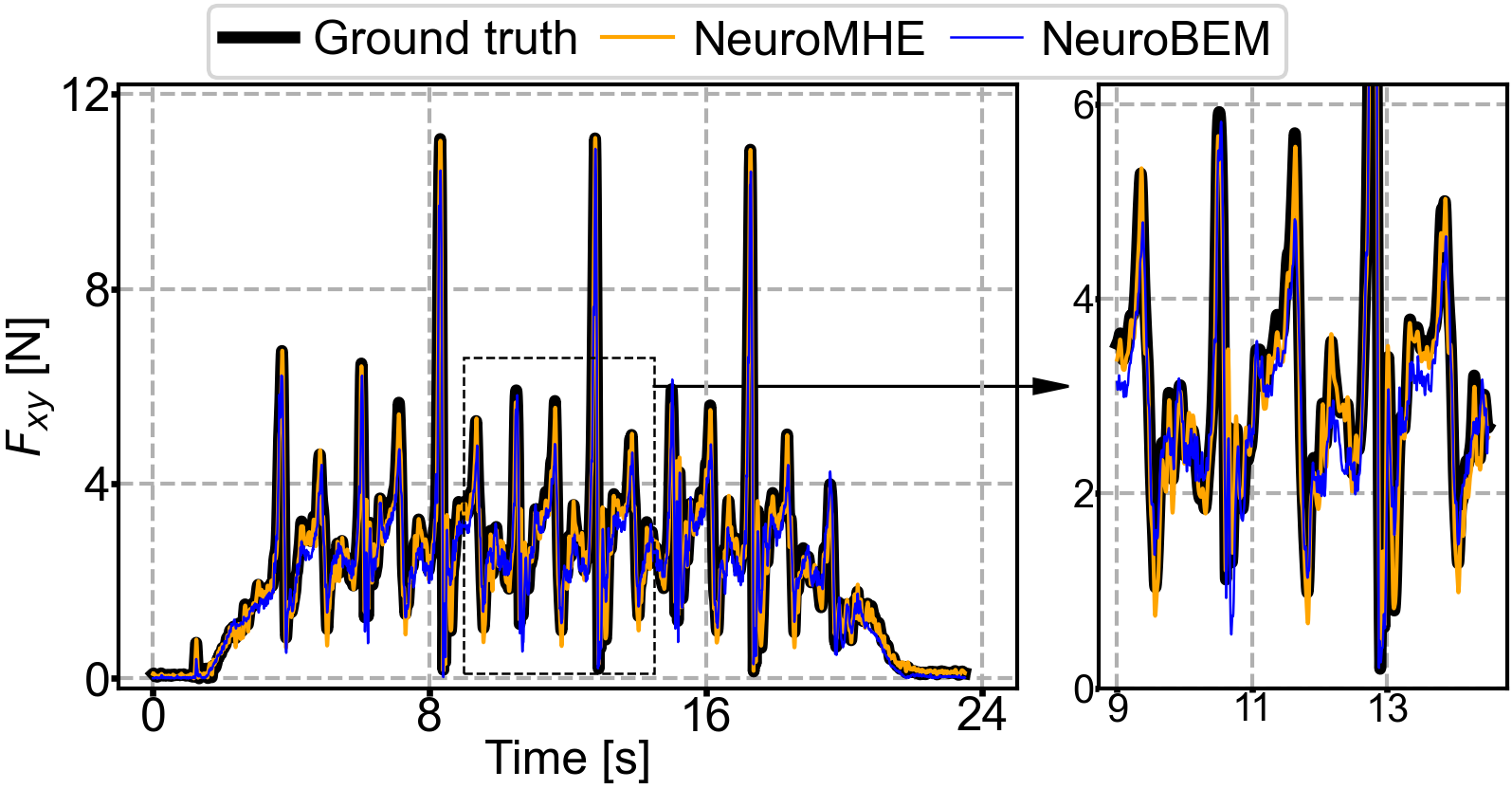}
\caption{\footnotesize Disturbance force in $xy$ plane.}
\label{fig:force xy real}
\end{subfigure}
\hfill
\begin{subfigure}[b]{0.49\textwidth}
\centering
\includegraphics[width=0.875\textwidth]{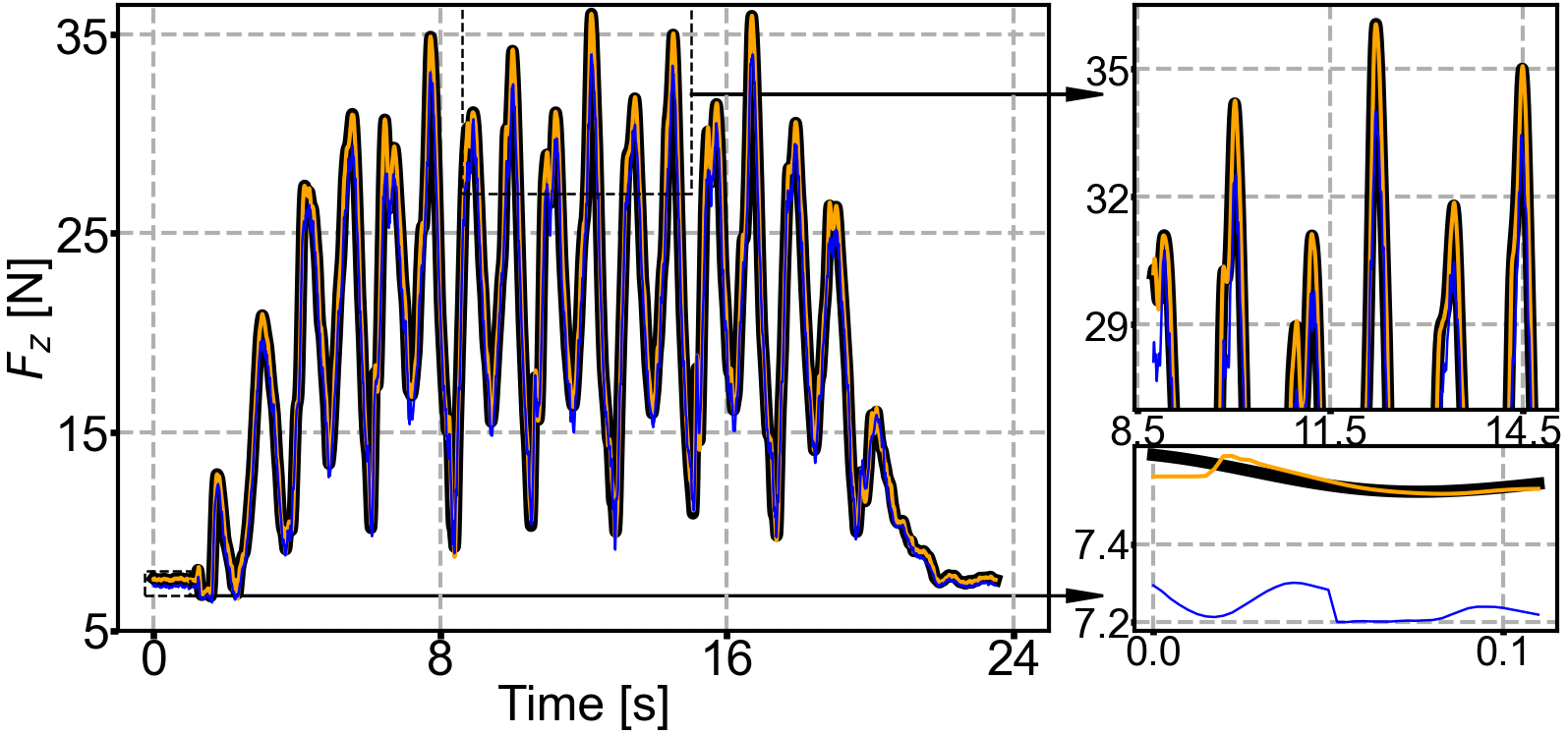}
\caption{\footnotesize Disturbance force in $z$ axis.}
\label{fig:force z real}
\end{subfigure}
\vskip\baselineskip
\begin{subfigure}[b]{0.49\textwidth}
\centering
\includegraphics[width=0.875\textwidth]{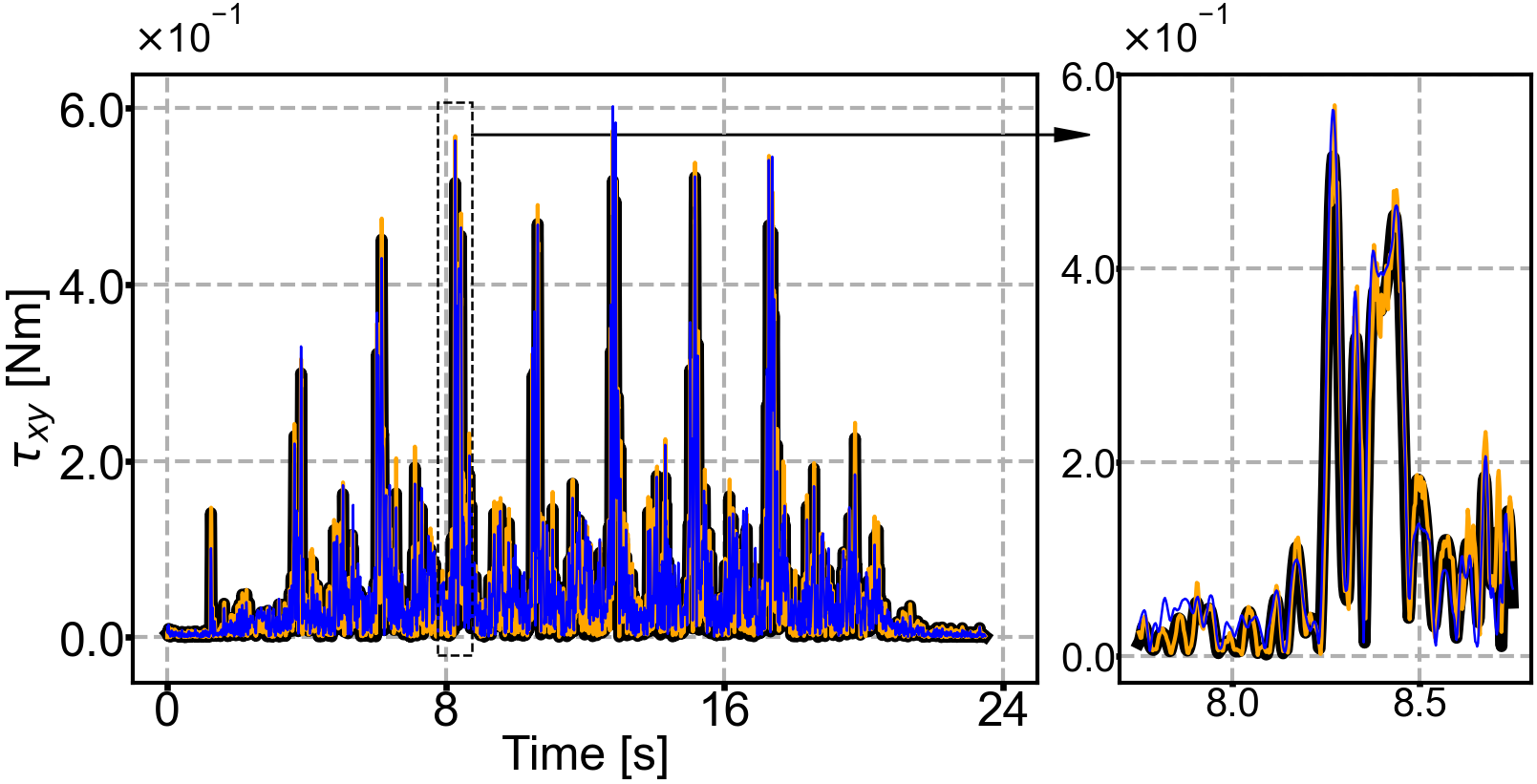}
\caption{\footnotesize Disturbance torque in $xy$ plane.}
\label{fig:torque xy real}
\end{subfigure}
\hfill
\begin{subfigure}[b]{0.49\textwidth}
\centering
\includegraphics[width=0.875\textwidth]{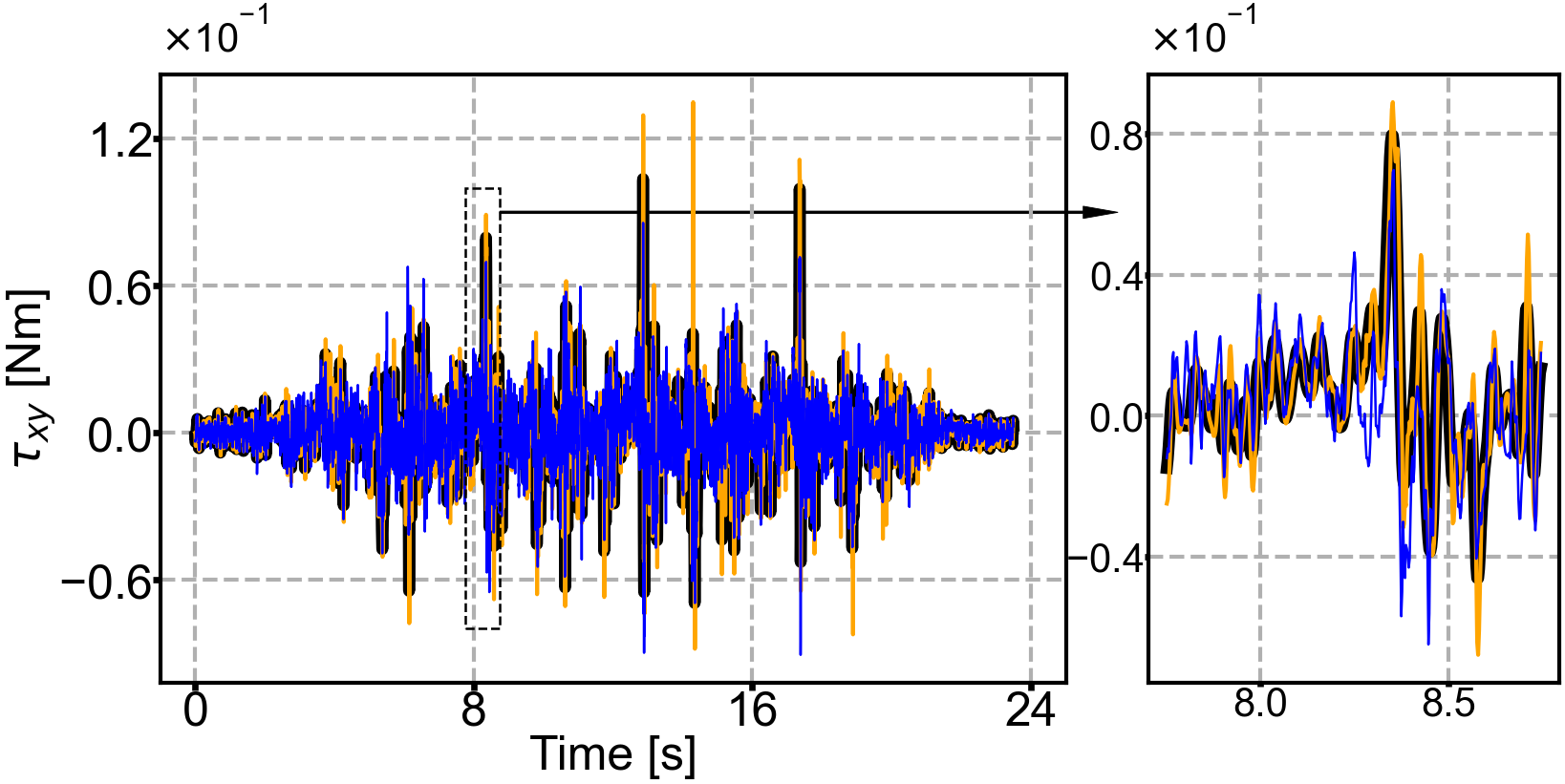}
\caption{\footnotesize Disturbance torque in $z$ axis.}
\label{fig:torque z real}
\end{subfigure}
\caption{\footnotesize The force and torque estimation performance of NeuroMHE on a highly aggressive {Figure-8} flight test dataset as used in~\cite{bauersfeld2021neurobem}. {Given that the NeuroBEM's estimated force is expressed in the body frame in the dataset, we transform our NeuroMHE-estimated force into the same frame to facilitate the comparison.} At $t=0$, the initial guess of the disturbance used in the arrival cost is set to ${\hat {\bm d}_0} = \left[ {0;0;mg;0;0;0} \right]$ since the quadrotor is nearly in hover at the beginning. Fig.~\ref{fig:force xy real} and \ref{fig:torque xy real} are obtained by plotting ${{F_{xy}}} = \sqrt {F_x^2 +  F_y^2}$ and ${{{{\tau} _{xy}}}} = \sqrt {{\tau} _x^2 + {\tau }_y^2} $, respectively. }
\label{fig: estimation real data}
\end{figure*}

For the training data, we choose a $10$-second-long trajectory segment from a { wobbly circle} flight, which { covers a limited velocity range} of $0.19\ {\rm {m} \mathord{\left/
 {\vphantom {m s}} \right.
 \kern-\nulldelimiterspace} \rm s}$ to $5.18\ {\rm {m} \mathord{\left/
 {\vphantom {m s}} \right.
 \kern-\nulldelimiterspace} \rm s} $. One training episode amounts to training the NeuroMHE over this $10$-second-long {slow dataset} once. For the test data, we use the same dataset as in~\cite{bauersfeld2021neurobem} to compare the performance of NeuroMHE with NeuroBEM over $13$ unseen agile flight trajectories. The corresponding disturbance estimation results of NeuroBEM are provided in the dataset. The parameters of these trajectories are summarized at \url{https://github.com/RCL-NUS/NeuroMHE}, where a visual comparison of the velocity-range space between the training and the test datasets is provided.

Before showing the estimation performance of NeuroMHE, we first test its computational efficiency in training by measuring the CPU runtime of Algorithm~\ref{alg: analytical solution} for different MHE horizons. As summarized in Table~\ref{table:cpu time}, the runtime is approximately linear to the horizon, thus the algorithm can scale efficiently to a large MHE problem with a long horizon. This is made possible because of the proposed Kalman filter-based gradient solver (Algorithm~\ref{alg: analytical solution}), which has a linear computational complexity with respect to the number of horizon $N$. Given the quadrotor's fast dynamics in extreme flights, we set $N=10$ in training.

\begin{table}[h]
\fontsize{7.5}{7.5}\selectfont
\caption{Runtime of Algorithm~\ref{alg: analytical solution} for Different MHE Horizons\label{table:cpu time}}
\centering
\begin{threeparttable}[t]
\begin{tabular}{ c|c c c c c c} 
\toprule[1pt]
Horizon $N$ & $10$ & $20$ & $40$ & $60$ & $80$ & $100$\\
\midrule[0.5pt]
Runtime $\left[ {\rm ms} \right]$ & $1.83$ & $3.74$ & $6.97$ & $12.36$ & $14.93$ & $18.71$\\
\bottomrule[0.5pt]
\end{tabular}
\begin{tablenotes}[flushleft]
      \footnotesize
      \item The time is measured in a workstation with a $11$th Gen Intel Core i7-11700K processor.
\end{tablenotes}
\end{threeparttable}
\end{table}

Fig.~\ref{fig: estimation real data} shows a comparison between the ground truth disturbances and the estimation performance of NeuroMHE and NeuroBEM on a highly aggressive flight dataset (See '{Figure-8\_4}' in Table~\ref{table:testset rmse}). The force estimates of NeuroMHE can converge quickly to the ground truth data from the initial guess within about $0.1\ {\rm s}$ (See the zoom-in figure in Fig.\ref{fig:force z real}). The estimation is of high accuracy even around the most fast-changing force spikes (See Fig.\ref{fig:force xy real} and \ref{fig:force z real}) where NeuroBEM, however, demonstrates sizeable estimation performance degradation. The torque estimates of NeuroMHE also show satisfactory performance but are noisier than its force estimates. To compare with NeuroBEM quantitatively, we utilize the Root-Mean-Square errors (RMSEs) to quantify the estimation performance and summarize the comparison in Table~\ref{table:rmse}. As can be seen, the estimation errors of force and torque in NeuroMHE are all much smaller than those in NeuroBEM. In particular, it outperforms NeuroBEM substantially in the planar and the vertical forces estimation with the estimation error reductions of $52.4\%$ and $78.1\%$, respectively.
\begin{table}[h]
\fontsize{7.5}{7.5}\selectfont
\caption{Estimation Errors (RMSEs) Comparisons\label{table:rmse}}
\centering
\begin{threeparttable}[t]
\begin{tabular}{ c|c c c c|c c } 
\toprule[1pt]
\multirow{2}{*}{Method}  & ${F_{xy}}$ & ${F_z}$ & ${\tau_{xy}}$ & ${\tau_{z}}$ & $F$ & ${\tau}$\\
       & $\left[ {\rm N} \right]$ & $\left[ {\rm N} \right]$ & $\left[ {\rm Nm} \right]$ & $\left[ {\rm Nm} \right]$ & $\left[ {\rm N} \right]$ & $\left[ {\rm Nm} \right]$\\
\midrule[0.5pt]
{ NeuroBEM}  & $0.508$ &  $1.084$ & $0.027$ & $0.009$ & $1.197$ & $0.028$\\
{ NeuroMHE}  & ${\bf 0.242}$ & ${\bf 0.237}$ &${\bf 0.017}$ & ${\bf 0.007}$ & ${\bf 0.339}$ & ${\bf 0.018}$\\
\bottomrule[0.5pt]
\end{tabular}

\end{threeparttable}
\end{table}

For a comprehensive comparison, we further evaluate the performance of the trained NeuroMHE on the entire NeuroBEM test dataset. The results are delineated in Table~\ref{table:testset rmse}. Notably, NeuroMHE demonstrates a significantly smaller RMSE in the overall force estimation than NeuroBEM across all of these trajectories{, achieving a reduction of up to $76.7\%$ (See the penultimate column). The only exception is the '3D Circle\_1' trajectory where both methods exhibit a similar RMSE value.} Furthermore, NeuroMHE exhibits a comparable performance in the overall torque estimation to that of NeuroBEM. The torque estimation performance could potentially be improved by using inertia-normalized quadrotor dynamics, wherein the force and torque magnitudes are similar. These findings underscore the superior generalizability of NeuroMHE to previously unseen challenging trajectories.

Thanks to the fusion of a portable neural network into the model-based MHE estimator, our auto-tuning/training algorithm achieves high data efficiency. For instance, NeuroBEM requires $3150$-second-long flight data to train a high-capacity neural network consisting of temporal-convolutional layers with $25{\rm k}$ parameters. In contrast, our method uses only $10$-second-long flight data to train a portable neural network that has only $7.7\%$ of the parameters of NeuroBEM ($1.915{\rm k}$), while improving the overall force estimation by up to $76.7\%$.

\subsection{Online Learning with Fast Adaptation}
\label{subsec:expB}
\subsubsection{Training setup}\label{subsubsection:training setup} we design a robust trajectory tracking control scenario in simulation to show the second advantage of our algorithm. We synthesize the external disturbances by integrating the random walk model (\ref{eq: random walk}) where ${\bm w _f} \sim {\cal N}\left( {{\mathbf 0},{\boldsymbol \sigma^{2}_f}} \right)$ and ${\bm w _\tau } \sim {\cal N}\left( {\mathbf 0,{\boldsymbol \sigma^{2}_\tau }} \right)$ are set to Gaussian noises. The standard deviations are modeled by two polynomials of the quadrotor state, i.e., ${\boldsymbol \sigma_f} = {\bm c_v}{\rm{dia}}{{\rm{g}}^2}\left(\bm v \right) + {\bm c_p}{\rm{dia}}{{\rm{g}}^2}\left( \bm p \right) + {\bm c_f}$ and ${\boldsymbol \sigma_\tau } = {\bm c_\omega }{\rm{dia}}{{\rm{g}}^2}\left( \bm \omega  \right) + {\bm c_\Theta }{\rm{dia}}{{\rm{g}}^2}\left( \bm \Theta_e  \right) + {\bm c_\tau }$ where $\bm \Theta_e$ denotes the Euler angles of the quadrotor, $\bm c_v$, $\bm c_p$, $\bm c_f$, $\bm c_{\omega}$, $\bm c_{\Theta}$, and $\bm c_{\tau}$ are $3\times 3$ diagonal positive definite coefficient matrices. The diagonal elements of $\bm c_v$, $\bm c_p$, and $\bm c_f$ are set to $2$, $1$, and $1$, respectively. The coefficients in $\bm \sigma_{\tau}$ are set to $0.1$, which is consistent with the relatively small aerodynamic torques in common flights. We generate a disturbance dataset offline for training by setting $\bm v$, $\bm p$, $\bm \Theta_e$, and $\bm \omega$ in $\bm \sigma_f$ and $\bm \sigma_{\tau}$ to be the desired quadrotor states on the reference trajectory. 

To highlight the benefits of the adaptive weightings, we conduct ablation studies by comparing NeuroMHE against fixed-weighting MHE. The latter is automatically tuned using gradient descent under the same conditions and is referred to as differentiable moving horizon estimator (DMHE)~\cite{wang2021differentiable}. Specifically, to train DMHE, we first use Algorithm~\ref{alg: analytical solution} to compute the gradients $\frac{{\partial {\hat {\bm {\mathop{ {\rm x}}\nolimits}} }}}{{\partial {\boldsymbol \theta} }}$, then obtain the gradient of the loss function with respect to $\bm \Theta$ using the chain rule $\frac{dL\left ( {\hat {\mathbf x}\left (\boldsymbol\Theta  \right )} \right )}{d \boldsymbol\Theta }=\frac{\partial L\left (\hat {\mathbf x}  \right )}{\partial \hat {\mathbf x}}\frac{\partial \hat {\mathbf x}\left ( \boldsymbol\theta \right )}{\partial \boldsymbol\theta}\frac{\partial \boldsymbol\theta}{\partial \boldsymbol\Theta}$ with the sparse parameterization defined in Sec.~\ref{subsec:parameterization}, and finally apply gradient descent to update $\boldsymbol \Theta$. Different from the adaptive weightings of NeuroMHE, the weightings of DMHE are fixed at their optimal values $\boldsymbol \Theta^*$ after training.

We develop the robust flight controller~\eqref{eq:general controller} based on a geometric controller proposed in~\cite{lee2010geometric}, which is treated as the baseline method in this simulation. The control objective is to have the quadrotor track the desired trajectory ${\bm p}_d\left ( t \right )$ and yaw angle $\psi _{d}\left ( t \right )$ for time $t$ under the external disturbances defined above. We generate ${\bm p}_d\left ( t \right )$ using the minimum snap method~\cite{mellinger2011minimum} and set $\psi _{d}\left ( t \right )$ to $0\ {\rm rad}$. The robust flight controller incorporates the current disturbance estimates $\hat{\bm d}_{f}$ and $\hat{\bm d}_{\tau }$ from NeuroMHE as feedforward terms in the baseline controller to compensate for the disturbances. Mathematically, the robust flight control force and torque are given by
\begin{subequations}
\begin{align}
    {\bm F}_{d}&=\bar{{\bm F}}_{d}-\hat{\bm d}_{f}
    \label{eq:desired active control force},\\
    f&={\bm F}_{d}\cdot \left ( {\bm R}{\bm e}_3 \right )
    \label{eq:projection of active force},\\
    {\boldsymbol\tau} _{m}&=\bar{\boldsymbol\tau} _{m}-\hat{\bm d}_{\tau }
    \label{eq:desired control torque},
\end{align}
\label{eq:active control}%
\end{subequations}
where $\bar{\bm F}_{d}$ is the nominal control force, ${\bm F}_{d}$ is the robust control force, $f$ is the desired collective thrust obtained by projecting ${\bm F}_{d}$ onto the current body-$z$ axis, $\bar{\boldsymbol\tau} _{m}$ is the nominal control torque, and ${\boldsymbol\tau} _{m}$ is the robust control torque. We denote ${\bm u}_{\rm b}=\left [ \bar{f};\bar{\boldsymbol\tau }_{m} \right ]$ as the baseline geometric control input where $\bar{f}=\bar{\bm F}_{d}\cdot \left ( {\bm R}{\bm e}_{3} \right )$, and ${\bm u}_{\rm r}=\left [ f;{\boldsymbol \tau}_{m}  \right ]$ as the robust geometric control input. The detailed design procedures of the geometric controller, specifically $\bar{\bm F}_{d}$ and $\bar{\boldsymbol\tau} _{m}$, are provided in Appendix-\ref{appendix:geometric}.

In the simulation environment, the quadrotor model (\ref{eq:quadrotor model}) is integrated using $4$th-order Runge-kutta method with a time step of $10\ {\rm ms}$ to update the measurement $\bm y$. We set the measurement noise $\boldsymbol \nu$ to Gaussian with a fixed covariance matrix much smaller than those of $\bm w_f$ and $\bm w_{\tau}$. The same numerical integration is used in NeuroMHE for state prediction. As explained earlier in this subsection, the noise covariances are defined as polynomials of the complete quadrotor states (i.e., $\bm p$, $\bm v$, $\boldsymbol{\Theta}_e$, and $\bm \omega$). Therefore, these states are selected as inputs for the neural network (\ref{eq:neural network expression}). Consistent with~\ref{subsec:expA}, we also fix the first diagonal element in ${\bm R}_t$ at $100$. Then, we train the network to tune the remaining $47$ diagonal elements and the two forgetting factors. Finally, the number of neurons in both hidden layers is $50$, and the MHE horizon is $10$. The simulations are done with the same workstation as in~\ref{subsec:expA}.

\begin{figure}[h]
	\centering
	{\includegraphics[width=0.65\linewidth]{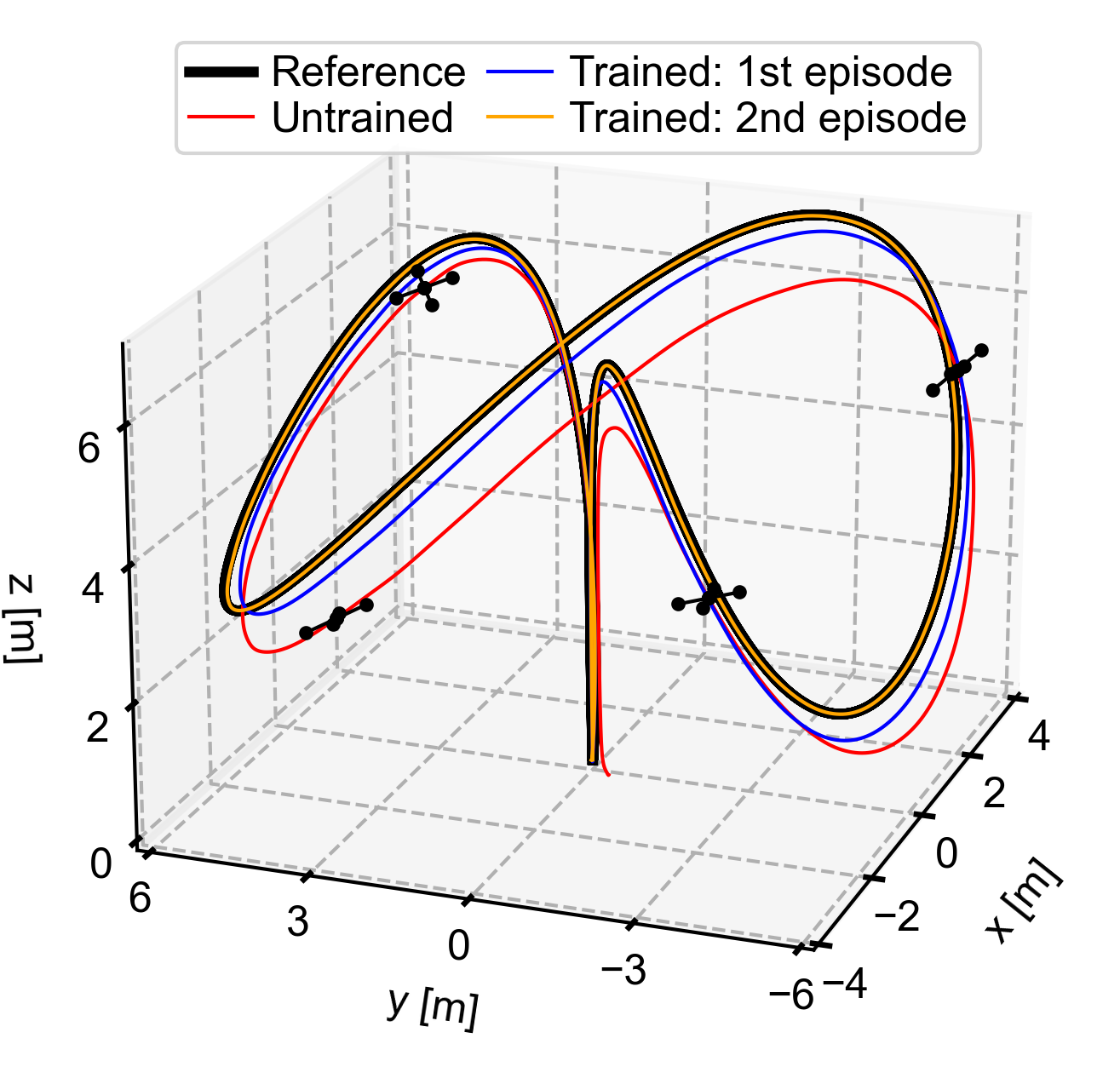}}
	\caption{\footnotesize Trajectory-tracking performance of NeuroMHE in training. One training episode amounts to training the NeuroMHE on the trajectory once. A video showing this training process can be found on \url{https://github.com/RCL-NUS/NeuroMHE}.}
	\label{fig:trajectory online}
\end{figure}

\begin{figure}[h]
\centering
\begin{subfigure}[b]{0.24\textwidth}
\centering
\includegraphics[width=1\textwidth]{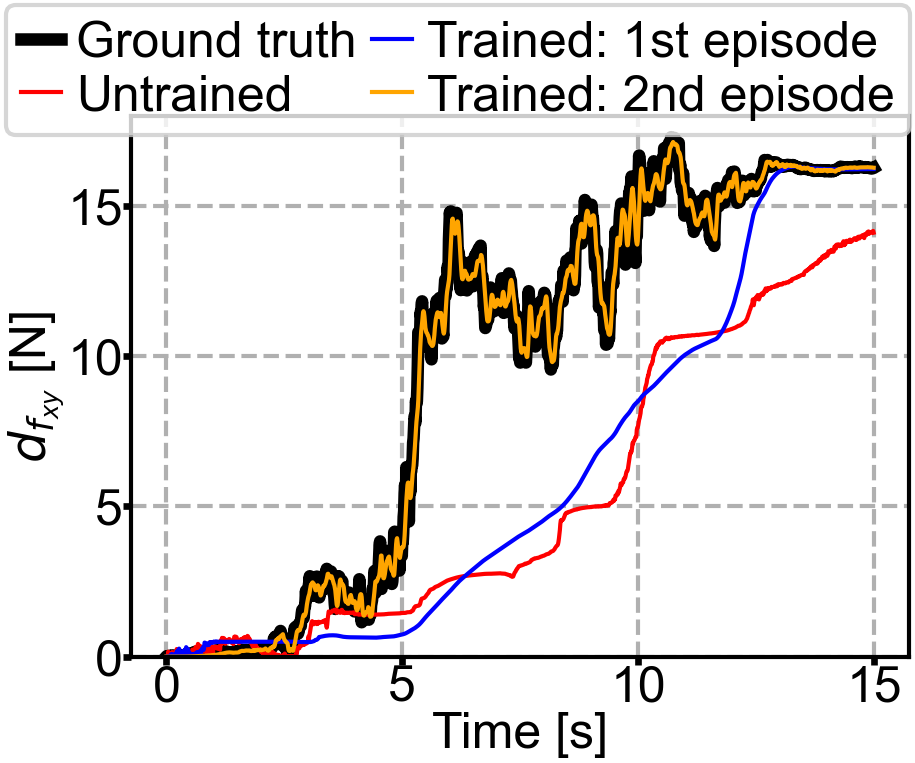}
\caption{\footnotesize Estimation of $d_{f_{xy}}$.}
\label{fig:fxy training}
\end{subfigure}
\hfill
\begin{subfigure}[b]{0.24\textwidth}
\centering
\includegraphics[width=1\textwidth]{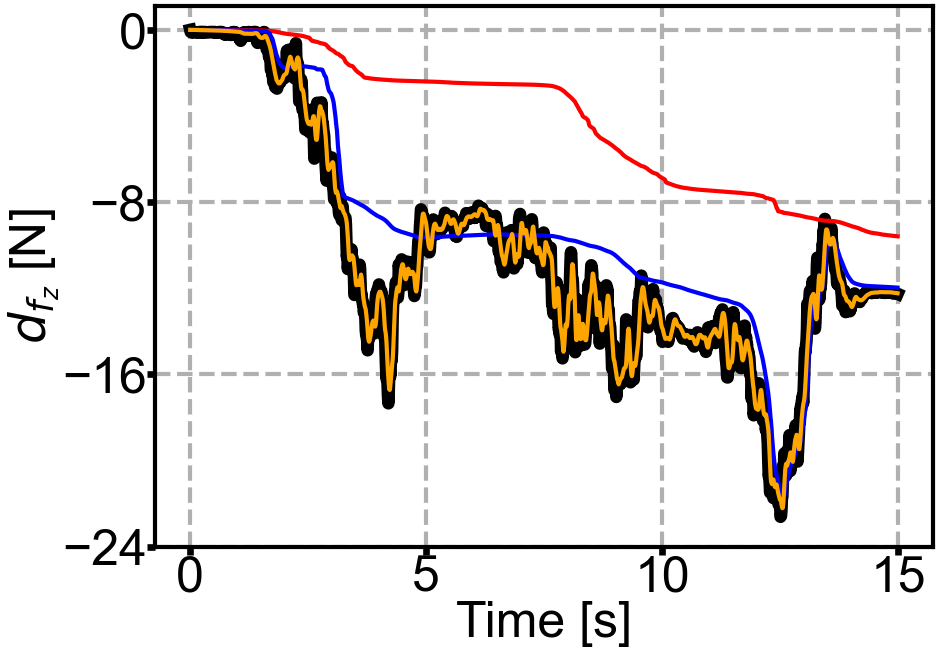}
\caption{\footnotesize Estimation of $d_{f_{z}}$.}
\label{fig:fz training}
\end{subfigure}
\hfill
\begin{subfigure}[b]{0.24\textwidth}
\centering
\includegraphics[width=1\textwidth]{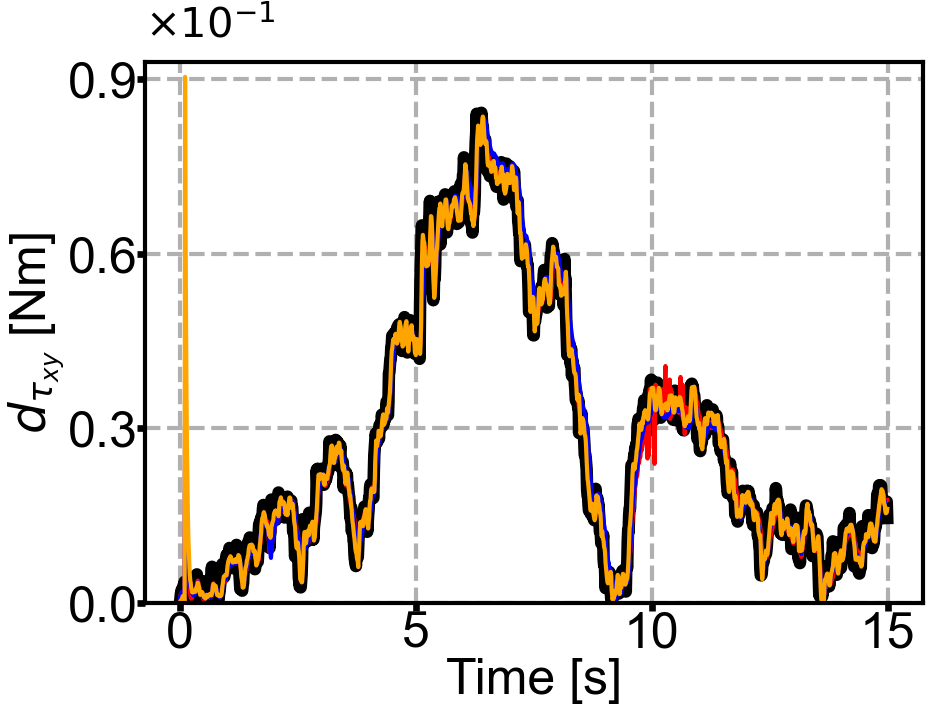}
\caption{\footnotesize Estimation of $d_{\tau_{xy}}$.}
\label{fig:torque xy training}
\end{subfigure}
\hfill
\begin{subfigure}[b]{0.24\textwidth}
\centering
\includegraphics[width=1\textwidth]{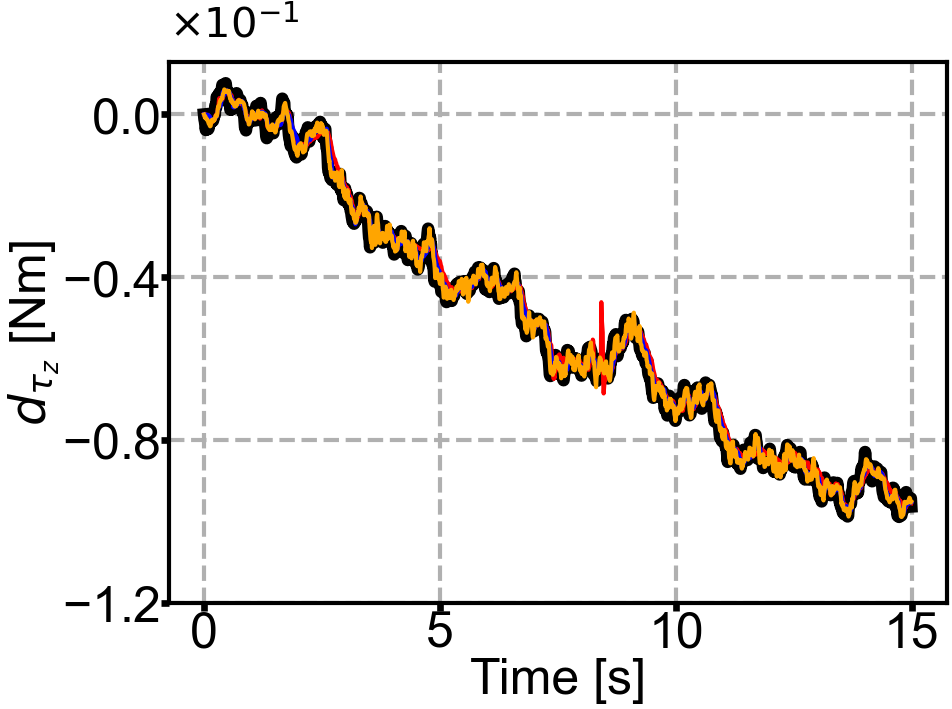}
\caption{\footnotesize Estimation of $d_{\tau_{z}}$.}
\label{fig:torque z training}
\end{subfigure}
\caption{\footnotesize Comparison of the estimation of disturbance force and torque in different training episodes. Fig.\ref{fig:fxy training} and Fig.\ref{fig:torque xy training} are obtained by plotting $d_{f_{xy}}=\sqrt{d_{f_{x}}^{2}+d_{f_{y}}^{2}}$ and $d_{\tau _{xy}}=\sqrt{d_{\tau_x}^{2}+d_{\tau_y}^{2}}$, respectively.}
\label{fig: force and torque in training}
\end{figure}

\begin{figure}[h]
\centering
\begin{subfigure}[b]{0.24\textwidth}
\centering
\includegraphics[width=1\textwidth]{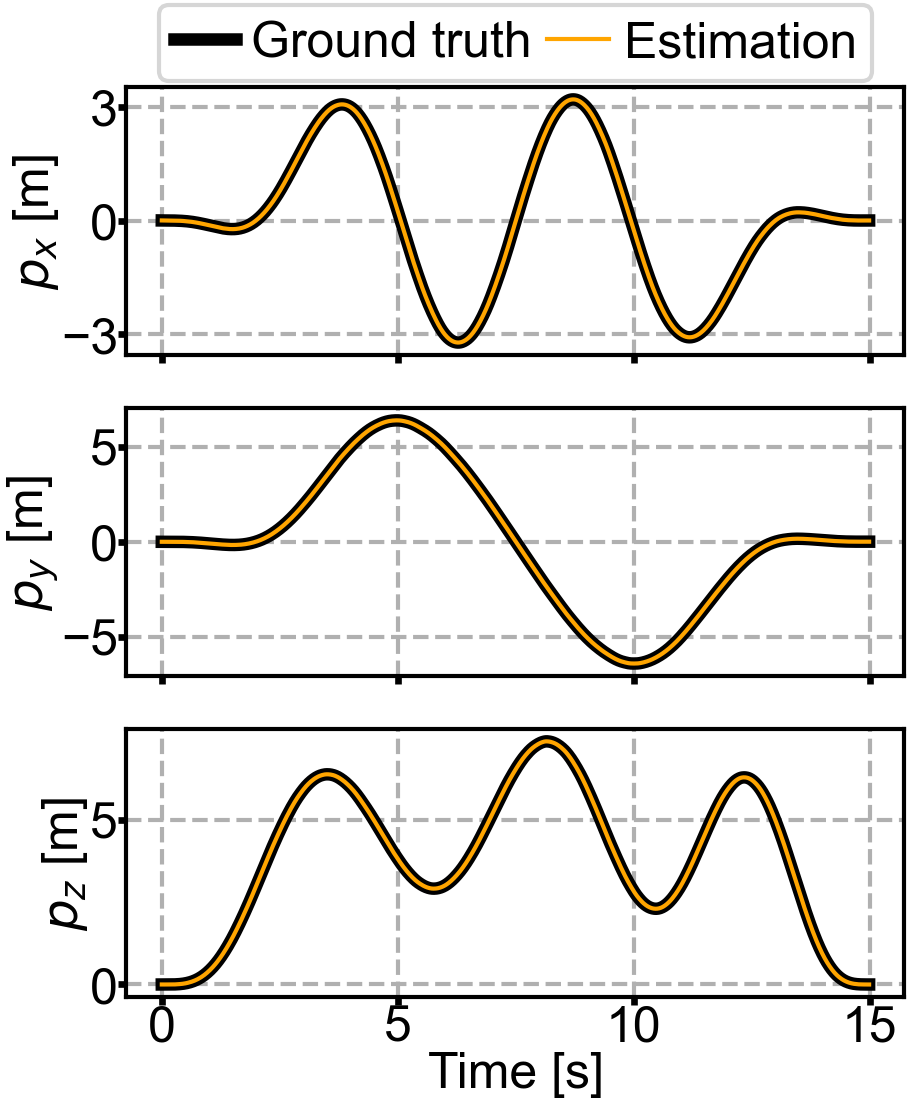}
\caption{\footnotesize Position estimation.}
\label{fig:position estimation}
\end{subfigure}
\hfill
\begin{subfigure}[b]{0.24\textwidth}
\centering
\includegraphics[width=1\textwidth]{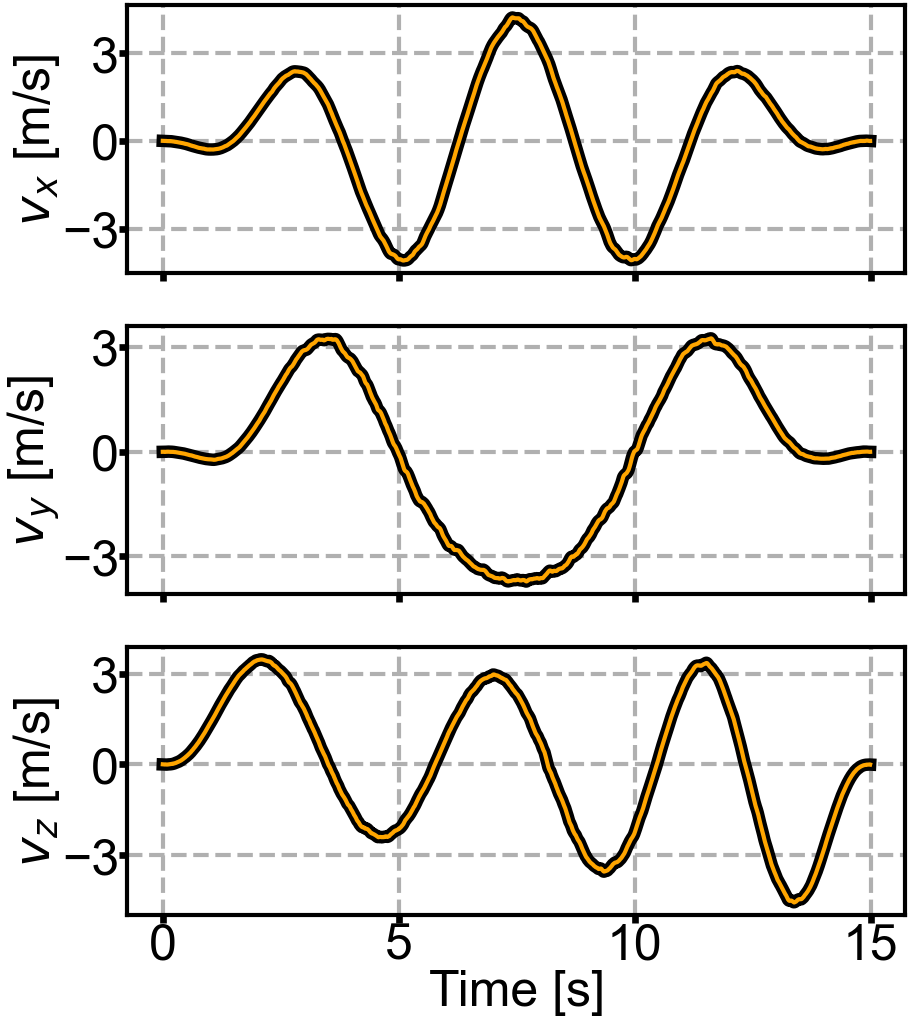}
\caption{\footnotesize Velocity estimation.}
\label{fig:velocity estimation}
\end{subfigure}
\hfill
\begin{subfigure}[b]{0.24\textwidth}
\centering
\includegraphics[width=1\textwidth]{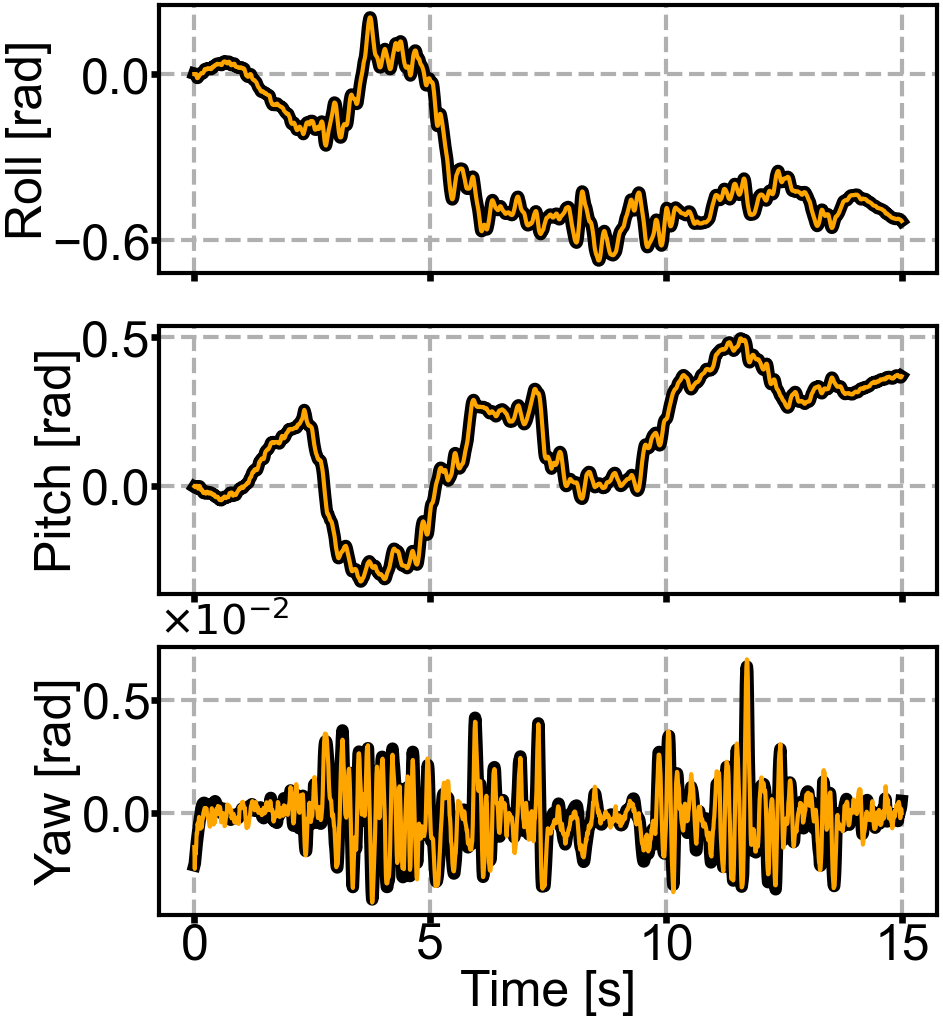}
\caption{\footnotesize Euler angle estimation.}
\label{fig:attitude estimation}
\end{subfigure}
\hfill
\begin{subfigure}[b]{0.24\textwidth}
\centering
\includegraphics[width=1\textwidth]{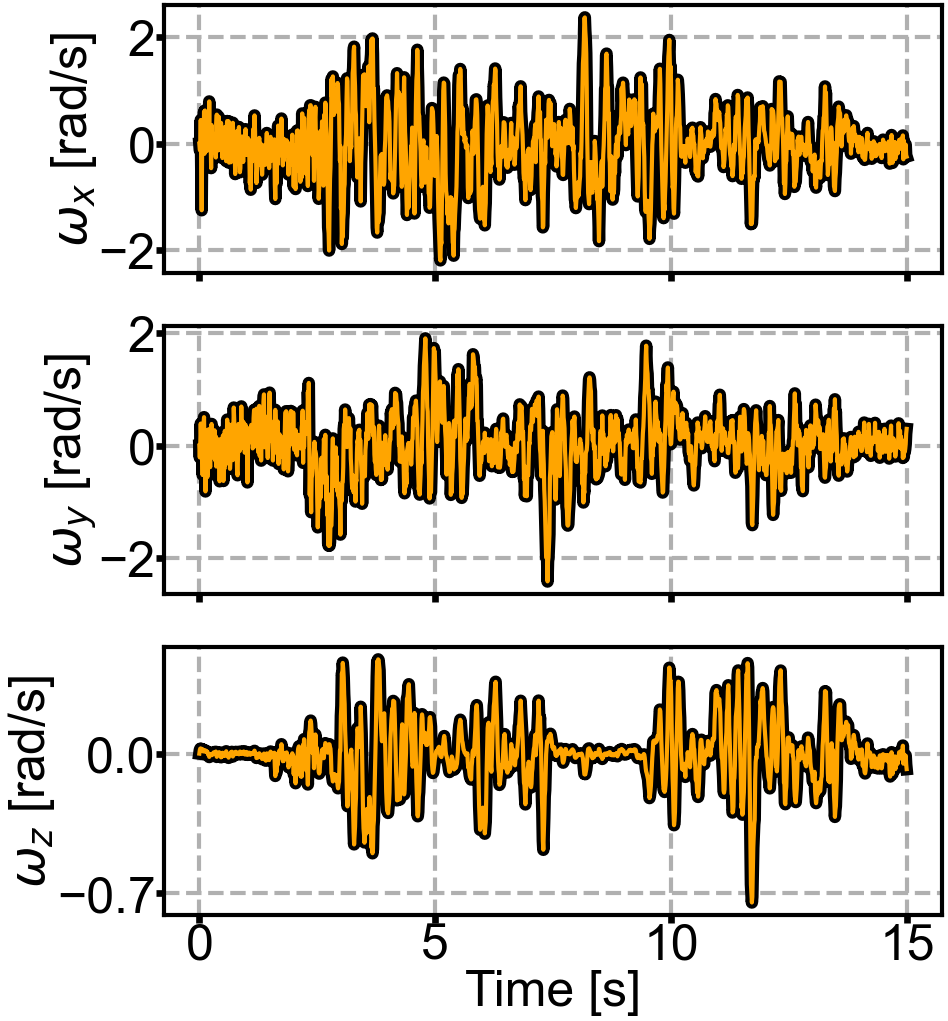}
\caption{\footnotesize Angular rate estimation.}
\label{fig:angular rate estimation}
\end{subfigure}
\caption{\footnotesize Estimation of the complete quadrotor states using the trained NeuroMHE on the training trajectory in the presence of the external disturbance. The standard deviations of the measurement noise $\boldsymbol \nu$ for the position, velocity, Euler angle, and angular rate are set to $\sigma _{p}=0.001\ {\rm m}$, $\sigma _{v}=0.005\ {\rm m/s}$, $\sigma _{\Theta_e }=0.001\ {\rm rad}$, and $\sigma _{\omega }=0.001\ {\rm rad/s}$. The RMSE of the estimation is $0.0003\ {\rm m}$ for $p_x$, $p_y$, $p_z$ positions. The RMSEs of the estimation are $0.0039\ {\rm m/s}$, $0.0048\ {\rm m/s}$, and $0.0048\ {\rm m/s}$ for $v_x$, $v_y$, $v_z$ velocities. The RMSEs of the estimation are $0.0003\ {\rm rad}$, $0.0004\ {\rm rad}$, $0.0003\ {\rm rad}$ for roll, pitch, yaw angles. Finally, the RMSEs of the estimation are $0.0009\ {\rm rad/s}$, $0.0009\ {\rm rad/s}$, $0.0007\ {\rm rad/s}$ for $\omega_x$, $\omega_y$, $\omega_z$ angular rates. The NeuroMHE estimations can track the ground truth values accurately and reduce the noise.}
\label{fig:state estimation training trajectory}
\end{figure}

\begin{figure}[h]
	\centering
	{\includegraphics[width=0.875\linewidth]{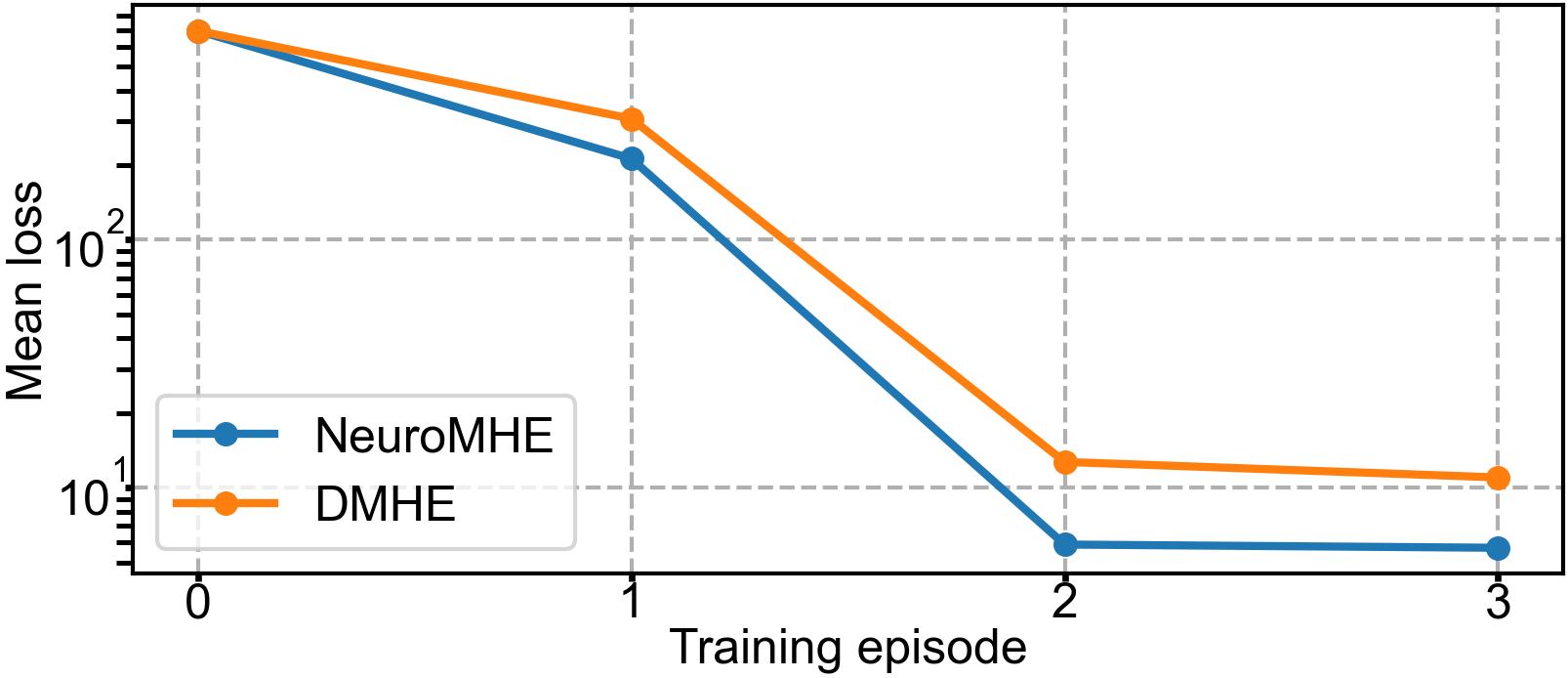}}
	\caption{\footnotesize Comparison of the mean loss between NeuroMHE and DMHE. The mean loss is obtained by averaging the loss (\ref{eq:loss function}) over the trajectory. For fair comparisons, the initial weightings $\boldsymbol \Theta_0$ of DMHE are manually tuned such that its untrained mean loss is very close to that of NeuroMHE.}
	\label{fig:mean loss}
\end{figure}

We begin with visualizing the training process by a 3D trajectory plot in Fig.~\ref{fig:trajectory online}. It shows that the tracking performance in all three directions is substantially improved online. Remarkably, the quadrotor can already track the reference trajectory accurately starting from the $2$nd training episode. This can be attributed to the rapid convergence  of the disturbance estimates in the planar and vertical directions, which is shown in Fig.~\ref{fig: force and torque in training} after only two training episodes. 
Fig.~\ref{fig:state estimation training trajectory} further illustrates the accuracy of NeuroMHE in estimating the complete quadrotor states on the training trajectory, which align closely with the ground truth values. We then train DMHE under the same conditions and compare its training loss with that of NeuroMHE in Fig.~\ref{fig:mean loss}. Although both methods exhibit efficient training, the mean loss of NeuroMHE stabilizes at a much smaller value, indicating better trajectory-tracking performance.

In addition to DMHE, we also compare NeuroMHE with a state-of-the-art adaptive controller proposed in~\cite{wu20221}. It employs $\mathcal{L}_1$ adaptive control ($\mathcal{L}_1$-AC) as an augmentation in the geometric controller to estimate and compensate for disturbances. We adopt the methodology presented in~\cite{wu20221} for designing the $\mathcal{L}_1$-AC, which comprises two primary components: 1) a state predictor with a tunable Hurwitz matrix $\bm A_s$, and 2) an adaptation law responsible for updating the estimations of both matched ($\hat{\bm d}_{\rm m}$) and unmatched ($\hat{\bm d}_{\rm um}$) disturbances. The control law, denoted by $\bm u_{\rm ad}$, for the $\mathcal{L}_1$-AC represents the estimated matched disturbance, which is processed via a low-pass filter. The detailed design procedures are provided in Appendix-\ref{appendix:l1-ac}. 

\begin{figure}[h]
\centering
\begin{subfigure}[b]{0.24\textwidth}
\centering
\includegraphics[width=0.95\textwidth]{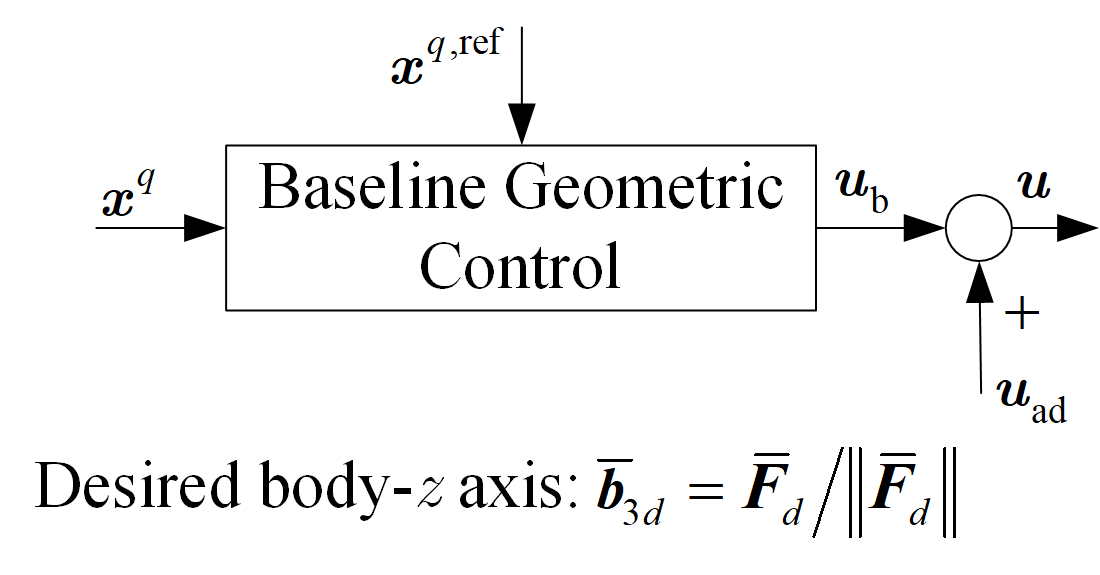}
\caption{\footnotesize Control architecture used in~\cite{wu20221}.}
\label{fig:l1 augmentation}
\end{subfigure}
\hfill
\begin{subfigure}[b]{0.24\textwidth}
\centering
\includegraphics[width=0.95\textwidth]{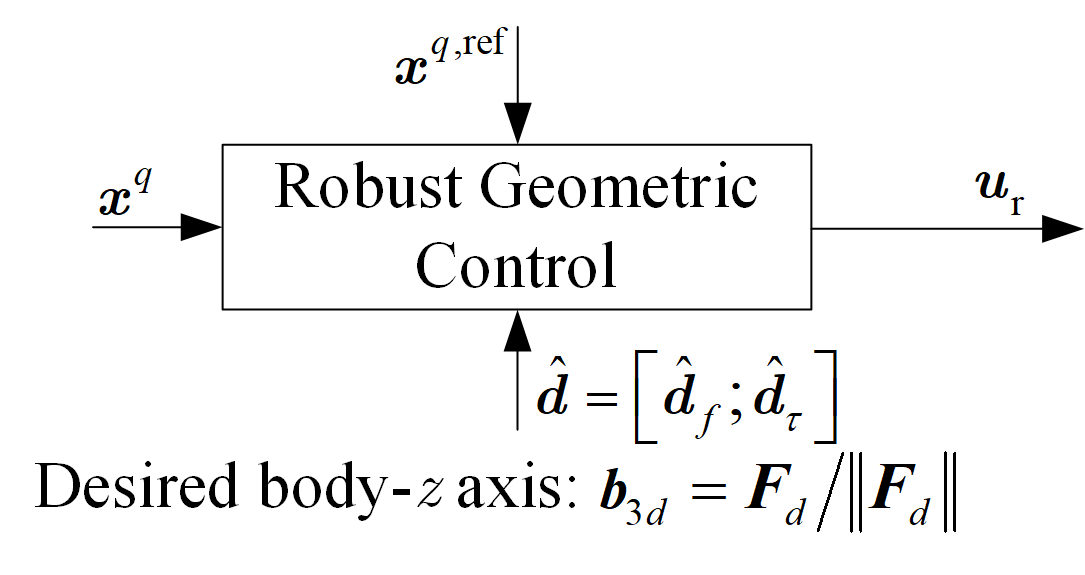}
\caption{\footnotesize Proposed control architecture.}
\label{fig:robust architecture}
\end{subfigure}
\caption{\footnotesize Comparison of the control architectures between the $\mathcal{L}_1$-augmented geometric controller~\cite{wu20221} and the proposed robust flight controller (\ref{eq:active control}). The definitions of $\bar{\bm F}_d$ and ${\bm F}_d$ are provided in (\ref{eq:desired nominal control force}) and (\ref{eq:desired active control force}), respectively, while the integration of the desired body-$z$ axis into the desired rotation matrix $\bm R_d$ is further elaborated in Appendix-\ref{appendix:geometric}.}
\label{fig: control framework comparison}
\end{figure}

Fig.~\ref{fig: control framework comparison} compares the control architectures between the $\mathcal{L}_1$-augmented geometric controller~\cite{wu20221} and our robust geometric controller~(\ref{eq:active control}). Fig.~\ref{fig:l1 augmentation} shows that the $\mathcal{L}_1$ adaptive control signal $\bm u_{\rm ad}$ is added directly to the baseline geometric control input ${\bm u}_{\rm b}$, leaving the desired attitude unaffected by the disturbance estimate $\hat{\bm d}$. This is reflected by the computation of the desired body-$z$ axis (\ref{eq: desired z axis}), which uses the nominal control force $\bar {\bm F}_d$. In contrast, our control architecture (Fig.~\ref{fig:robust architecture}) updates both the control input and the desired attitude using the estimation $\hat{\bm d}$. Our improved design enables the quadrotor to actively respond to the horizontal disturbances in the world frame and compensate for them as much as possible through rotation. Note that the $\mathcal{L}_1$-AC can also be applied using the proposed control architecture. To do so, we first express the estimated disturbance force from (\ref{eq:adaptation law}) in the world frame, calculate ${\bm R}_d$ for the robust geometric control input ${\bm u}_{\rm r}$ and finally replace ${\bm u}_{\rm b}+{\bm u}_{\rm ad}$ with ${\bm u}_{\rm r}$. We will refer to this improved $\mathcal{L}_1$-augmented robust geometric controller as Active $\mathcal{L}_1$-AC.

\begin{figure}[h]
\centering
\begin{subfigure}[b]{0.49\textwidth}
\centering
\includegraphics[width=0.815\textwidth]{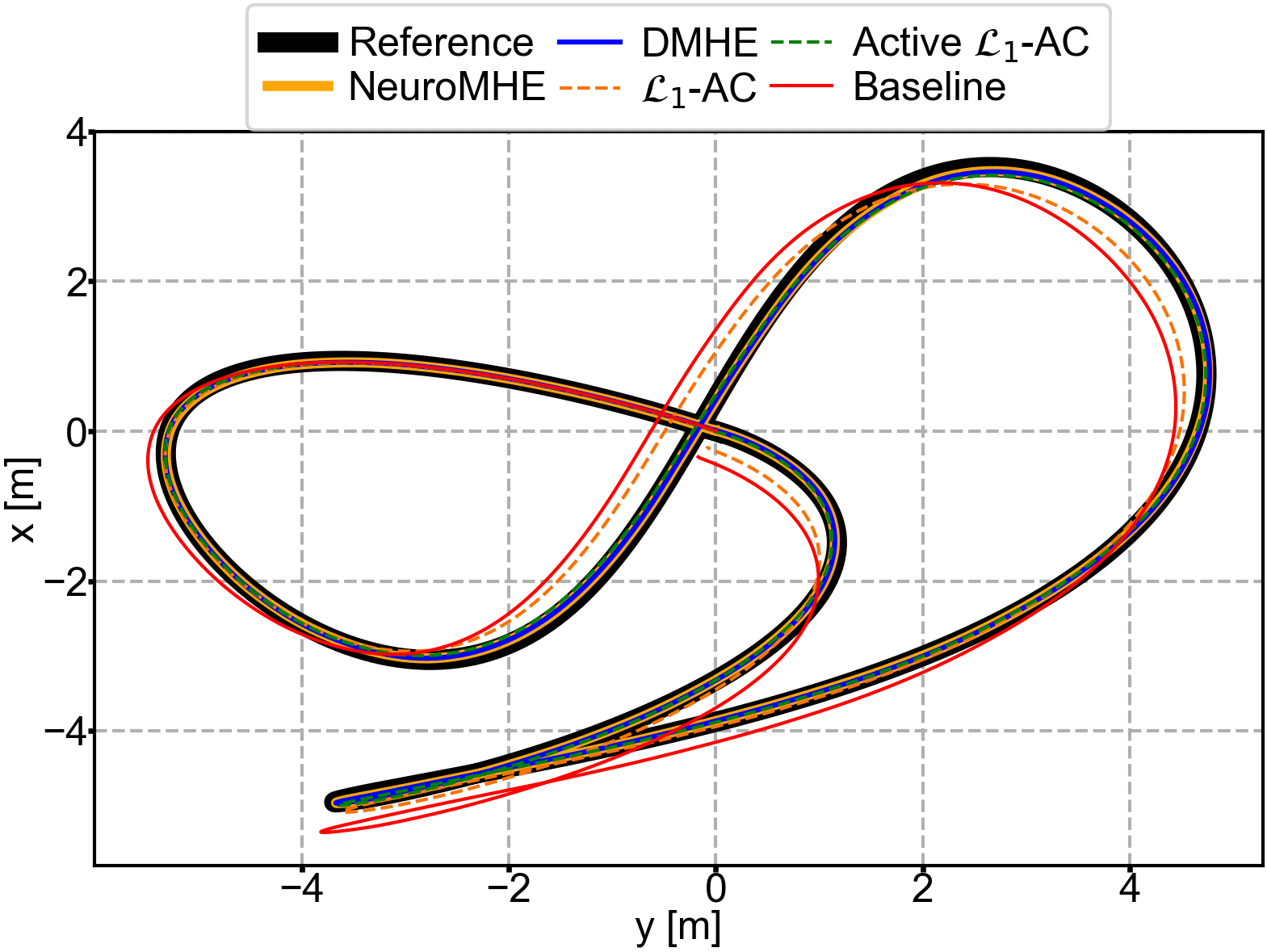}
\caption{\footnotesize Horizontal tracking performance.}
\label{fig:planar race-track}
\end{subfigure}
\hfill
\begin{subfigure}[b]{0.49\textwidth}
\centering
\includegraphics[width=0.86\textwidth]{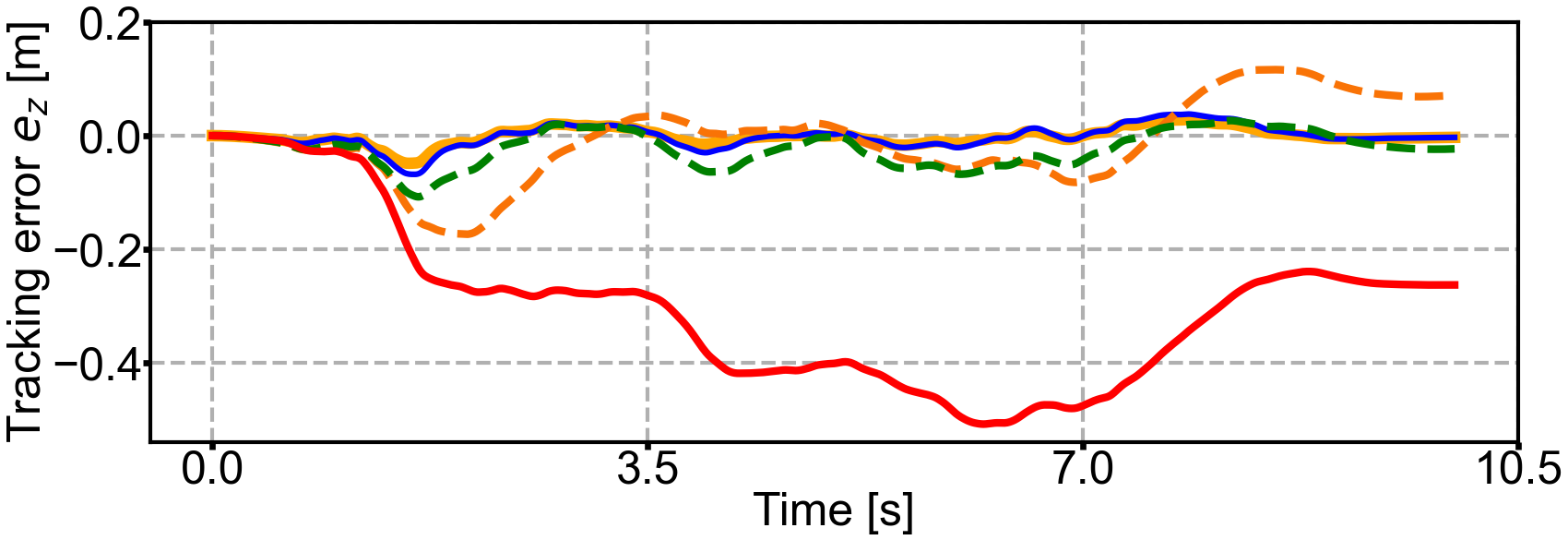}
\caption{\footnotesize Vertical tracking error.}
\label{fig:tracking error z race-track}
\end{subfigure}
\caption{\footnotesize Comparison between the tracking performances of NeuroMHE, DMHE, $\mathcal{L}_1$-AC, Active $\mathcal{L}_1$-AC, and the baseline geometric controller on a previously unseen race-track trajectory.}
\label{fig: tracking performance on race-track}
\end{figure}

\begin{figure}[h]
\centering
\begin{subfigure}[b]{0.24\textwidth}
\centering
\includegraphics[width=1\textwidth]{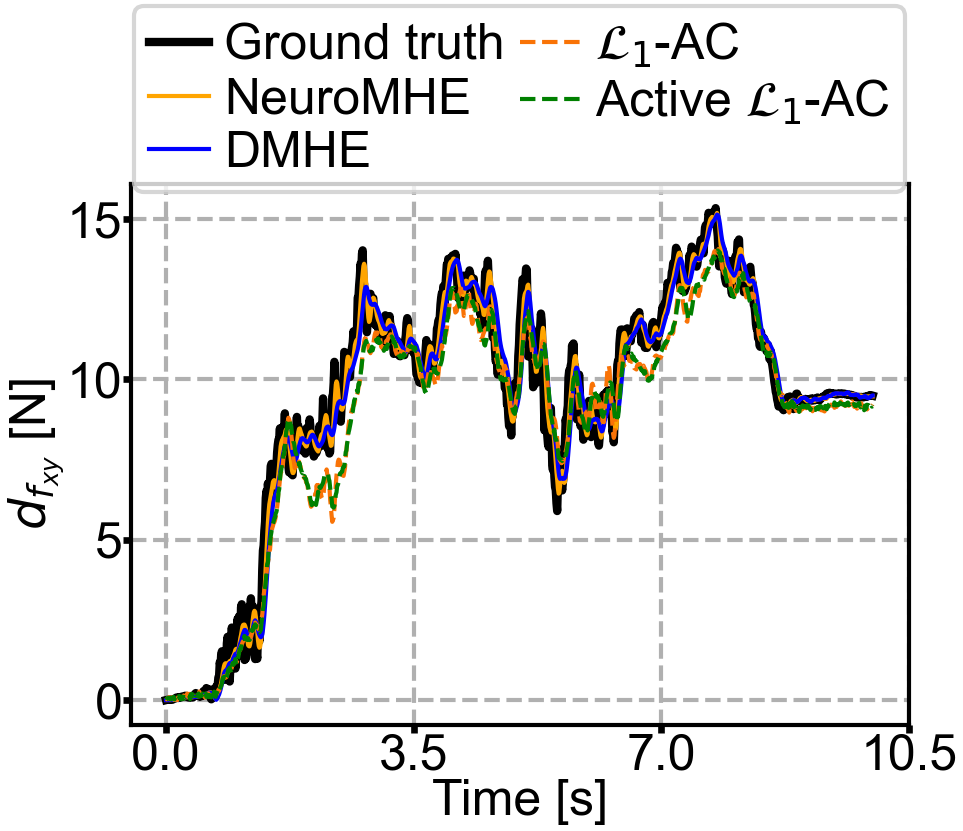}
\caption{\footnotesize Disturbance force in $xy$ plane.}
\label{fig:force xy synthetic}
\end{subfigure}
\hfill
\begin{subfigure}[b]{0.24\textwidth}
\centering
\includegraphics[width=1\textwidth]{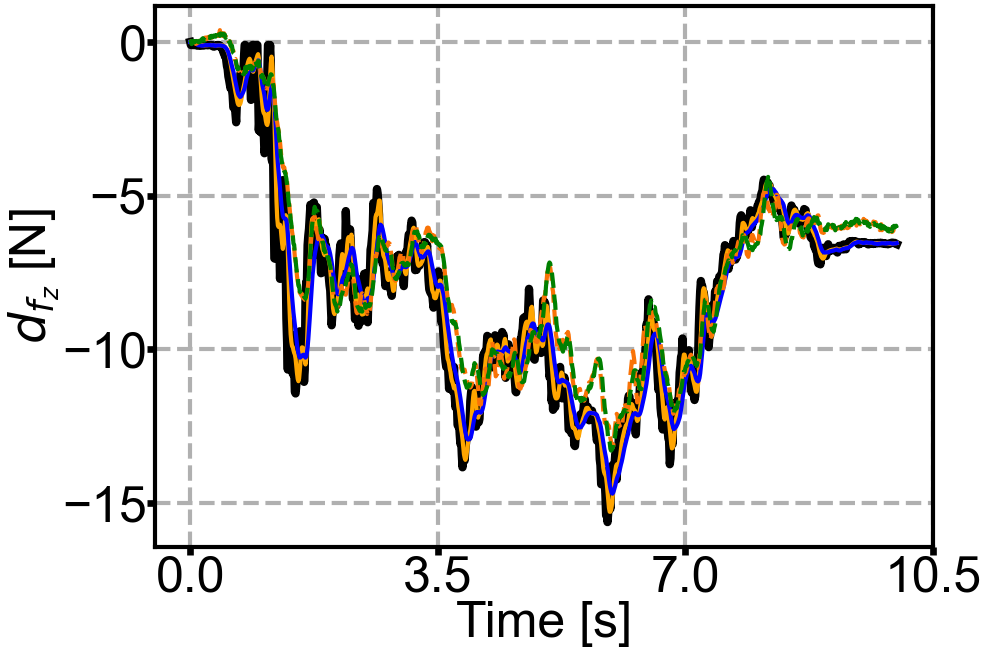}
\caption{\footnotesize Disturbance force in $z$ axis.}
\label{fig:force z synthetic}
\end{subfigure}
\hfill
\begin{subfigure}[b]{0.24\textwidth}
\centering
\includegraphics[width=1\textwidth]{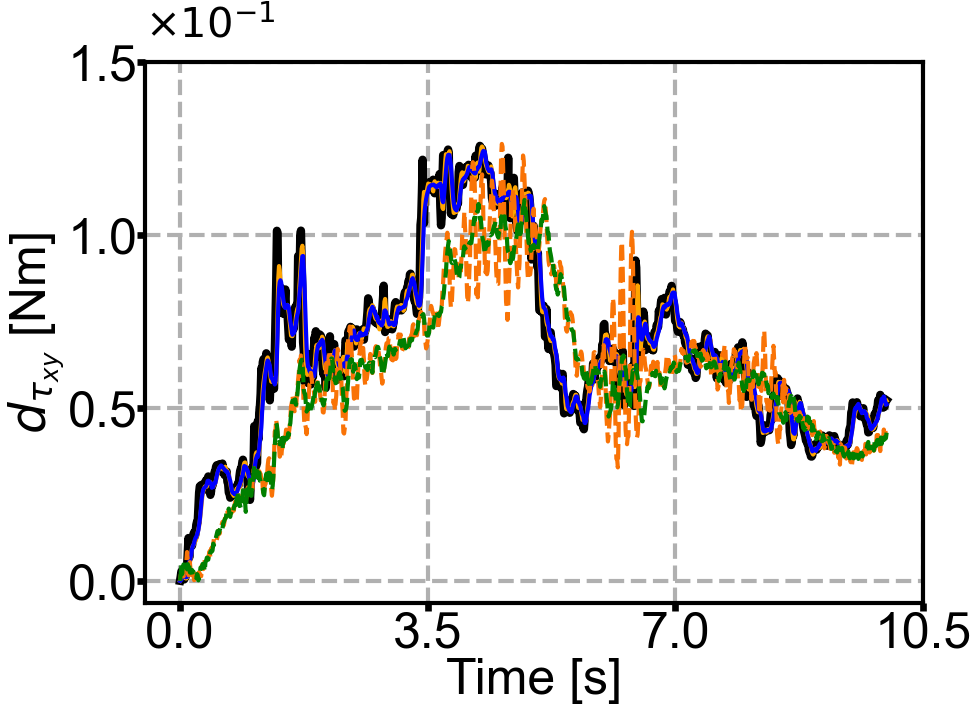}
\caption{\footnotesize Disturbance torque in $xy$ plane.}
\label{fig:torque xy synthetic}
\end{subfigure}
\hfill
\begin{subfigure}[b]{0.24\textwidth}
\centering
\includegraphics[width=1\textwidth]{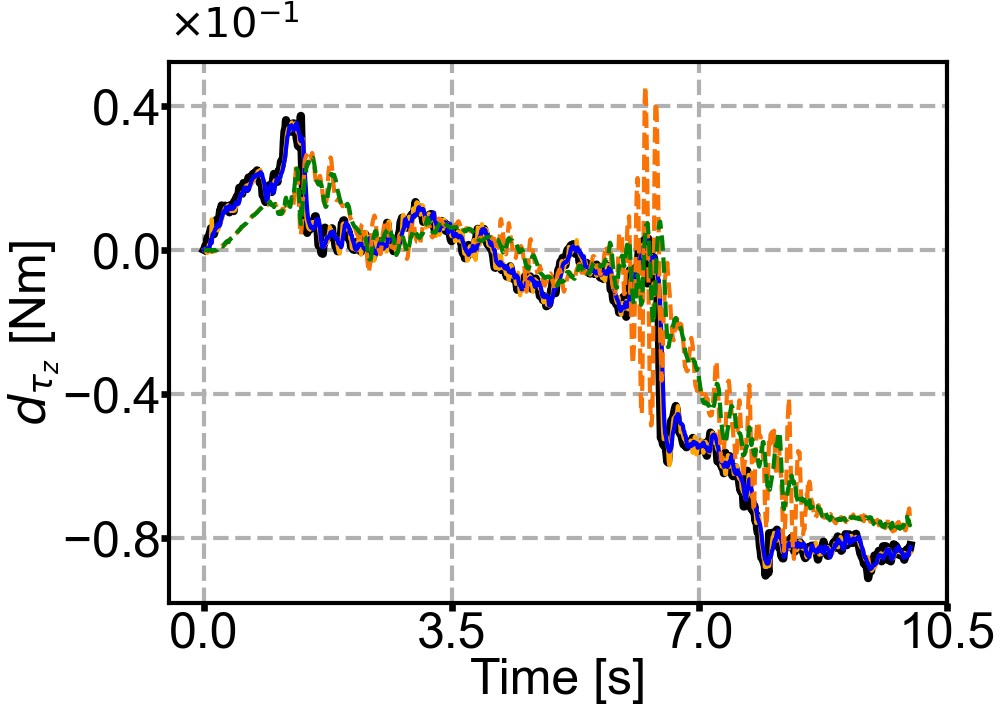}
\caption{\footnotesize Disturbance torque in $z$ axis.}
\label{fig:torque z synthetic}
\end{subfigure}
\caption{\footnotesize Comparison between the disturbance estimation performances of NeuroMHE, DMHE, $\mathcal{L}_1$-AC, and Active $\mathcal{L}_1$-AC on an unseen race-track trajectory with an unseen synthetic disturbance dataset. The plotted estimated disturbance forces are in the world frame. The disturbance estimations from the $\mathcal{L}_1$-AC methods are filtered by the LPF as required by the $\mathcal{L}_1$ adaptive control law. }
\label{fig: disturbance estimation synthetic data}
\end{figure}

\begin{figure}[h]
\centering
\begin{subfigure}[b]{0.24\textwidth}
\centering
\includegraphics[width=1\textwidth]{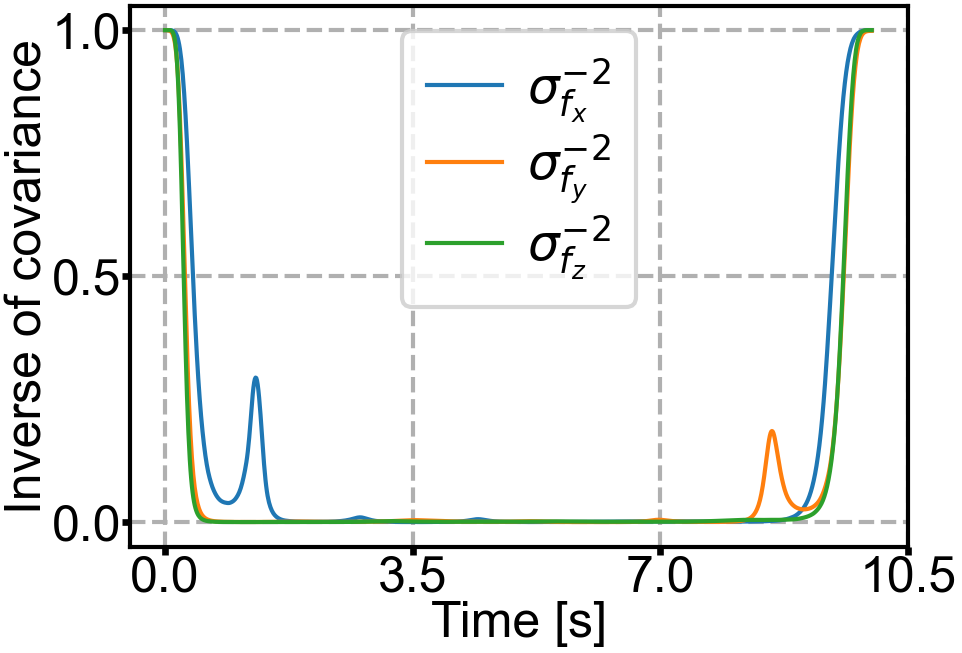}
\caption{\footnotesize Inverse of the covariance of $\bm w_{f}$.}
\label{fig:inverse force noise}
\end{subfigure}
\hfill
\begin{subfigure}[b]{0.24\textwidth}
\centering
\includegraphics[width=1\textwidth]{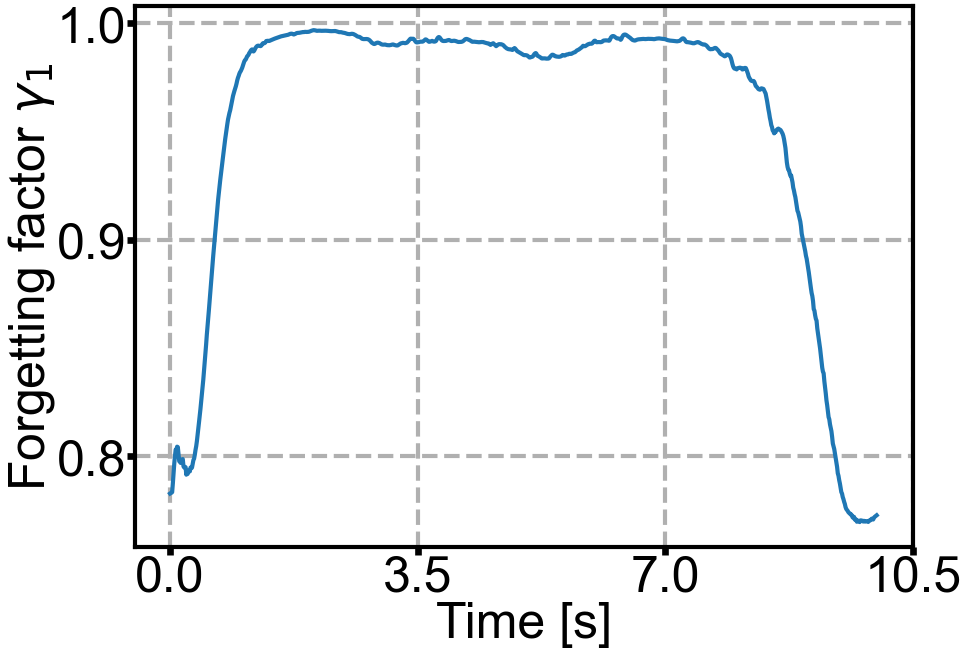}
\caption{\footnotesize Forgetting factor $\gamma_1$.}
\label{fig:forgetting factor 1}
\end{subfigure}
\hfill
\begin{subfigure}[b]{0.24\textwidth}
\centering
\includegraphics[width=1\textwidth]{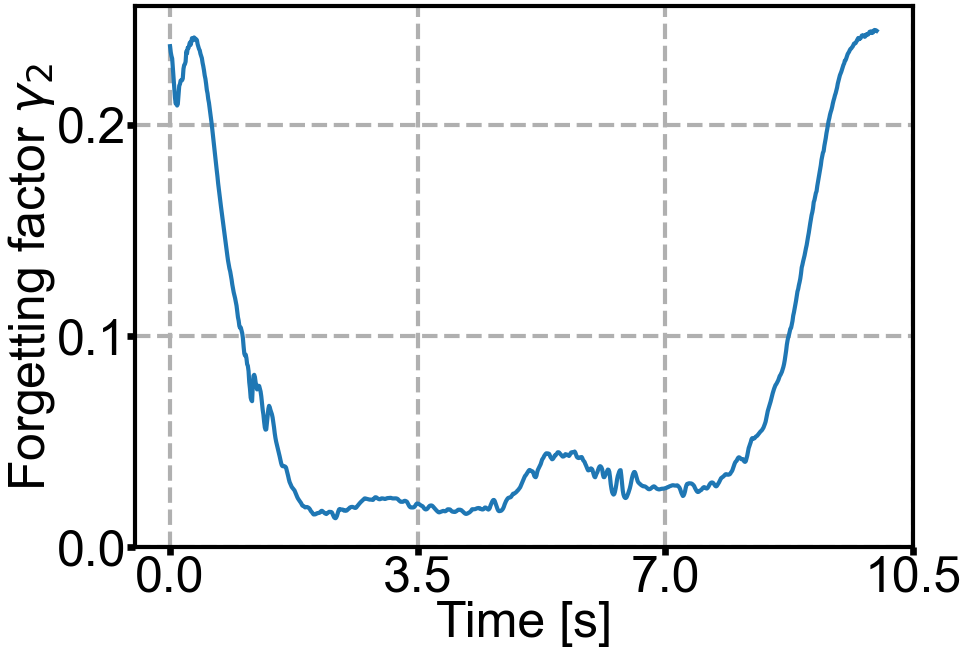}
\caption{\footnotesize Forgetting factor $\gamma_2$.}
\label{fig:forgetting factor 2}
\end{subfigure}
\hfill
\begin{subfigure}[b]{0.24\textwidth}
\centering
\includegraphics[width=1\textwidth]{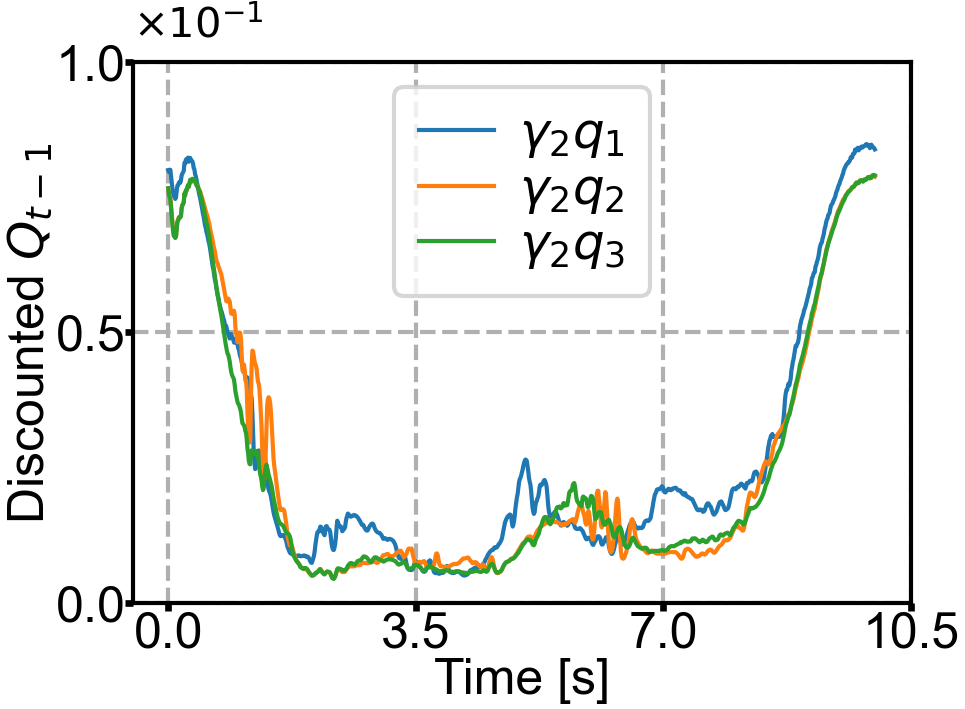}
\caption{\footnotesize Discounted ${\bm Q}_{t-1}$ with $\gamma_2$}
\label{fig:discounted Q elements}
\end{subfigure}
\caption{\footnotesize Comparison of the inverse of the noise covariance parameters and the NeuroMHE weighting matrices on the unseen race-track trajectory. Here, we show one example of the discounted $\gamma_{2}{\bm Q}_{t-1}$ which penalizes $\bm w_f$ in the MHE cost function (\ref{eq:mhe cost}). The remaining ${\bm Q_k} = \gamma _2^{t - 1 - k}{\bm Q_{t - 1}}$ with $k = t - N, \cdots ,t - 3$ have the similar changing pattern.}
\label{fig: optimal weighting matrices evaluation}
\end{figure}

\subsubsection{Evaluation and Comparison on an Unseen Trajectory}\label{subsubsection: evaluation unseen trajectory}
we evaluate the trained NeuroMHE and compare it with DMHE, $\mathcal{L}_1$-AC, Active $\mathcal{L}_1$-AC, and the baseline geometric controller on an unseen trajectory to show the third advantage of our method. First, we create an offline disturbance dataset for the purpose of equitable evaluation and comparison within a single episode. It is generated by integrating the random walk model (\ref{eq: random walk}) with the same noise covariance parameters as in training, but on a previously unseen race-tracking trajectory. For the $\mathcal{L}_1$-AC and the Active $\mathcal{L}_1$-AC in evaluation, we set $T_{s}=10\ {\rm ms}$ and ${\bm A}_{s}=-{\rm diag}\left (\left [ 5,5,5,10,10,10 \right ]  \right )$. For the associated LPFs, we use a first-order LPF with $\omega _{c}=8\ {\rm {rad} \mathord{\left/
 {\vphantom {rad s}} \right.
 \kern-\nulldelimiterspace} \rm s}$ for the force channel and two cascaded first-order LPFs with $\omega _{c1}=4\ {\rm {rad} \mathord{\left/
 {\vphantom {rad s}} \right.
 \kern-\nulldelimiterspace} \rm s}$ and $\omega _{c2}=6\ {\rm {rad} \mathord{\left/
 {\vphantom {rad s}} \right.
 \kern-\nulldelimiterspace} \rm s}$ for the torque channel. These parameters of $\mathcal{L}_1$-AC are manually tuned to achieve the best estimation and trajectory tracking performance while keeping the system stable. The control gains are the same as those in training.
 
Fig.~\ref{fig: tracking performance on race-track} illustrates the tracking performance of each controller on the unseen race-track trajectory. Significant tracking errors are observed in both the horizontal and vertical directions when using the baseline controller alone. $\mathcal{L}_1$-AC only slightly improves the tracking error in the horizontal plane over the baseline controller, whereas Active $\mathcal{L}_1$-AC, DMHE, and NeuroMHE all achieve much more accurate tracking. This comparison clearly shows the advantage of the proposed robust control architecture. Fig.~\ref{fig: disturbance estimation synthetic data} compares the disturbance estimation performance among all methods except for the baseline controller. Both NeuroMHE and DMHE demonstrate more accurate estimation than $\mathcal{L}_1$-AC and Active $\mathcal{L}_1$-AC, particularly in the disturbance torque estimation, where the latter two algorithms have large oscillations and biases. Fig.~\ref{fig:force xy synthetic} and Fig.~\ref{fig:force z synthetic} also shows that NeuroMHE outperforms DMHE in terms of smaller time lag and estimation error.

The advantages of NeuroMHE over DMHE are attributed to its MLP-modeled weightings, which effectively adapt to the inverse of the noise covariance parameters on the unseen trajectory, as shown in Fig.~\ref{fig: optimal weighting matrices evaluation}. The neural network's increased $\gamma_1$ and decreased $\gamma_2$ impose a greater penalty on the predicted measurement error than on the process noise during the period of $1.5\sim 8.5\ {\rm s}$. This time-varying pattern allows for rapid changes of the force estimates, thus enabling NeuroMHE to track the fast-changing disturbance forces during that period. In contrast, without the neural network, the optimized values of $\gamma_1$ and $\gamma_2$ in the trained DMHE are fixed at $0.788$ and $0.183$, respectively.

\begin{figure}[h]
\centering
\begin{subfigure}[b]{0.49\textwidth}
\centering
\includegraphics[width=0.86\textwidth]{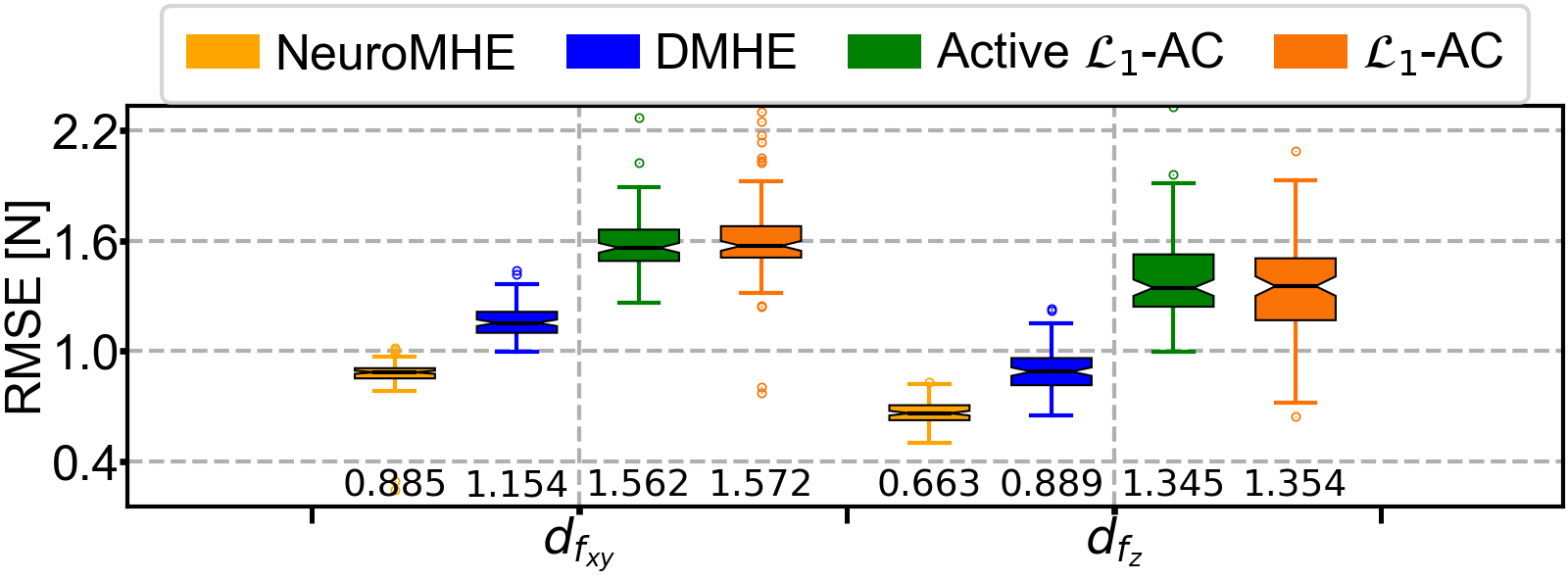}
\caption{\footnotesize Comparison of the force estimation RMSEs.}
\label{fig:rmse force}
\end{subfigure}
\hfill
\begin{subfigure}[b]{0.49\textwidth}
\centering
\includegraphics[width=0.86\textwidth]{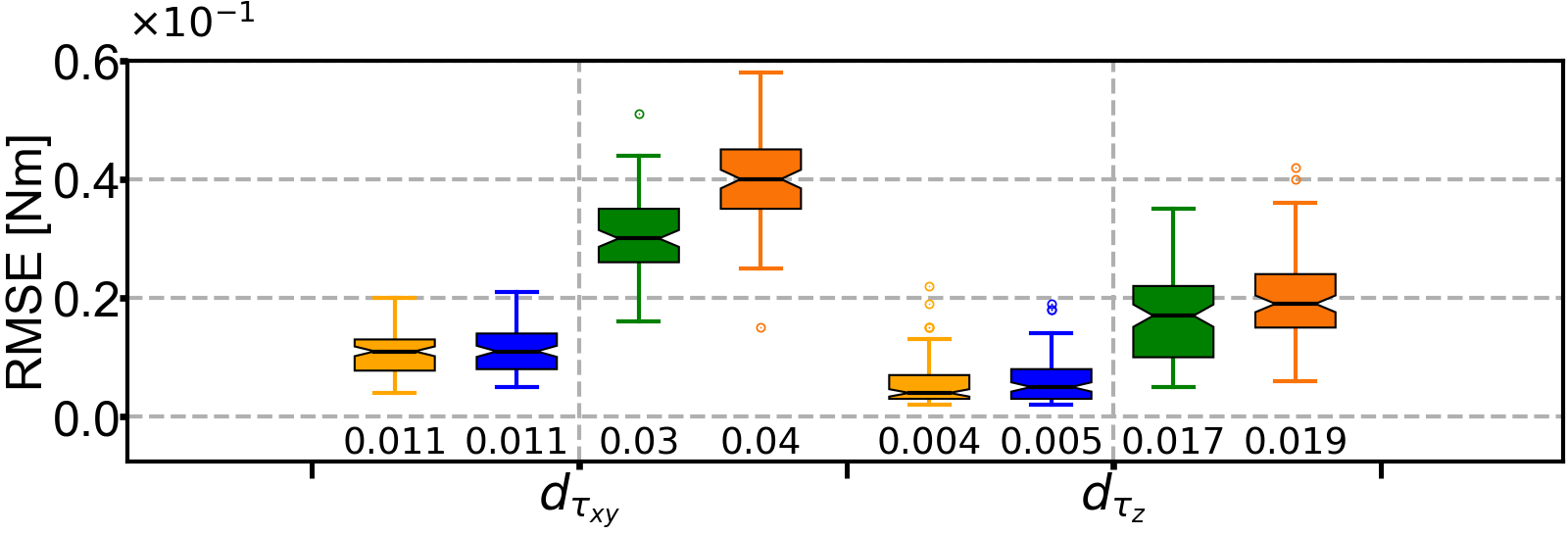}
\caption{\footnotesize Comparison of the torque estimation RMSEs.}
\label{fig:rmse torque}
\end{subfigure}
\hfill
\begin{subfigure}[b]{0.49\textwidth}
\centering
\includegraphics[width=0.86\textwidth]{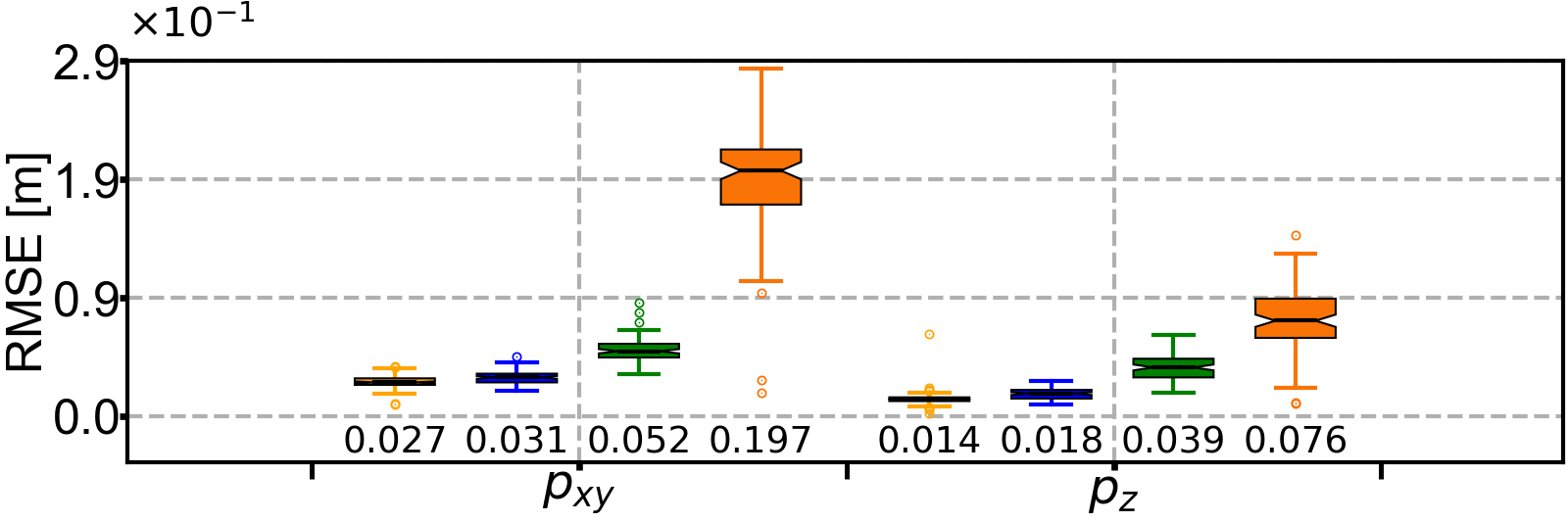}
\caption{\footnotesize Comparison of the position tracking RMSEs.}
\label{fig:rmse position}
\end{subfigure}
\caption{\footnotesize Boxplots of evaluated estimation and trajectory tracking performance in terms of the RMSEs over $100$ episodes. The solid black line in each box denotes the median value, which is also indicated at the bottom of each subplot. All outliers are defined following the widely-used $1.5\times$IQR (Interquartile Range) rule. The RMSE values for the planar data (i.e., $d_{f_{xy}}$, $d_{\tau_{xy}}$, and $p_{xy}$) are computed using the vector error, as defined in the caption of Table~\ref{table:testset rmse}.}
\label{fig: rmse comparison}
\end{figure}

We further compare NeuroMHE, DMHE, $\mathcal{L}_1$-AC, and Active $\mathcal{L}_1$-AC over $100$ episodes. In each episode, the quadrotor is controlled to follow the same race-track trajectory as in evaluation, and the external disturbances are updated online by integrating the random walk model (\ref{eq: random walk}) with the state-dependent noise using the current quadrotor state. In Fig.~\ref{fig: rmse comparison}, we present boxplots of the force and torque estimation error as well as the trajectory tracking error in terms of RMSE. Fig.~\ref{fig:rmse force} and Fig.~\ref{fig:rmse torque} demonstrate that NeuroMHE outperforms all other methods in all spacial directions, reducing force and torque estimation errors by up to $51\%$ and $78.9\%$, respectively. The accurate estimation of NeuroMHE leads to superior trajectory-tracking performance with up to $86.3\%$ tracking error reduction over the other methods, as shown in Fig.~\ref{fig:rmse position}.


Overall, our simulation results indicate:

\begin{enumerate}
    \item A stable NeuroMHE with a fast dynamic response can be efficiently learned online in merely a few episodes from the quadrotor trajectory tracking error without the need for the ground-truth disturbance data.
    \item The neural network can generate the adaptive weightings online, thus improving both the estimation and control performance of the NeuroMHE-augmented flight controller when the noise covariances are dynamic.
\end{enumerate}

\subsection{Physical Experiments}
\label{subsec:expC}
\begin{figure}[h]
	\centering
	{\includegraphics[width=0.95\linewidth]{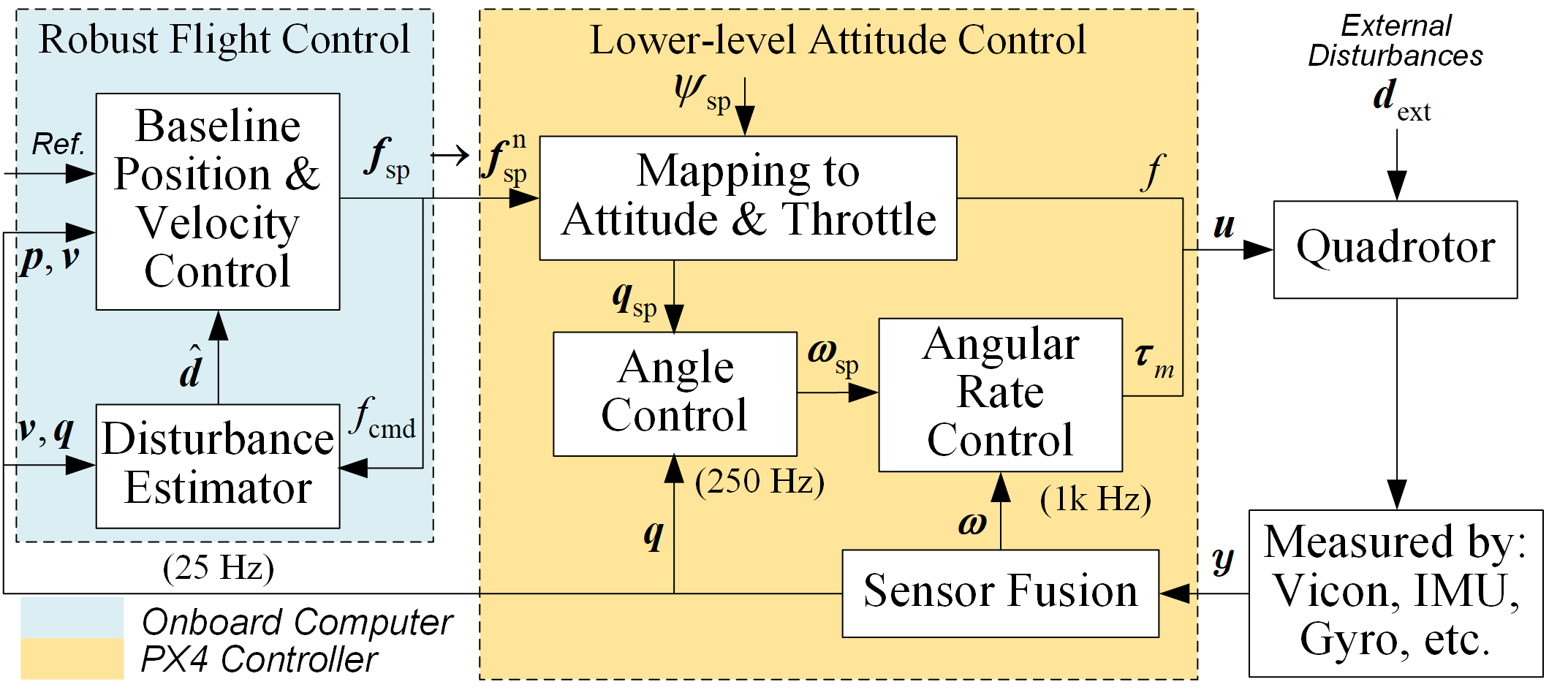}}
	\caption{\footnotesize Diagram of the cascaded control architecture for the experiments. In the robust flight controller, $f_{\rm cmd}$ is the collective thrust command obtained by projecting the desired control force setpoint $\bm f_{\rm sp}\in \mathbb{R}^{3}$ in $\cal {\bm I} $ onto the current body-$z$ axis: $f_{\rm cmd}={\bm f}_{\rm sp}\cdot \left ( {\bm R}{\bm e}_{3} \right )$. We set the yaw angle setpoint to $\psi _{\rm sp}=0\ {\rm rad}$. Note that $f$ and $\boldsymbol\tau_m$ denote the true collective thrust and torque generated by the rotor aerodynamics. For simplicity, we have omitted the motor dynamics block in the figure.}
	\label{fig: control diagram overview}
\end{figure}

Finally, we validate the performance of the proposed NeuroMHE-based robust flight controller and compare it with that of DMHE-based and $\mathcal{L}_1$-AC-based robust controllers, as well as the baseline controller, on a real quadrotor under various external disturbance forces. The baseline controller is a Proportional-Derivative (PD) controller with gravity compensation. In the robust version of the controller, the estimated disturbance force is included as a feedforward term to compensate for the disturbance. To demonstrate the effectiveness of our approach in improving tracking and stabilization performance under state-dependent disturbances and complex aerodynamic effects, we conduct experiments in two settings:
\begin{enumerate}
    \item Setting A: Trajectory-tracking control in the presence of the state-dependent cable forces; and
    \item Setting B: Holding the position against the challenging aerodynamic disturbances from both an electric fan and the downwash flow generated by a second quadrotor.
\end{enumerate}
\begin{figure}[h]
	\centering
	{\includegraphics[width=0.8\linewidth]{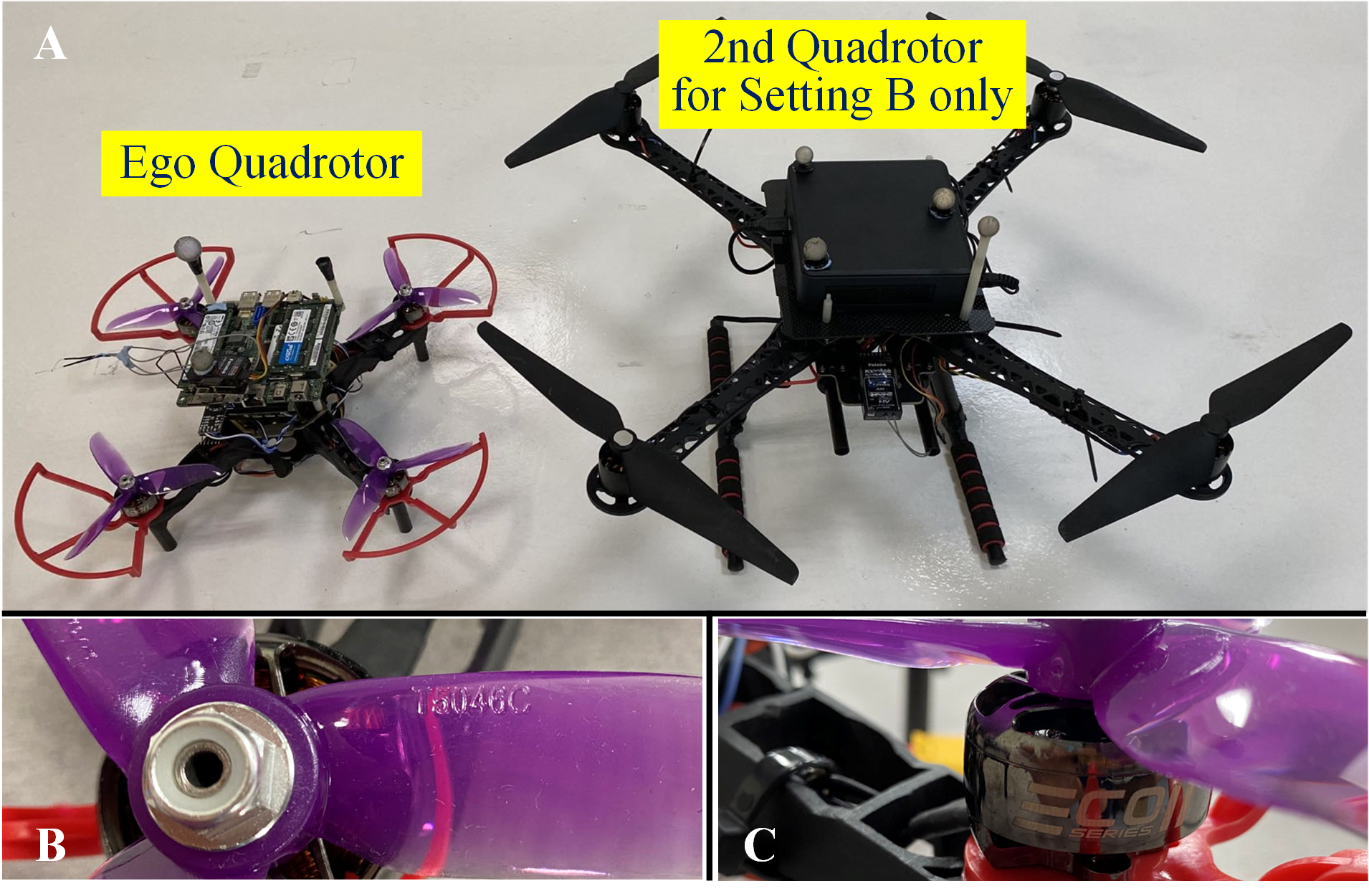}}
	\caption{\footnotesize Overview of the two quadrotors for the experiments. A): The left one is the ego quadrotor used in both two settings. The right one, which is larger and heavier, is used only in Setting B where it flies above the ego one to generate the downwash flow. B): The model of the propeller of the ego quadrotor is DALPROP Cyclone T5046C. C): The motor type of the ego quadrotor is ECOII2306-2400KV.}
	\label{fig: quadrotor overview}
\end{figure}
In the experiments, we employ two quadrotors as shown in Fig.~\ref{fig: quadrotor overview}. The ego quadrotor weighs $1\ {\rm kg}$ and is equipped with a Pixhawk autopilot and a 64-bit Intel NUC onborad computer featuring an Intel Core i7-5557U CPU. The onboard computer runs a Linux operating system to perform real-time computation of the robust flight control algorithms and interface with the autopilot through the Robot Operating System (ROS). We implement our method based on the off-the-shelf PX4 control firmware, as illustrated in Fig.~\ref{fig: control diagram overview}. Specifically, we modify the official v1.11.1 PX4 firmware to bypass the PX4's position and velocity controllers. Since the PX4 attitude control pipeline requires the normalized force setpoint ${\bm f}_{\rm sp}^{\rm n}$, we convert $\bm f_{\rm sp}$ to ${\bm f}_{\rm sp}^{\rm n}$ via ${\bm f}_{\rm sp}^{\rm n}={\bm f}_{\rm sp}/\left ( 4f_{\max} \right )$ where ${f_{\max }} = 1.25g\ {\rm N}$ is the maximum thrust\footnote{The maximum thrust, recorded as $1480\ {\rm g}$ for the used propeller and motor, is available as a reference at \url{https://emaxmodel.com/products/emax-eco-ii-series-2306-1700kv-1900kv-2400kv-brushless-motor-for-rc-drone-fpv-racing}. It is obtained at $16.8\ {\rm V}$, representing the ideal maximum battery voltage. However, practical conditions lead to a reduced actual maximum thrust due to factors such as battery depletion. 
To ascertain a more feasible $f_{\max}$ value, we perform ground tests at various voltages and fit a polynomial model to the thrust data. This process yields an updated $f_{\max}$ of $1250\ {\rm g}$ (i.e., $1.25g\ {\rm N}$) at a nominal voltage of around $15\ {\rm V}$.} provided by each rotor with the local gravity constant $g = 9.78\ {{\rm{m}} \mathord{\left/
 {\vphantom {{\rm{m}} {{{\rm{s}}^2}}}} \right.
 \kern-\nulldelimiterspace} {{{\rm{s}}^2}}}$ in Singapore. A Vicon motion capture system, which records the quadrotor's pose at $100\ {\rm Hz}$, transmits the pose data to the onboard computer through Wi-Fi. The PX4 employs an Extended Kalman Filter (EKF) to fuse the pose measurements with the data from other sensors, such as the Inertial Measurement Unit (IMU) and the gyroscope, to estimate the quadrotor's state. By leveraging the sensor fusion, we compare the disturbance estimation performance of NeuroMHE, DMHE, and $\mathcal{L}_1$-AC, and evaluate their impact on the control performance.

\begin{figure}[h]
	\centering
	{\includegraphics[width=0.8\linewidth]{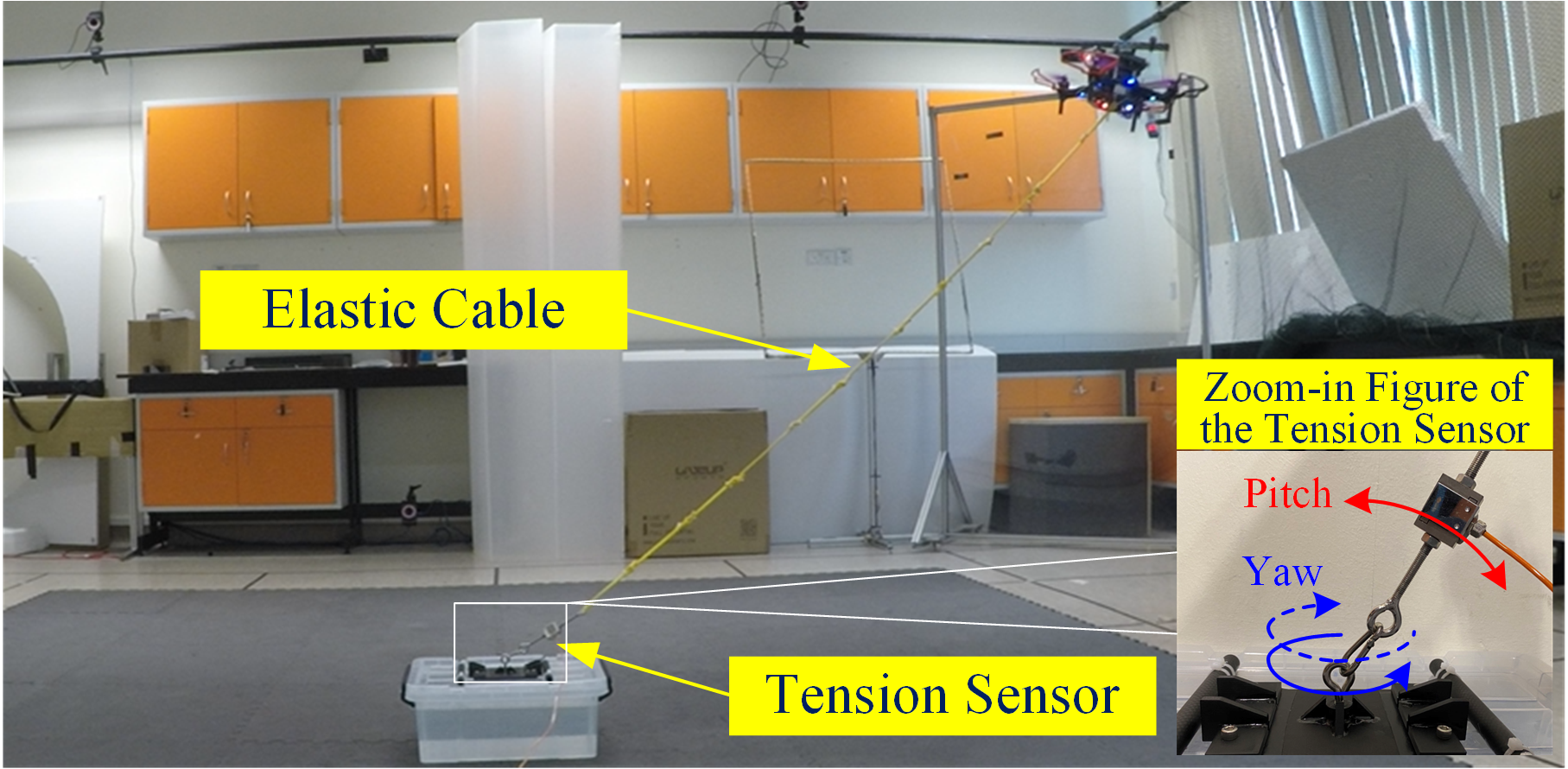}}
	\caption{\footnotesize Experimental setup of Setting A. The elastic cable weighs about $40\ {\rm g}$. The tension sensor, linking the cable and a heavy water box, has a weight of about $120\ {\rm g}$. The box is at the origin of East-North-Up (ENU) world frame. As depicted in the zoom-in figure, our setup enables the tension sensor to move freely in the pitch and yaw directions, facilitating an accurate measurement of the tension force within the cable.}
	\label{fig: experiment setting 1}
\end{figure}

\begin{remark}
    Our disturbance estimator utilizes the collective thrust command $f_{\rm cmd}$ instead of the true collective thrust $f$. Acquiring the latter accurately is challenging in practice due to the complicated rotor aerodynamics, the open-loop rotor speed control accuracy, the heuristic force command normalization in PX4 autopilot, and the impact of battery depletion, etc. By using the commanded value, NeuroMHE actually estimates the aggregation of external disturbances $\bm d_{\rm ext}$, such as the tension force, and internal uncertainties arising from the unmeasurable discrepancy $\Delta f=f-f_{\rm cmd}$. Specifically, the estimated disturbance is the sum of two components: $\bm d={\bm R}\Delta f{\bm e}_{3}+{\bm d}_{\rm ext}$. This approach provides a more convenient strategy for enhancing the robustness of existing flight controllers. Additionally, although the tension sensor is allowed to move freely in pitch and yaw (See Fig.~\ref{fig: experiment setting 1}), there are times when it does not align with the cable precisely due to its own weight. Hence, in these instances the force sensor only measures the cable tension component projected onto its measured direction, which additionally contributes to the difference.
\label{remark: discrepancy}
\end{remark}

In Setting A (See Fig.~\ref{fig: experiment setting 1}), we attach an elastic cable to the center of the ego quadrotor to generate a tension force. The control objective is to track a desired trajectory that passes through multiple waypoints. These waypoints are chosen such that the tension force vector in the world frame exhibits comparable components in three directions. The training of NeuroMHE is conducted by implementing Algorithm~\ref{alg: online} in numerical simulation with the same workstation as in Sec~\ref{subsec:expA} and \ref{subsec:expB}. In training, we numerically simulate the tension force vector using the model ${\bm f}_{\rm t}=-k\left ( l-l_{0} \right )\ast {\bm p}/\left \| \bm p \right \|$ where $k=50\ {\rm N/m}$ is the cable stiffness, $l_0=1.5\ {\rm m}$ is the cable's natural length, and $l$ is the cable's actual length that depends on the quadrotor's position $\bm p$. Given that the force estimation is the primary concern in the experiment, we modify the neural network's output to be $\boldsymbol \Theta =\left [ p_{1:6},\bar{\gamma}_1,r_{1:3},\bar{\gamma }_2,q_{1:3}  \right ]\in \mathbb R^{14}$, excluding the weightings related to the quadrotor's position, attitude, and angular velocity. We select the quadrotor's velocity as an input feature for the neural network, as it reflects the force change. The number of neurons is $20$ in both hidden layers, and the MHE horizon is $10$. Finally, under the same conditions, we also train DMHE and manually tune the Hurwitz matrix and the LPF's bandwidth used in $\mathcal{L}_1$-AC to $\bm A_{s}=-{\rm diag}\left ( \left [ 1,1,1 \right ] \right )$ and $\omega _{c}=5\ {\rm rad}/{\rm s}$, respectively. The trained NeuroMHE and DMHE, as well as the tuned $\mathcal{L}_1$-AC are deployed to the real quadrotor without extra tuning.

\begin{figure}[h]
\centering
\begin{subfigure}[b]{0.24\textwidth}
\centering
\includegraphics[width=1\textwidth]{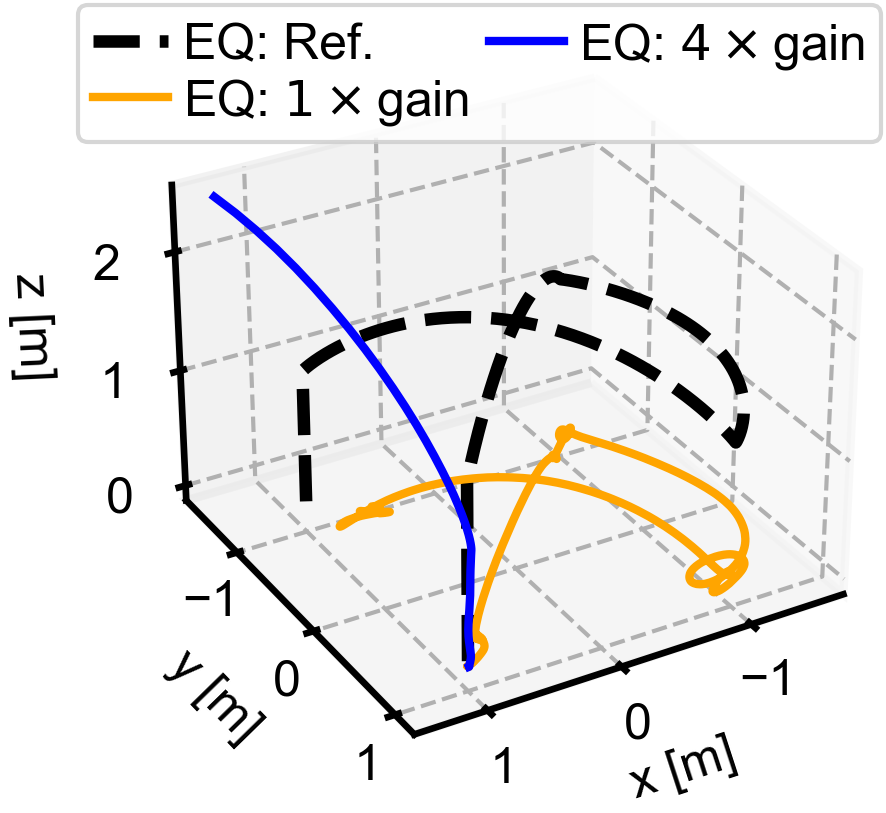}
\caption{\footnotesize Baseline controller.}
\label{fig:tracking baseline experiment}
\end{subfigure}
\hfill
\begin{subfigure}[b]{0.24\textwidth}
\centering
\includegraphics[width=1\textwidth]{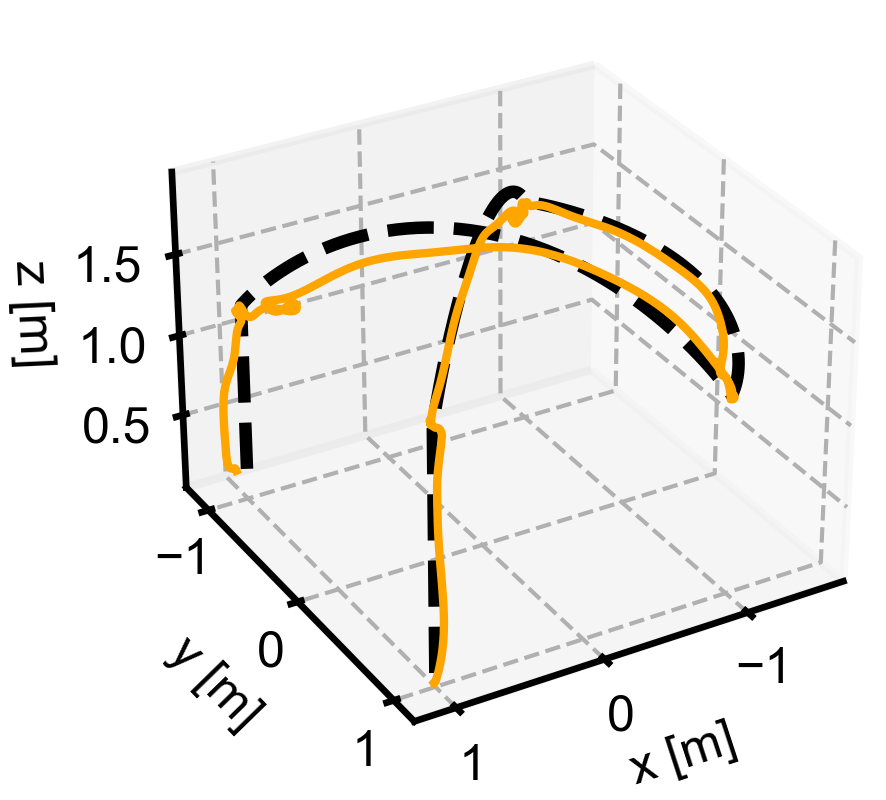}
\caption{\footnotesize NeuroMHE-based controller.}
\label{fig:tracking neuromhe experiment}
\end{subfigure}
\hfill
\begin{subfigure}[b]{0.24\textwidth}
\centering
\includegraphics[width=1\textwidth]{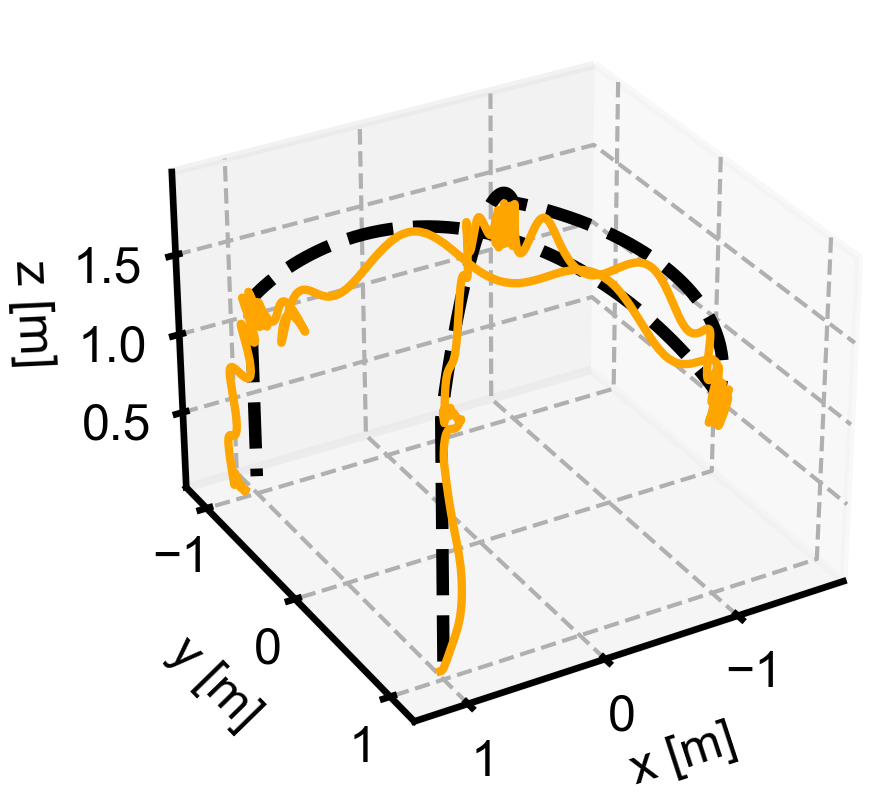}
\caption{\footnotesize DMHE-based controller.}
\label{fig:tracking dmhe experiment}
\end{subfigure}
\hfill
\begin{subfigure}[b]{0.24\textwidth}
\centering
\includegraphics[width=1\textwidth]{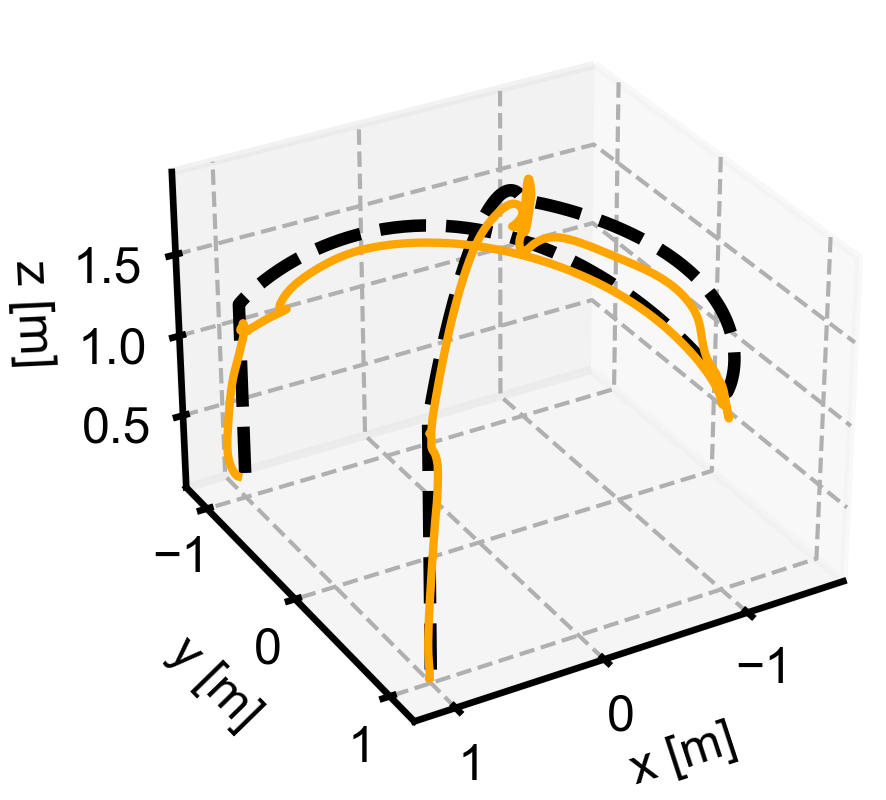}
\caption{\footnotesize $\mathcal{L}_1$-AC-based controller.}
\label{fig:tracking l1 experiment}
\end{subfigure}
\caption{\footnotesize Comparison between the tracking performance of NeuroMHE and that of DMHE, $\mathcal{L}_1$-AC, and the baseline controller in Setting A. The EQ stands for the ego quadrotor.}
\label{fig: path tracking experiment}
\end{figure}


\begin{figure*}[h]
\centering
\begin{subfigure}[b]{0.32\textwidth}
\centering
\includegraphics[width=1\textwidth]{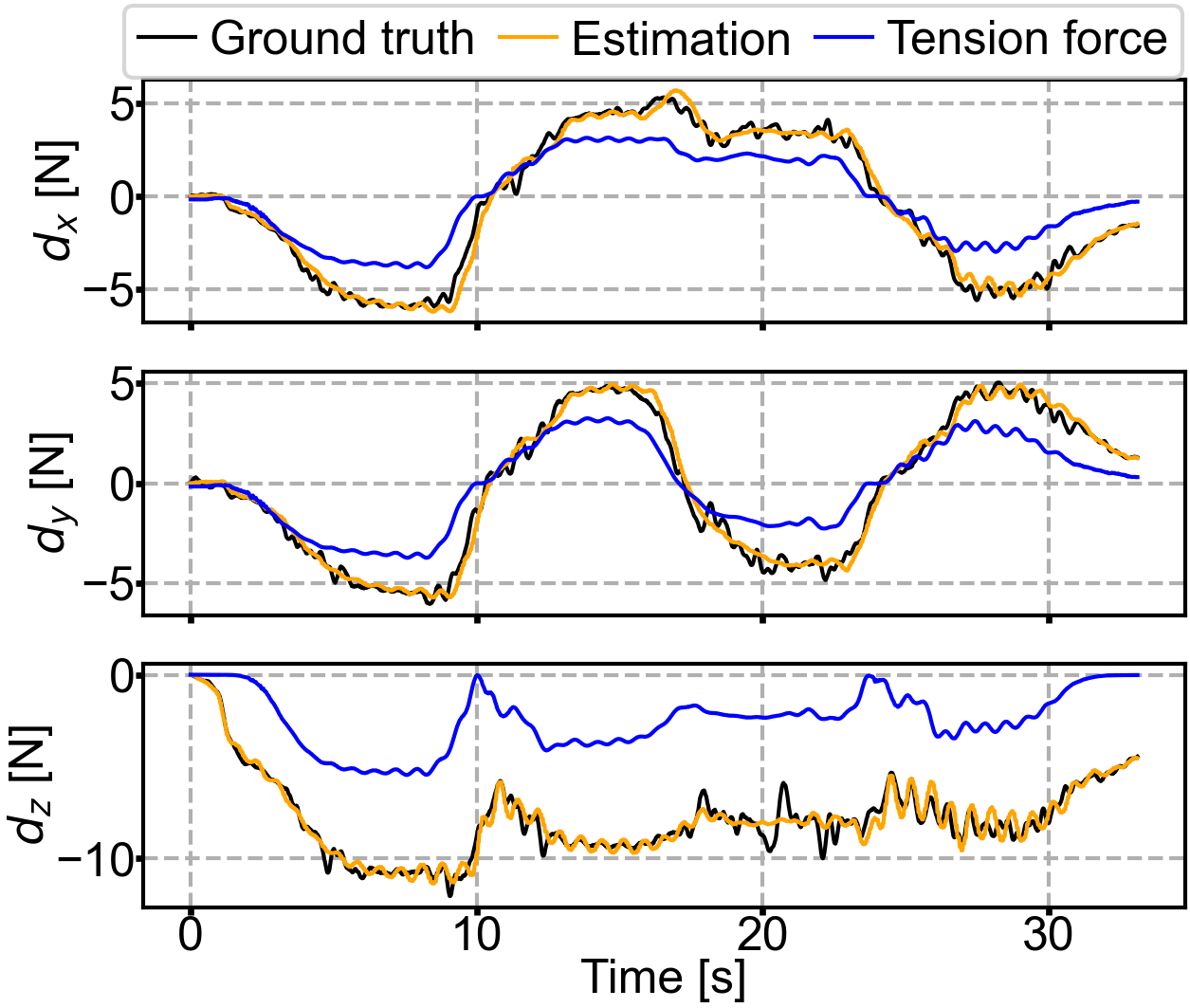}
\caption{\footnotesize NeuroMHE estimation performance.}
\label{fig:neuromhe estimation experiment}
\end{subfigure}
\hfill
\begin{subfigure}[b]{0.32\textwidth}
\centering
\includegraphics[width=1\textwidth]{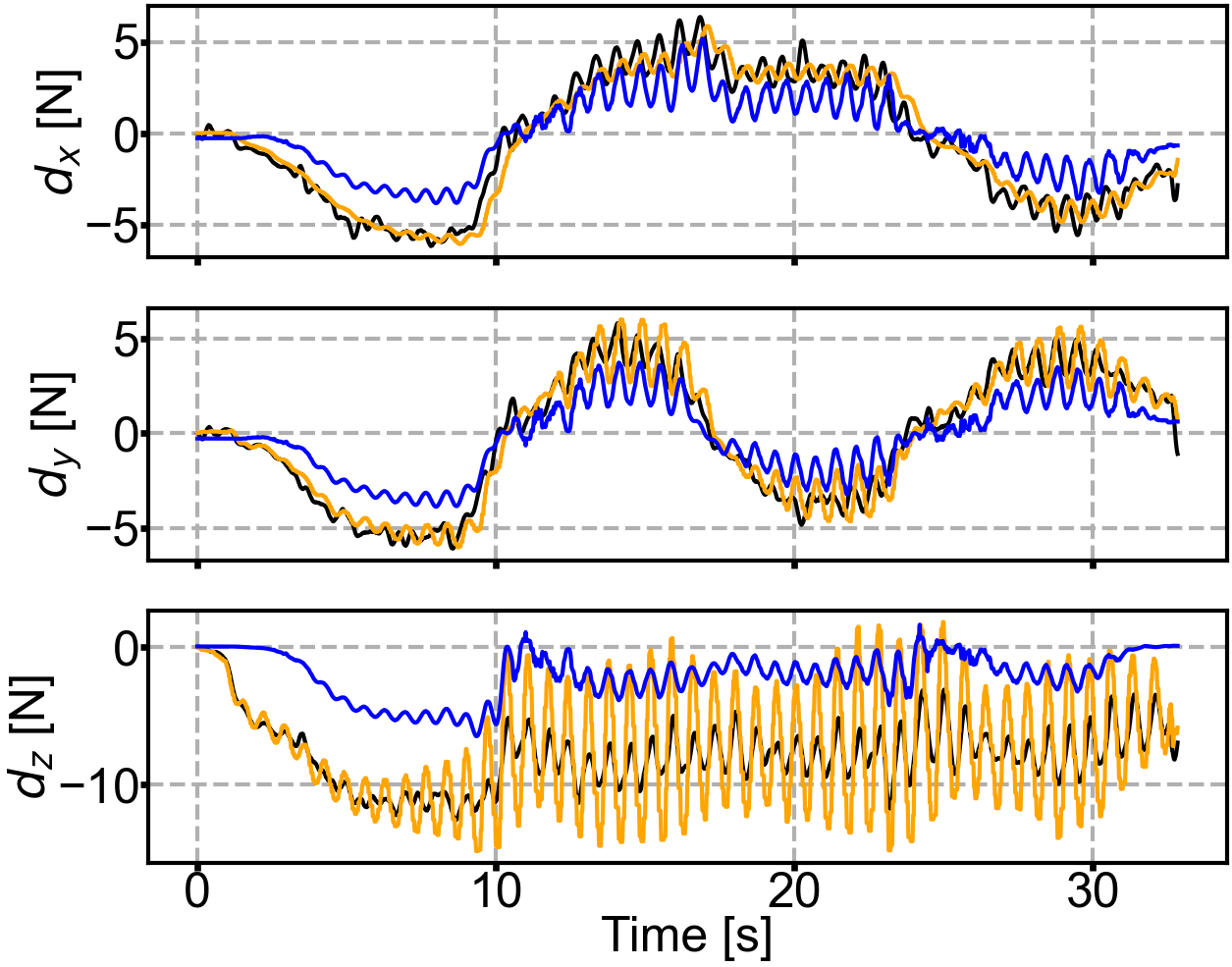}
\caption{\footnotesize DMHE estimation performance.}
\label{fig:dmhe estimation experiment}
\end{subfigure}
\hfill
\begin{subfigure}[b]{0.32\textwidth}
\centering
\includegraphics[width=1\textwidth]{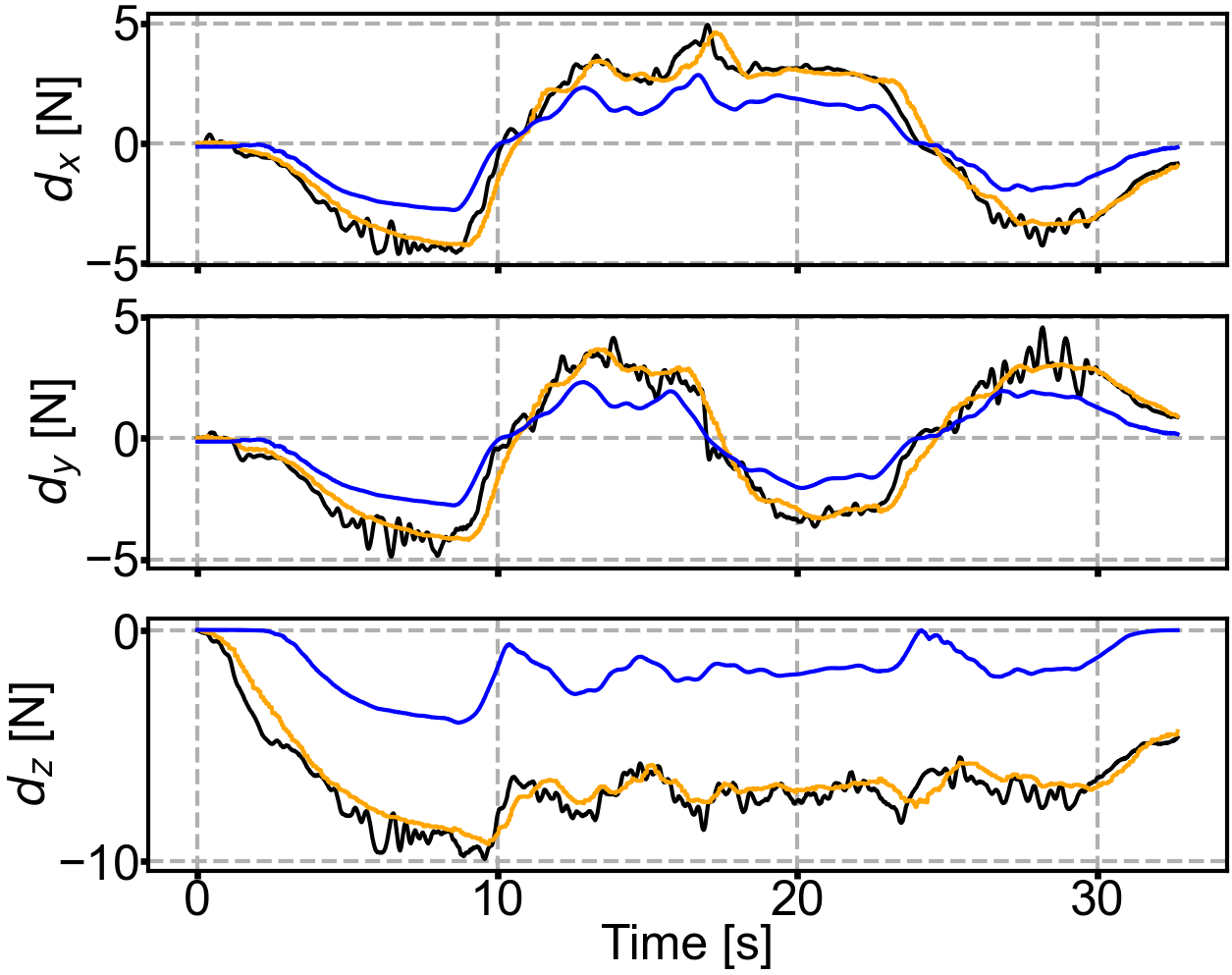}
\caption{\footnotesize $\mathcal{L}_1$-AC estimation performance.}
\label{fig:l1 estimation experiment}
\end{subfigure}
\caption{\footnotesize Comparison between the disturbance estimation performance of NeuroMHE and that of DMHE and $\mathcal{L}_1$-AC in Setting A. The components $d_x$, $d_y$, and $d_z$ are expressed in the ENU world frame. The ground truth disturbance $\bm d$ is calculated based on the quadrotor's position dynamics $\dot{\bm v}=-g{\bm e}_{3}+m^{-1}\left ( {\bm R}f_{\rm cmd}{\bm e}_{3}+{\bm d} \right )$. The true acceleration $\dot{\bm v}$ is obtained from the IMU and smoothed by a finite-impulse-response high-order low-pass filter with a cutoff frequency of $2\ {\rm Hz}$. The current rotation matrix $\bm R$ is obtained from the quaternions $\bm q$.  The primary purpose of providing the tension force is to illustrate the dynamic pattern of the external disturbance. Please refer to Remark~\ref{remark: discrepancy} for the explanation of the difference between the tension force and the ground truth disturbance. In particular, note that the quadrotor's attitude only experiences small perturbations relative to the hovering state during the flight. Hence, the difference between the tension force and the ground truth is more prominent in $z$-axis.}
\label{fig: disturbance estimation experiment}
\end{figure*}

Fig.~\ref{fig: path tracking experiment} shows a comparison of the tracking performance of all control methods under the tension effect. Notably, using the baseline controller alone results in significant tracking errors. In an attempt to mitigate these errors, we increase the control gain to four times its baseline value. However, this causes the quadrotor to become unstable as soon as the cable is taut, indicating extremely poor robustness to the state-dependent disturbance and leading to a severe crash\footnote{A video showing the crash can be found on \url{https://www.youtube.com/watch?v=H6NPRawJ74g}.}. In contrast, augmenting the baseline controller with NeuroMHE, DMHE, and $\mathcal{L}_1$-AC for the disturbance compensation substantially improves the tracking performance, despite using the baseline gain. Fig.~\ref{fig: disturbance estimation experiment} compares the disturbance estimation performance of NeuroMHE with that of DMHE and $\mathcal{L}_1$-AC. It is evident that the disturbance estimate of DMHE suffers from severe oscillations, particularly in the $z$ direction, when compared to that of NeuroMHE and $\mathcal{L}_1$-AC. Due to this poor estimation performance, the trajectory of the quadrotor using the DMHE-based controller is also highly oscillatory (See Fig.~\ref{fig:tracking dmhe experiment}). In addition, we observe in Fig.~\ref{fig:neuromhe estimation experiment} and Fig.~\ref{fig:l1 estimation experiment} that NeuroMHE exhibits smaller estimation time lag and error than $\mathcal{L}_1$-AC.

\begin{table}[h]
\fontsize{7.5}{7.5}\selectfont
\caption{Tracking and Estimation RMSEs in Setting A\label{table:rmse_exp1}}
\centering
\begin{threeparttable}[t]
\begin{tabular}{ c|c c c | c c c } 
\toprule[1pt]
\multirow{2} {*} {Method}  & ${d_{x}}$ & $d_y$ & $d_z$ & ${p_x}$ & ${p_y}$ & ${p_{z}}$\\
       & $\left[ {\rm N} \right]$ & $\left[ {\rm N} \right]$ & $\left[ {\rm N} \right]$ & $\left[ {\rm m} \right]$ & $\left[ {\rm m} \right]$ & $\left[ {\rm m} \right]$ \\
\midrule[0.5pt]
NeuroMHE  & $\bf 0.425$ & $\bf 0.412$ & $\bf 0.524$ & $\bf 0.081$ & $0.099$ & $\bf 0.048$ \\
DMHE  & $0.797$ & $0.823$ & $2.725$ & $0.124$ & $\bf 0.085$ & $0.133$ \\
$\mathcal{L}_1$-AC & $0.454$ & $0.547$ & $0.647$ & $\bf 0.081$ & $0.092$ & $0.131$ \\
Baseline & ${\rm N}/{\rm A}$ & ${\rm N}/{\rm A}$&${\rm N}/{\rm A}$& $0.175$ & $0.145$ & $1.217$ \\
\bottomrule[0.5pt]
\end{tabular}
\begin{tablenotes}[flushleft]
      \footnotesize
      \item {The tracking RMSEs of the baseline controller with a larger gain are not included in the table due to the resulting crash.}
\end{tablenotes}
\end{threeparttable}
\end{table}

These findings are further supported by the quantitative comparisons in terms of RMSEs, which are summarized in Table~\ref{table:rmse_exp1}. The data in the first three columns indicates that our method outperforms DMHE and $\mathcal{L}_1$-AC in all directions. Specifically, we achieve up to $80.8\%$ reduction in force estimation errors. On the other hand, the data in the last three columns reveals that all robust control methods are able to reduce the tracking RMSEs significantly when compared to the baseline controller. Moreover, comparing the tracking RMSE of NeuroMHE with that of DMHE and $\mathcal{L}_1$-AC, we observe that while NeuroMHE exhibits comparable tracking performance to these two methods in $x$ and $y$ directions, it reduces the tracking error by up to $63.9\%$ in the $z$ direction.

\begin{figure}[h]
	\centering
	{\includegraphics[width=0.8\linewidth]{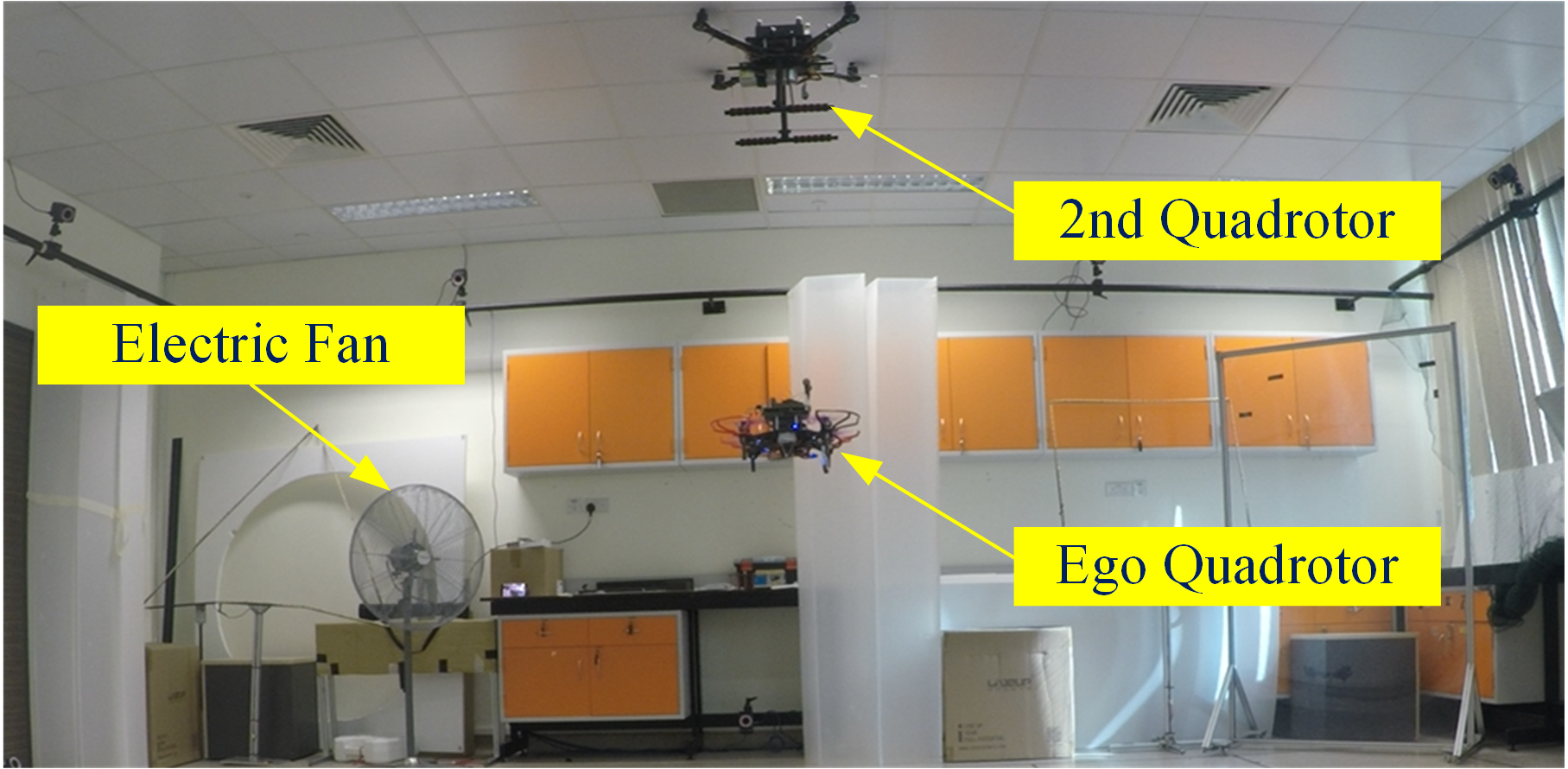}}
	\caption{\footnotesize Experimental setup of Setting B. The 2nd quadrotor with the PX4 controller hovers at about $1\ {\rm m}$ above the ego one to produce the downwash effect in the vertical direction. Simultaneously, the electric fan's turbulence generates aerodynamic disturbance forces primarily in the horizontal direction.}
	\label{fig: experiment setting 2}
\end{figure}

\begin{figure*}[h]
\centering
\begin{subfigure}[b]{0.32\textwidth}
\centering
\includegraphics[width=1\textwidth]{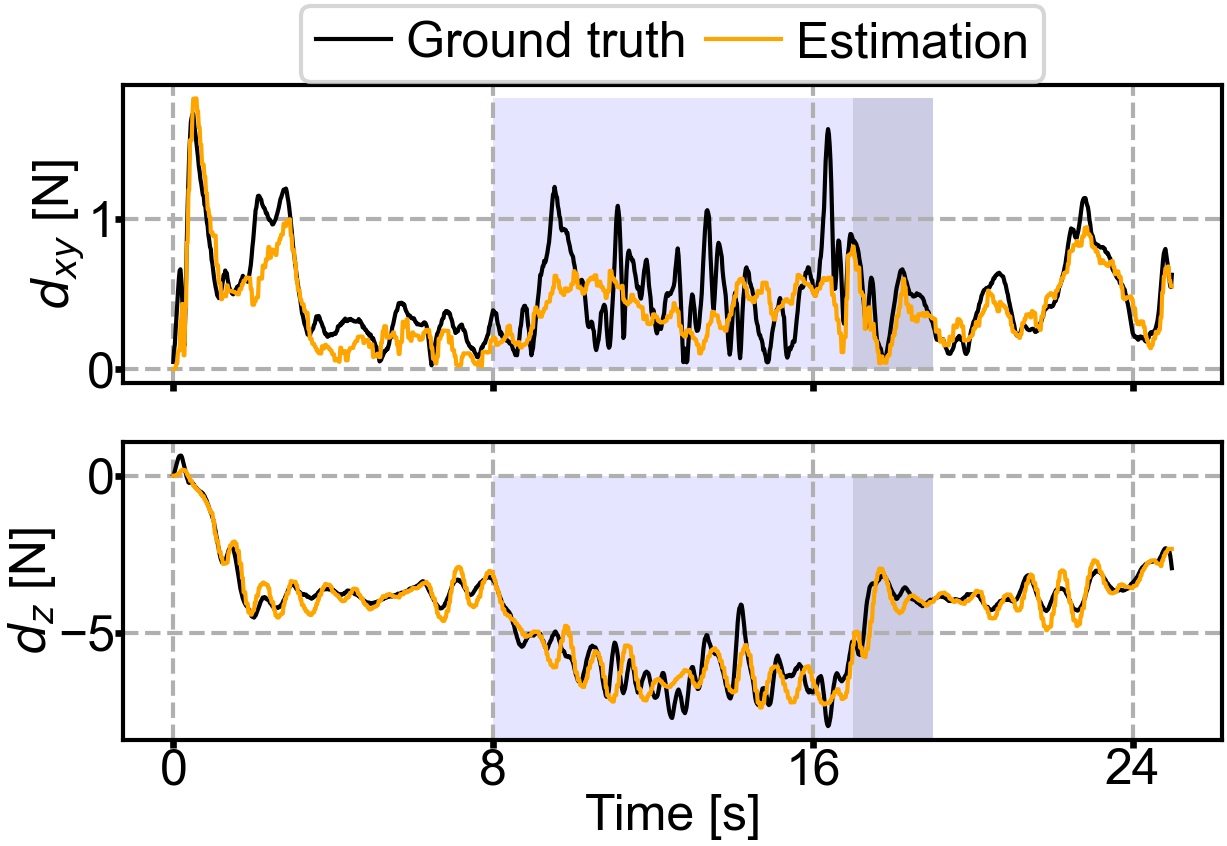}
\caption{\footnotesize NeuroMHE estimation performance.}
\label{fig:neuromhe estimation experiment2}
\end{subfigure}
\hfill
\begin{subfigure}[b]{0.32\textwidth}
\centering
\includegraphics[width=1\textwidth]{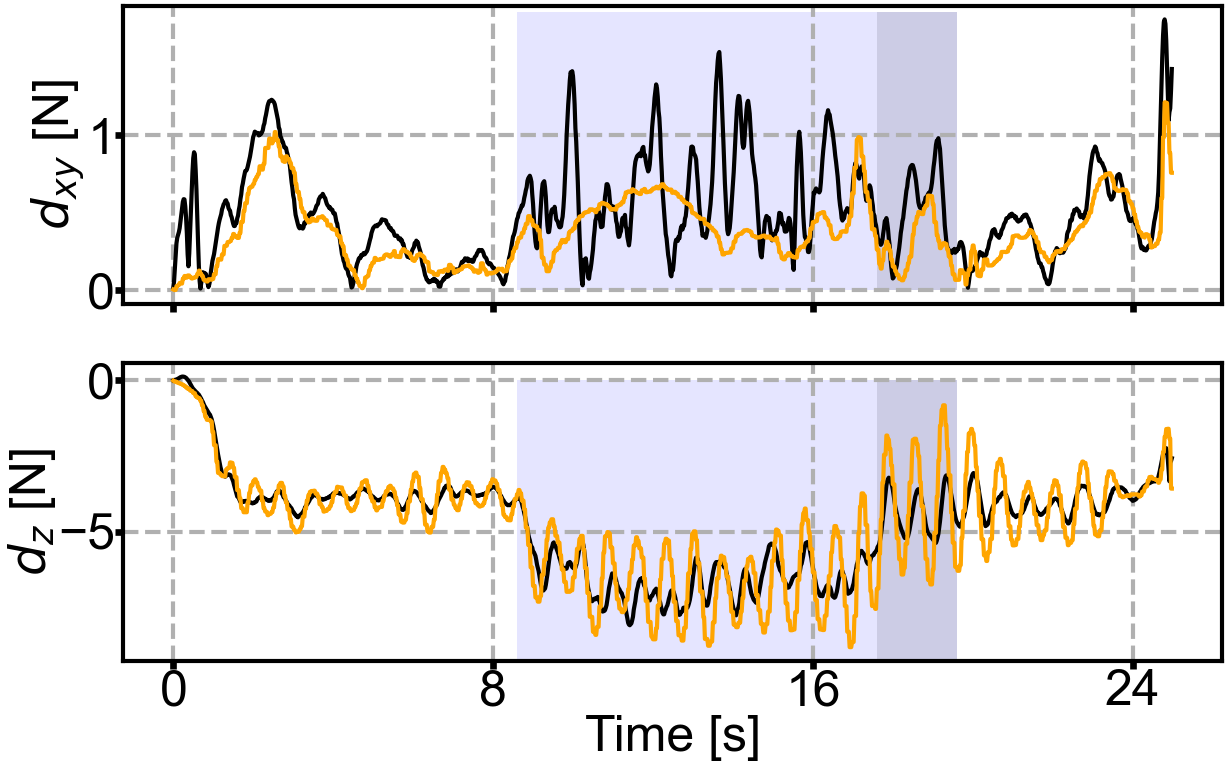}
\caption{\footnotesize DMHE estimation performance.}
\label{fig:dmhe estimation experiment2}
\end{subfigure}
\hfill
\begin{subfigure}[b]{0.32\textwidth}
\centering
\includegraphics[width=1\textwidth]{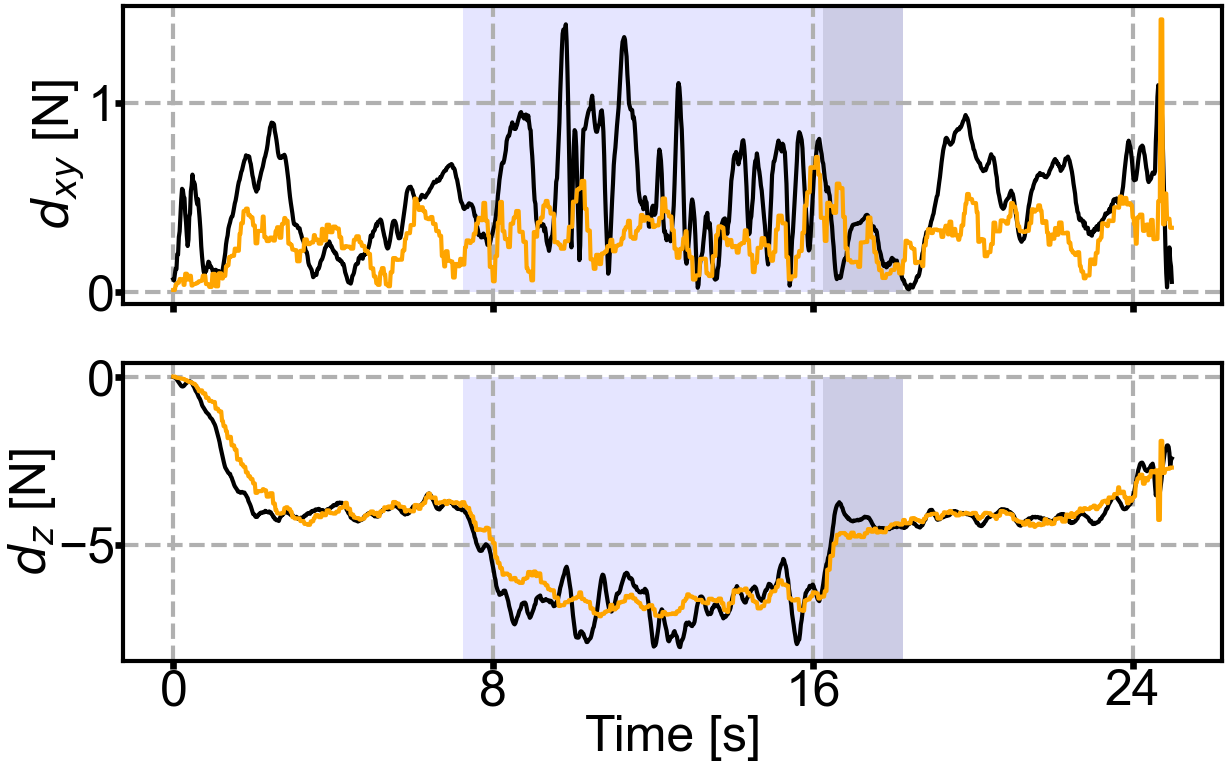}
\caption{\footnotesize $\mathcal{L}_1$-AC estimation performance.}
\label{fig:l1 estimation experiment2}
\end{subfigure}
\caption{\footnotesize Comparison between the disturbance estimation performance of NeuroMHE and that of DMHE and $\mathcal{L}_1$-AC in Setting B. The disturbance ground truth is obtained using the same method as in Setting A. In each subfigure, the left shadow block denotes the downwash stage, while the narrow right shadow block indicates the recovery stage. The time at which the ego quadrotor takes off is manually set, resulting in a slight variation in the starting time of the downwash effect.}
\label{fig: disturbance estimation experiment2}
\end{figure*}

\begin{figure}[h]
\centering
\begin{subfigure}[b]{0.24\textwidth}
\centering
\includegraphics[width=1\textwidth]{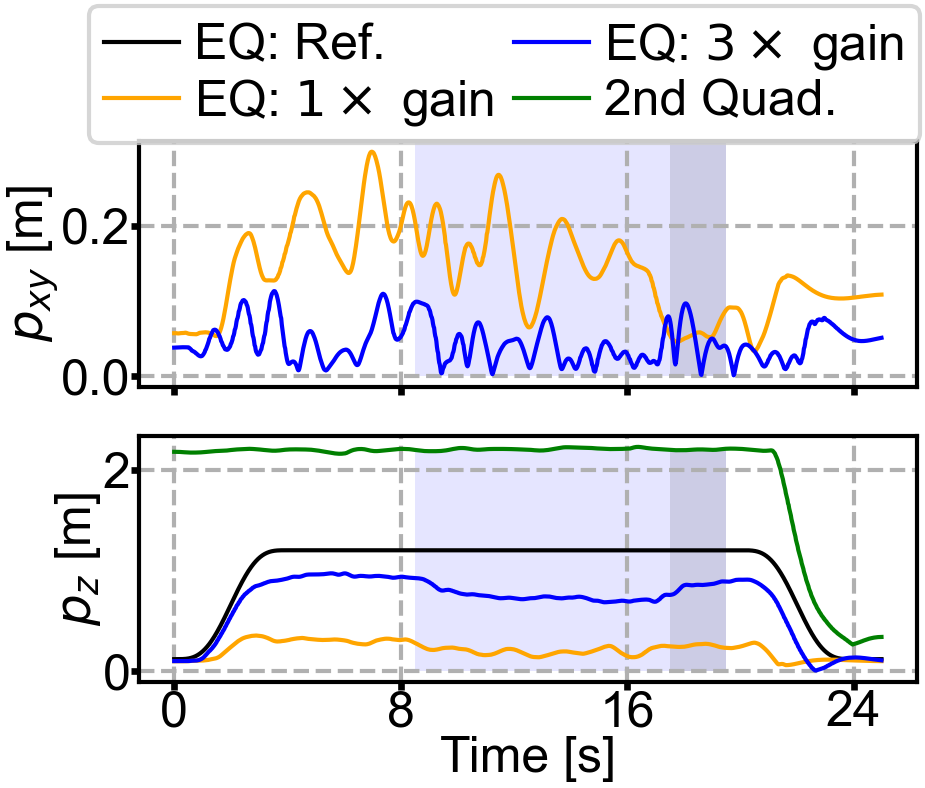}
\caption{\footnotesize Baseline controller.}
\label{fig:tracking baseline experiment2}
\end{subfigure}
\hfill
\begin{subfigure}[b]{0.24\textwidth}
\centering
\includegraphics[width=1\textwidth]{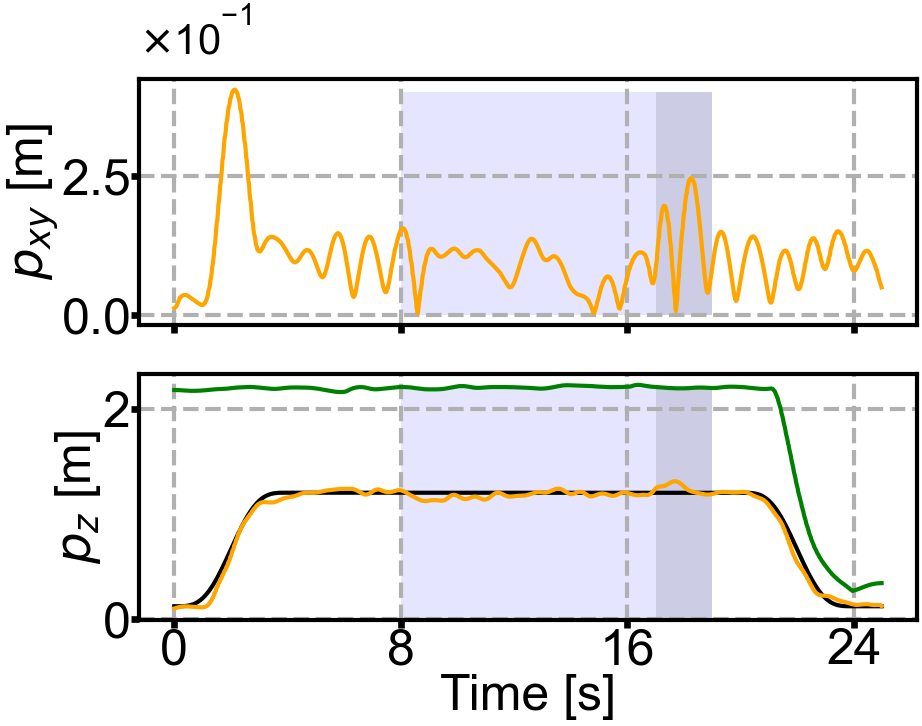}
\caption{\footnotesize NeuroMHE-based controller.}
\label{fig:tracking neuromhe experiment2}
\end{subfigure}
\hfill
\begin{subfigure}[b]{0.24\textwidth}
\centering
\includegraphics[width=1\textwidth]{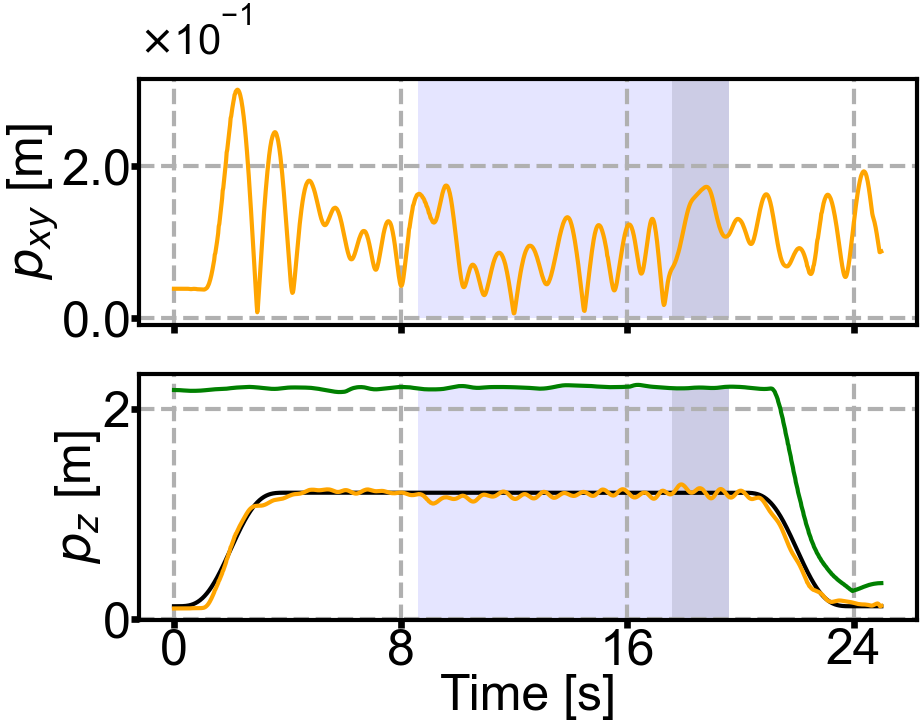}
\caption{\footnotesize DMHE-based controller.}
\label{fig:tracking dmhe experiment2}
\end{subfigure}
\hfill
\begin{subfigure}[b]{0.24\textwidth}
\centering
\includegraphics[width=1\textwidth]{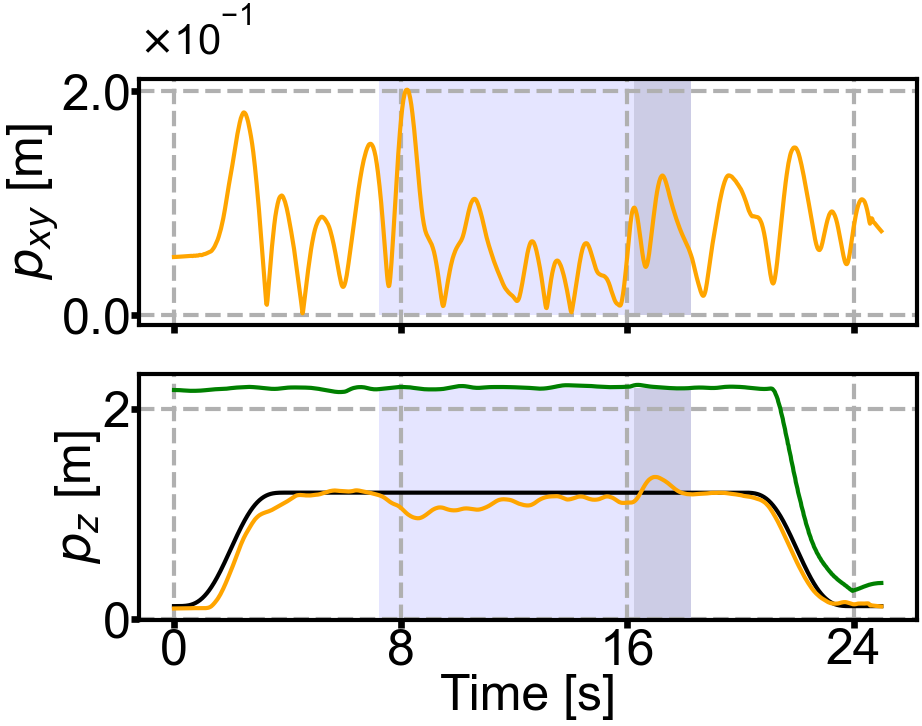}
\caption{\footnotesize $\mathcal{L}_1$-AC-based controller.}
\label{fig:tracking l1 experiment2}
\end{subfigure}
\caption{\footnotesize Comparison between the stabilization performance of NeuroMHE and that of DMHE, $\mathcal{L}_1$-AC, and the baseline controller in Setting B.}
\label{fig: tracking experiment2}
\end{figure}

The subsequent experiment conducted in Setting B aims to evaluate the robustness of the proposed method in challenging aerodynamic conditions. All controllers are the same as those used in Setting A. The experimental setup, shown in Fig.~\ref{fig: experiment setting 2}, involves the generation of external aerodynamic disturbances along the $x$, $y$, $z$ axes via an electric fan and the 2nd quadrotor. The experiment begins with the 2nd quadrotor hovering $2.2\ {\rm m}$ above its starting point $\left [ 1.5;1.5;0 \right ]\ {\rm m}$. Next, the ego quadrotor ascends to hover at $1.2\ {\rm m}$ above the origin, followed by the 2nd quadrotor flying towards it, hovering $1\ {\rm m}$ above it for $9\ {\rm s}$, and subsequently departing to land. This maneuver generates a downwash disturbance force in a square-wave-like pattern, as shown in Fig.~\ref{fig: disturbance estimation experiment2}.

\begin{table}[h]
\fontsize{7.5}{7.5}\selectfont
\caption{Performance Comparisons in Downwash Stage\label{table:rmse_exp2}}
\centering
\begin{threeparttable}[t]
\begin{tabular}{ c|c c | c c c c } 
\toprule[1pt]
\multirow{2} {*} {Method}  & $e_{xy}^{\max}$ &${\bar e}_{xy}$  & $e_{z}^{\max}$ & $\bar{e}_z^{\rm ss}$ & $\sigma_{z}^{\rm ss}$ & $d_{z}$\\
       & $\left[ {\rm m} \right]$ & $\left[ {\rm m} \right]$ & $\left[ {\rm m} \right]$ & $\left[ {\rm m} \right]$ & $\left[ {\rm m} \right]$ & $\left[ {\rm N} \right]$ \\
\midrule[0.5pt]
NeuroMHE  & $ 0.156$ & $0.092$  & $\bf 0.079$ & $\bf 0.020$   & $\bf 0.018$ & $\bf 0.596$ \\
DMHE  & $0.174$ & $0.094$ & $0.103$ & $0.026$ & $ 0.023$ & $ 1.086$ \\
$\mathcal{L}_1$-AC & $0.201$ & $0.074$ & $0.244$ & $0.061$ & $0.020$ & $0.621$ \\
Baseline & $0.267$ & $0.165$&$1.062$& $1.009$ & $0.036$ & ${\rm N}/{\rm A}$ \\
$3\times$Baseline  & $\bf 0.099$ & $\bf 0.045$ & $ 0.516$ & $0.485$ & $0.020$ & ${\rm N}/{\rm A}$\\
\bottomrule[0.5pt]
\end{tabular}
\begin{tablenotes}[flushleft]
      \footnotesize
      \item {In the table, $e_{xy}^{\max}$ and $e_{z}^{\max}$ are the maximum horizontal and vertical tracking errors, $\bar{e}_{xy}$ is the RMSE of horizontal error, $\bar{e}_{z}^{\rm ss}$ is the RMSE of vertical error in steady-state, $\sigma_z^{\rm ss}$ is the standard deviation of vertical error in steady-state, the data in $d_z$ column denotes the RMSE of vertical disturbance estimation. {$3\times$Baseline } uses a larger control gain which is three times the baseline value.}
\end{tablenotes}
\end{threeparttable}
\end{table}

Our evaluation of the estimation and control performance consists of two stages: the downwash stage and the recovery stage after the second quadrotor flies away. Fig.~\ref{fig: tracking experiment2} illustrates the stabilization performance of all control methods, and quantitative comparisons are summarized in Table~\ref{table:rmse_exp2}. During the downwash stage, the baseline controller yields significant tracking errors in all directions. A straightforward way to reduce the tracking errors is to increase the control gains (i.e., "high-gain" control). However, its improved performance is often at the cost of reduced system stability margin~\cite{tsypkin1999high}, as exemplified by the crash in Setting A. Practically, it can only handle relatively small disturbances before the system becomes unstable. To further push the performance boundary against large disturbances, disturbance estimation and compensation are typically needed. Note that the $3\times$Baseline controller outperforms all robust controllers only in horizontal tracking. Its vertical tracking error remains considerably large with a steady-state value of about $0.5\ {\rm m}$. This is exactly due to the fact that the horizontal disturbance is relatively small (about $0.5\ {\rm N}$, see Fig.~\ref{fig: disturbance estimation experiment2}) and more amenable to high-gain control, whereas the substantial vertical disturbance is beyond the capability of stable high-gain control (about $7\ {\rm N}$, see Fig.~\ref{fig: disturbance estimation experiment2}). Hence, we refrain from amplifying the baseline control gain by more than three times to prevent the potential instability. By comparing this experiment setting with Setting A (where both horizontal and vertical directions have large disturbances), we hope to elucidate the limitation of high-gain control and the practical usage of disturbance estimation-based robust control. Moreover, the unsatisfactory horizontal tracking performance of all the robust controllers is due to their inadequate estimation accuracy, with the best among them achieving an RMSE of around $0.2\ {\rm N}$ (NeuroMHE). This is due to the relatively low running frequency (For fair comparisons, we run all the controllers at the same frequency of $25\ {\rm Hz}$). It is possible to further improve the estimation accuracy by increasing the running frequency, which can lead to improved tracking performance.

In the vertical direction, all the robust controllers benefit from the disturbance compensation using the estimation, reducing the steady-state height tracking error to within $0.06\ {\rm m}$. In particular, our NeuroMHE-based method achieves the most accurate height tracking performance with the smallest oscillation among all methods, as evidenced by the values of $\bar{e}_{z}^{\rm ss}$, $e_{z}^{\max}$, and $\sigma_z^{\rm ss}$. This is due to the accurate disturbance estimation of NeuroMHE, which outperforms DMHE and $\mathcal{L}_1$-AC by up to $45.1\%$. During the recovery stage, the NeuroMHE-base, DMHE-based, and $\mathcal{L}_1$-AC-based robust controllers result in the height overshoot of $0.110\ {\rm m}$, $0.079\ {\rm m}$, and $0.149\ {\rm m}$, respectively, relative to the desired height of $1.2\ {\rm m}$. Despite the slightly larger overshoot produced by NeuroMHE compared to DMHE, the height trajectory using NeuroMHE is much smoother, indicating better stability.

Overall, the experiment results demonstrate:
\begin{enumerate}
    \item Our method outperforms the state-of-the-art methods in terms of both control performance and adaptation to different flight scenarios, exhibiting superior simulation-to-real transfer capability;
    \item The proposed NeuroMHE can be easily integrated with the widely-used PX4 control firmware and effectively robustify an existing baseline controller against various external and internal disturbances.

\end{enumerate}

\section{Discussion}\label{section: discussion}
NeuroMHE features auto-tuning and a portable neural network for modeling the online adaptive MHE weightings. It outperforms the state-of-the-art NeuroBEM in terms of training efficiency and estimation accuracy across various agile flight trajectories. NeuroMHE-based robust controller exhibits superior trajectory-tracking performance and adaptation to different scenarios over the state-of-the-art $\mathcal{L}_1$-AC-based controller, particularly under state-dependent disturbances. The recently proposed PI-TCN \cite{saviolo2022physics} improves over NeuroBEM using more advanced learning techniques. Both methods try to learn an accurate quadrotor dynamics model. However, these methods will fail when the model is subject to changes, such as load variation or thrust loss due to battery depletion. In contrast, NeuroMHE is able to accurately account for such unmodeled dynamics by referring to the nominal model, and hence compensate for a variety of external and internal disturbances. 

On the other hand, NeuroMHE demands more computational resources for real-time estimation than the other three state-of-the-art methods due to its reliance on solving a nonlinear optimization problem. Our existing onboard hardware can only run it at about $25\ {\rm Hz}$ to support outer-loop position control in real-world flight. Torque compensation is in the inner-loop attitude control, demanding a much higher running frequency of hundreds of Hertz (e.g., $400\ {\rm Hz}$~\cite{wu2023l1quad}). To achieve it in real-world flight, more powerful hardware and particularly optimized numerical solvers would be needed. NeuroBEM and PI-TCN can be employed in model predictive control to explicitly account for future disturbances. It is nevertheless impractical to utilize NeuroMHE for such predictive control. Finally, $\mathcal{L}_1$-AC is the only method that has theoretical stability guarantee, while the stability of nonlinear MHE and neural network based estimators still remains largely an open problem.


\section{Conclusion} \label{section: conclusion}
This paper proposed a novel estimator NeuroMHE that can accurately estimate disturbances and adapt to different flight scenarios. Our critical insight is that NeuroMHE can automatically tune the key parameters generated by a portable neural network online from the trajectory tracking error to achieve optimal performance. At the core of our approach is the computationally efficient method to obtain the analytical gradients of the MHE estimates with respect to the weightings, which explores a recursive form using a Kalman filter. We have shown that the proposed NeuroMHE enjoys efficient training, fast online environment adaptation, and improved disturbance estimation performance over the state-of-the-art estimator and adaptive controller, via extensive simulations using both real and synthetic datasets. We have also conducted physical experiments on a real quadrotor to demonstrate that NeuroMHE can effectively robustify a baseline flight controller against various challenging disturbances.
Our future work includes developing more efficient training algorithms and analyzing the stability of NeuroMHE.

\appendix

\subsection{Approximation of $\frac{{\partial {{{\hat{ x}}_{t - N}}}}}{{\partial { \theta} }}$}\label{appendix:approximation}
As defined in (\ref{eq:mhe}), $\hat{\bm x}_{t-N}$ is the MHE estimate ${{\hat {\bm x}}_{t - N\left| {t-1} \right.}}$ of $\bm x_{t-N}$ made at $t-1$. Therefore, at the current step $t$, the gradient $\frac{{\partial {{{\hat{ \bm x}}_{t - N}}}}}{{\partial {\boldsymbol \theta} }}$ can be computed by the chain rule:
\begin{equation}
    \frac{\partial \hat{\bm x}_{t-N}}{\partial \boldsymbol\theta_{t} }=\frac{\partial \hat{\bm x}_{t-N\left| {t-1} \right.}}{\partial \boldsymbol\theta _{t-1}}\frac{\partial \boldsymbol\theta _{t-1}}{\partial \boldsymbol\theta _{t}}.
    \label{eq:chain rule of arrival gradient}
\end{equation}
The first gradient in the right-hand side (RHS) of (\ref{eq:chain rule of arrival gradient}) can be obtained from the gradient trajectory $\left \{ \frac{\partial \hat{\bm x}_{k\left| {t-1} \right.}}{\partial \boldsymbol \theta } \right \}_{k=t-N}^{t-1}$ made at $t-1$, as will be detailed in Lemma~\ref{lemma:analytical gradient}. 

Next, we approximate the second gradient in the RHS for two cases: 1) direct updating of $\boldsymbol\theta$ using gradient descent, and 2) modelling of $\boldsymbol\theta$ using a neural network. For Case 1, a solution for $\frac{d \boldsymbol\theta _{t-1}}{d \boldsymbol\theta _{t}}$ can be obtained from the one-step update: $\boldsymbol \theta _{t}=\boldsymbol \theta _{t-1}-\alpha \triangledown_{\theta }L$ as:
\begin{equation}
    \frac{d \boldsymbol\theta _{t-1}}{d \boldsymbol\theta _{t}}=\left ( \bm I-\alpha \mathcal{H}_{\theta }L \right )^{-1},
    \label{eq:second gradient solution}
\end{equation}
where $\alpha$ is the learning rate and $\mathcal{H}_{\theta }L$ is the Hessian of the loss $L$ with respect to $\boldsymbol \theta$. Since $\alpha$ is typically very small, the RHS of (\ref{eq:second gradient solution}) approaches the identity matrix $\bm I$. This results in the desired approximation:
\begin{equation}
    \frac{\partial \hat{ \bm x}_{t-N}}{\partial \boldsymbol\theta_{t} }\approx \frac{\partial \hat{\bm x}_{t-N\left| {t-1} \right.}}{\partial \boldsymbol\theta _{t-1}}.
    \label{eq:desired gradient approximation}
\end{equation}

For Case 2, since the input features of the neural network (e.g., the quadrotor states) are typically not subject to sudden changes, the gradient $\frac{d \boldsymbol\theta _{t-1}}{d \boldsymbol\theta _{t}}$ can still approach the identity matrix $\bm I$. Thus, the approximation (\ref{eq:desired gradient approximation}) holds for the output of the network. 


\subsection{Proof of Lemma~\ref{lemma: auxiliary MHE}}\label{appendix:lemma1}
For the auxiliary MHE system (\ref{eq: auxiliary MHE}), we define the following Lagrangian
\begin{equation}
    {{\cal L}_2} = \frac{1}{2}{\rm{Tr}}\left\| {{{ {\bm X}}_{t - N}} - \hat{\bm X}_{t-N}} \right\|_{\bm P}^2 + {\bar {\cal L}_2},
    \label{eq: Lagrangian for auxiliary MHE}
\end{equation}
where
\begin{equation}
    \begin{aligned}
{{\bar {\cal L}}_2}& ={\rm{Tr}}\sum\limits_{k = t - N}^t {\left( {\frac{1}{2}{\bm X}_{k}^T\bar {\bm L}_k^{xx}{{{\bm X}}_{k}} + {{\bm W}_{k}}^T {\bm L}_k^{w x} {\bm X}_{k} } \right)} \\
&\quad + {\rm{Tr}}\sum\limits_{k = t - N}^{t - 1} {\left( {\frac{1}{2}{\bm W}_{k}^T {\bm L}_k^{w w}{{\bm W}_k} +{{{\left( {{\bm L}_k^{w \theta }} \right)}^T}{{\bm W}_{k}}} }\right)}\\
&\quad + {\rm{Tr}}\sum\limits_{k = t - N}^{t} {\left( {{\left( {{\bm L}_k^{x\theta }} \right)}^T} {{{\bm X}}_{k}}  \right)} \\
&\quad + {\rm{Tr}}\sum\limits_{k = t - N}^{t - 1} {\bm \Lambda _k^T\left( {{{{\bm X}}_{k + 1}} - {\bm F_k}{{ {\bm X}}_{k}} - {\bm G_k}{{\bm W}_{k}}} \right)}.
\end{aligned}
\nonumber
\end{equation}
The optimal estimates $\hat{\bm {\mathop{ {\rm X}}\nolimits}}$ and $\hat{\bm {\mathop{ {\rm W}}\nolimits}}$, together with the new optimal dual variables  $\bm \Lambda^*  = \left\{ {{\bm \Lambda^* _k}} \right\}_{k = t - N}^{t - 1}$, satisfy the following KKT conditions:
\begin{subequations}
\begin{align}
\begin{split}
{\nabla _{{\hat X}_{t - N\left| t \right.}}}{{\cal { L}}_{2}} & = \left( {\bm P + \bar {\bm L}_{t - N}^{xx}} \right){{\hat {\bm X}}_{t - N\left| t \right.}} - {\bm P}{\hat{\bm X}_{t-N}} + {\bm L}_{t - N}^{x\theta }\\
 &\quad+ {\bm L}_{t - N}^{xw}{\hat{\bm W}_{t-N\left| t \right.}} - {\bm F}_{t - N}^T{\bm \Lambda^* _{t - N}} = \bm 0,
\end{split}
\label{eq:auxiliary kkt boundary}\\
\begin{split}
{\nabla _{{\hat X}_{k\left| t \right.}}}{{\cal {L}}_{2}} & = \bar {\bm L}_k^{xx}{{\hat {\bm X}}_{k\left| t \right.}} + {\bm L}_k^{xw}{\hat{\bm W}_{k\left| t \right.}} - {\bm F}_k^T{\bm \Lambda^* _k} + {\bm \Lambda^* _{k - 1}}\\
&\quad + {\bm L}_k^{x\theta } = \bm 0,\ k = t - N + 1, \cdots ,t, 
\end{split}
\label{eq:auxiliary kkt X}\\
\begin{split}
{\nabla _{{\hat W}_{k\left| t \right.}}}{{\cal {L}}_{2}} & = {\bm L}_k^{w x}{{\hat {\bm X}}_{k\left| t \right.}} + {\bm L}_k^{w w}{\hat{\bm W}_{k\left| t \right.}} - {\bm G}_k^T{\bm \Lambda^* _k}\\
&\quad + {\bm L}_k^{w \theta } = \bm 0,\ k = t - N, \cdots ,t - 1,
\end{split}
\label{eq:auxiliary kkt W}\\
\begin{split}
{\nabla _{\Lambda_k^*} }{{\cal {L}}_{2}} & = {{\hat {\bm X}}_{k + 1\left| t \right.}} - {\bm F_k}{{\hat {\bm X}}_{k\left| t \right.}}- {\bm G_k}{\hat{\bm W}_{k\left| t \right.}} = \bm 0,\\
&\quad \ k = t - N, \cdots ,t - 1.
\end{split}
\label{eq:auxiliary kkt Lambda}
\end{align}
\label{eq: auxiliary KKT appendix}%
\end{subequations}
The above equations (\ref{eq: auxiliary KKT appendix}) are the same as the differential KKT conditions (\ref{eq:differential KKT}), and thus (\ref{eq: equivalent gradient}) holds. This completes the proof.

\subsection{Proof of Lemma~\ref{lemma:analytical gradient}}\label{appendix:lemma2}
We establish a proof by induction to demonstrate that if (\ref{eq: forward state}) is satisfied by ${\hat {\bm X}_{k\left| t \right.}}$, it will also hold true for ${\hat {\bm X}_{k+1\left| t \right.}}$. First, we solve for $\hat{\bm W}_{k\left| t \right.}$ from (\ref{eq:auxiliary kkt W}) as 
\begin{equation}
    \hat{\bm W}_{k|t}=\left ( {\bm L}_{k}^{ww} \right )^{-1}\left ( {\bm G}_{k}^{T} {\boldsymbol\Lambda} _{k}^{*}-{\bm L}_{k}^{wx}\hat{\bm X}_{k|t}-{\bm L}_{k}^{w\theta }\right ).
    \label{eq:solution for W}
\end{equation}
After substituting $\hat{\bm W}_{k\left| t \right.}$ in (\ref{eq:auxiliary kkt X}) and (\ref{eq:auxiliary kkt Lambda}), we obtain
\begin{equation}
    {\bm \Lambda^* _{k - 1}} = {\bar {\bm F}^{T}_k}{\bm \Lambda^* _k} + {\bm S_k}{\hat {\bm X}_{k\left| t \right.}} + {\bm T_k},
    \label{eq:lambda without W}
\end{equation}
and
\begin{equation}
    {\hat {\bm X}_{k + 1\left| t \right.}} = {\bar {\bm F}_k}{\hat {\bm X}_{k\left| t \right.}} - {\bm G_k}{\left( { {\bm L}_k^{w w }} \right)^{ - 1}}{\bm L}_k^{w \theta } + {\bm G_k}{\left( { {\bm L}_k^{w w }} \right)^{ - 1}}{\bm G}_k^T{\bm \Lambda^* _k},
    \label{eq:state prediction without W}
\end{equation}
respectively, where $\bar{\bm F}_{k}={\bm F}_{k}-{\bm G}_{k}\left ( {\bm L}_{k}^{ww} \right )^{-1}{\bm L}_{k}^{wx}$, ${\bm S}_{k}={\bm L}_{k}^{xw}\left ( {\bm L}_{k}^{ww} \right )^{-1}{\bm L}_{k}^{wx}-\bar{\bm L}_{k}^{xx}$, and $\bm T_{k}={\bm L}_{k}^{xw}\left ( {\bm L}_{k}^{ww} \right )^{-1}{\bm L}_{k}^{w\theta }-{\bm L}_{k}^{x\theta }$.

Suppose that ${\hat {\bm X}_{k\left| t \right.}}$ satisfies (\ref{eq: forward state}) for $k \in \left[ {t - N,t-1} \right]$. Then, from (\ref{eq:state prediction without W}), we have
\begin{equation}
    \begin{aligned}
{{\hat {\bm X}}_{k + 1\left| t \right.}} & = {{\bar {\bm F}}_k}{{\hat {\bm X}}_{k\left| k \right.}^{\rm KF}} - {\bm G_k}{\left( { {\bm L}_k^{w w }} \right)^{ - 1}}{\bm L}_k^{w \theta }\\
 &\quad+ \left[ {{{\bar {\bm F}}_k}{\bm C_k}\bar {\bm F}_k^T + {\bm G_k}{{\left( { {\bm L}_k^{w w }} \right)}^{ - 1}}\bm G_k^T} \right]{\bm \Lambda^* _k}.
\end{aligned}
\label{eq:state prediction with kalman estimate}
\end{equation}
Using (\ref{eq: kalman filter prediction}) for $k+1$, we can simplify (\ref{eq:state prediction with kalman estimate}) to 
\begin{equation}
    {\hat {\bm X}_{k + 1\left| t \right.}} = {\hat {\bm X}_{k + 1\left| k \right.}} + \left[ {{{\bar {\bm F}}_k}{\bm C_k}\bar {\bm F}_k^T + {\bm G_k}{{\left( { {\bm L}_k^{w w }} \right)}^{ - 1}}\bm G_k^T} \right]{\bm \Lambda^* _k}.
    \label{eq:simplified state prediction in general case}
\end{equation}
Substituting ${\bm P_{k + 1}}$ from (\ref{eq: kalman filter covariance}) and $\bm \Lambda^*_k$ from (\ref{eq:lambda without W}) for $k+1$, we obtain
\begin{equation}
    \begin{aligned}
{{\hat {\bm X}}_{k + 1\left| t \right.}} & = {{\hat {\bm X}}_{k + 1\left| k \right.}} + {\bm P_{k + 1}}{\bm T_{k + 1}} + {\bm P_{k + 1}}{\bm S_{k + 1}}{{\hat {\bm X}}_{k + 1\left| t \right.}}\\
 &\quad+ {\bm P_{k + 1}}{{\bar {\bm F}}_{k + 1}}^T{\bm \Lambda^* _{k + 1}}.
\end{aligned}
\label{eq:simplified state prediction 2 in general case}
\end{equation}
The above equation is equivalent to the following form:
\begin{equation}
    \begin{aligned}
\left( {\bm I - {\bm P_{k + 1}}{\bm S_{k + 1}}} \right){{\hat {\bm X}}_{k + 1\left| t \right.}} & = \left( {\bm I - {\bm P_{k + 1}}{\bm S_{k + 1}}} \right){{\hat {\bm X}}_{k + 1\left| k \right.}}\\
&\quad + {\bm P_{k + 1}}{\bm S_{k + 1}}{{\hat {\bm X}}_{k + 1\left| k \right.}} \\
& \quad+ {\bm P_{k + 1}}{\bm T_{k + 1}}+ {\bm P_{k + 1}}{{\bar {\bm F}}^{T}_{k + 1}}{\bm \Lambda^* _{k + 1}}.
\end{aligned}
\label{eq:equivalent state prediction general case}
\end{equation}
Multiplying both sides by $\left( {\bm I - {\bm P_{k + 1}}{\bm S_{k + 1}}} \right)^{-1}$, and substituting $\bm C_{k+1}$ from (\ref{eq: kalman filter covariance update}) and ${\hat {\bm X}_{k + 1\left| {k + 1} \right.}^{\rm KF}}$ from (\ref{eq: kalman filter estimate}) for $k+1$, we obtain the desired relation as below:
\begin{equation}
    {\hat {\bm X}_{k + 1\left| t \right.}} = {\hat {\bm X}_{k + 1\left| {k + 1} \right.}^{\rm KF}} + {\bm C_{k + 1}}\bar {\bm F}_{k + 1}^T{\bm \Lambda^* _{k + 1}}.
    \label{eq:desired relation for k+1}
\end{equation}
It reflects the relation between the solution to the auxiliary MHE ${\hat {\bm X}_{k + 1\left| t \right.}}$ and the solution to the Kalman filter  ${\hat {\bm X}_{k + 1\left| {k + 1} \right.}^{\rm KF}}$. The latter's initial value $\hat{\bm X}_{t-N|t-N}^{\rm KF}$ can be derived from (\ref{eq:auxiliary kkt boundary}) by removing the term related to $\boldsymbol\Lambda_{t-N}^{*}$. Specifically, plugging $\hat{\bm W}_{t-N|t-N}=\left ( \bm L_{t-N}^{ww} \right )^{-1}\left ( -{\bm L}_{t-N}^{wx}\hat{\bm X}_{t-N|t-N}^{\rm KF}-{\bm L}_{t-N}^{w\theta } \right )$ from (\ref{eq:auxiliary kkt W}) into (\ref{eq:auxiliary kkt boundary}), we obtain $\hat{\bm X}_{t-N|t-N}^{\rm KF}=\hat{\bm X}_{t-N}+{\bm P}_{t-N}{\bm S}_{t-N}\hat{\bm X}_{t-N|t-N}^{\rm KF}+{\bm P}_{t-N}{\bm T}_{t-N}$. Using the same technique as in (\ref{eq:equivalent state prediction general case}), we can achieve the initial condition defined in (\ref{eq: initial condition for x}). This completes the proof.

\subsection{Comparison on NeuroBEM Test Dataset}\label{appendix:comparison}

\begin{table*}[!t]
\fontsize{7.5}{7.5}\selectfont
\caption{Estimation errors (RMSEs) comparisons on the NeuroBEM test dataset\label{table:testset rmse}}
\centering
\begin{threeparttable}[t]
\begin{tabular}{ c|c| c c c c c c| c c| c c } 
\toprule[1pt]
 \multirow{2} {*}{Trajectory} & \multirow{2} {*}{Method} & $F_x$ & $F_y$ & $F_z$ & $\tau_x$ & $\tau_y$ & $\tau_z$ & $F_{xy}$ & $\tau_{xy}$ & $F$ & $\tau$\\
 & & $\left [ \rm N \right ]$ & $\left [ \rm N \right ]$ & $\left [ \rm N \right ]$ & $\left [ \rm Nm \right ]$ & $\left [ \rm Nm \right ]$ & $\left [ \rm Nm \right ]$ & $\left [ \rm N \right ]$ & $\left [ \rm Nm \right ]$ & $\left [ \rm N \right ]$ & $\left [ \rm Nm \right ]$\\
\midrule[0.5pt]
\multirow{2} {*}{3D Circle\_1} & NeuroBEM & $\bf 0.196$ & $\bf 0.211$ & $0.215$ & $0.005$ & $0.006$ & $0.003$ & $\bf 0.288$ & $0.008$ & $\bf 0.360$ & $0.009$\\
& NeuroMHE & $0.258$ & $0.269$ & $\bf 0.108$ & $ \bf 0.003$ & $\bf 0.002$ & $ 0.003$ & $0.373$ & $\bf 0.004$ & $0.388$ & $\bf 0.005$\\
\midrule[0.5pt]
\multirow{2} {*}{Linear oscillation} & NeuroBEM & $0.164$ & $0.185$ & $0.456$ & $ 0.013$ & $0.011$ & $0.006$ & $0.247$ & $0.017$ & $0.518$ & $0.018$\\
& NeuroMHE & $\bf 0.119$ & $\bf 0.105$ & $\bf 0.186$ & $ \bf 0.011$ & $\bf 0.007$ & $\bf 0.005$ & $\bf 0.159$ & $\bf 0.013$ & $\bf 0.244$ & $\bf 0.014$\\
\midrule[0.5pt]
\multirow{2} {*}{Figure-8\_1} & NeuroBEM & $0.065$ & $ 0.056$ & $0.235$ & $0.004$ & $0.003$ & $ 0.002$ & $0.085$ & $0.005$ &$0.250$ &$0.006$\\
& NeuroMHE & $\bf 0.039$ & $0.056$ & $\bf 0.039$ & $\bf 0.002$ & $\bf 0.001$ & $0.002$ & $\bf 0.069$ & $\bf 0.002$ & $\bf 0.079$ & $\bf 0.003$\\
\midrule[0.5pt]
\multirow{2} {*}{Race track\_1} & NeuroBEM & $0.169$ & $0.158$ & $0.463$ & $0.009$ & $0.009$ & $0.004$ & $0.231$ & $0.013$ & $0.517$ & $0.013$\\
& NeuroMHE & $\bf 0.141$ & $\bf 0.092$ & $\bf 0.115$ & $\bf 0.007$ & $\bf 0.004$ & $0.004$ & $\bf 0.168$ & $\bf 0.009$ & $\bf 0.204$ & $\bf 0.009$\\
\midrule[0.5pt]
\multirow{2} {*}{Race track\_2} & NeuroBEM & $0.262$ & $0.248$ & $0.552$ & $0.014$ & $0.012$ & $\bf 0.007$ & $0.360$ & $0.019$ & $0.659$ & $\bf 0.020$ \\
& NeuroMHE & $\bf 0.245$ & $\bf 0.175$ & $\bf 0.208$ & $ \bf 0.012$ & $\bf 0.008$ & $0.018$ & $\bf 0.301$ & $\bf 0.014$ & $\bf 0.366$ & $0.023$ \\
\midrule[0.5pt]
\multirow{2} {*}{3D Circle\_2} & NeuroBEM & $\bf 0.110$ & $\bf 0.129$ & $0.470$ & $0.006$ & $0.009$ &$0.004$ & $\bf 0.170$ &$0.011$ & $0.499$ & $0.011$\\
& NeuroMHE & $0.140$ & $0.135$ & $\bf 0.075$ & $\bf 0.003$ & $\bf 0.002$ & $0.004$ & $0.194$ & $\bf 0.004$ & $\bf 0.208$ & $\bf 0.006$\\
\midrule[0.5pt]
\multirow{2} {*}{Figure-8\_2} & NeuroBEM & $0.051$ & $\bf 0.036$ & $0.339$ & $0.002$ & $0.002$ & $0.002$ & $0.063$ & $0.003$ & $0.345$ & $0.003$\\
& NeuroMHE & $\bf 0.020$ & $0.058$ & $\bf 0.029$ & $0.002$ & $\bf 0.001$ &$0.002$ & $\bf 0.061$ & $\bf 0.002$ & $\bf 0.068$ & $0.003$\\
\midrule[0.5pt]
\multirow{2} {*} {Melon\_1} & NeuroBEM & $0.099$ & $0.108$ & $0.397$ & $0.004$ & $0.005$ & $0.003$ & $0.147$ & $0.007$ & $0.423$ & $0.007$\\
& NeuroMHE & $\bf 0.053$ & $\bf 0.059$ & $\bf 0.060$ & $\bf 0.003$ & $\bf 0.001$ & $\bf 0.002$ & $\bf 0.079$ & $\bf 0.003$ & $\bf 0.099$ & $\bf 0.004$\\
\midrule[0.5pt]
\multirow{2} {*} {Figure-8\_3} & NeuroBEM & $0.145$ & $ 0.168$ & $0.584$ & $0.010$ & $0.012$ & $0.006$ & $0.221$ & $0.015$ & $0.624$ & $0.017$\\
& NeuroMHE & $\bf 0.118$ & $\bf 0.133$ & $\bf 0.151$ & $0.010$ & $\bf 0.006$ & $\bf 0.005$ & $\bf 0.178$ & $\bf 0.012$ & $\bf 0.233$ & $\bf 0.013$\\
\midrule[0.5pt]
\multirow{2} {*} {Figure-8\_4} & NeuroBEM & $0.400$ & $0.313$ & $1.084$ & $0.020$ & $0.018$ & $0.009$ & $0.508$ & $0.027$ & $1.197$ & $0.028$ \\
& NeuroMHE & $\bf 0.169$ & $\bf 0.174$ & $\bf 0.237$ & $\bf 0.014$ & $\bf 0.010$ & $\bf 0.007$ & $\bf 0.242$ & $\bf 0.017$ & $\bf 0.339$ & $\bf 0.018$\\
\midrule[0.5pt]
\multirow{2} {*} {Melon\_2} & NeuroBEM & $\bf 0.244$ & $\bf 0.198$ & $0.921$ & $0.009$ & $0.012$ & $0.006$ & $\bf 0.314$ & $0.015$ & $0.974$ & $0.016$\\
& NeuroMHE & $0.254$ & $0.213$ & $\bf 0.094$ & $\bf 0.005$ & $\bf 0.003$ & $\bf 0.004$ & $0.331$ & $\bf 0.005$ & $\bf 0.344$ & $\bf 0.007$\\
\midrule[0.5pt]
\multirow{2} {*} {Random points} & NeuroBEM & $0.161$ & $0.183$ & $0.471$ & $\bf 0.008$ & $0.008$ & $0.005$ & $0.244$ & $0.012$ & $0.530$ & $0.013$\\
& NeuroMHE & $\bf 0.115$ & $\bf 0.114$ & $\bf 0.204$ & $0.010 $ & $\bf 0.006$ & $\bf 0.004$ & $\bf 0.162$ & $0.012$ & $\bf 0.260$ & $\bf 0.012$\\
\midrule[0.5pt]
\multirow{2} {*} {Ellipse} & NeuroBEM & $0.204$ & $0.315$ & $1.039$ & $0.012$ & $0.018$ & $0.008$ & $0.375$ & $0.022$ & $1.105$ & $0.023$\\
& NeuroMHE & $\bf 0.176$ & $\bf 0.165$ & $\bf 0.089$ & $\bf 0.005$ & $\bf 0.003$ & $\bf 0.005$ & $\bf 0.242$ & $\bf 0.006$ & $\bf 0.258$ & $\bf 0.008$ \\
\bottomrule[0.5pt]
\end{tabular}
\begin{tablenotes}[flushleft]
      \footnotesize
      \item { The RMSEs of the planar and the overall disturbances are computed using the vector error (i.e., $\left \| {\bm d}_{f} - {\hat {\bm d}}_{f} \right \|_{2}$ and $\left \| {\bm d}_{\tau } - {\hat {\bm d}}_{\tau } \right \|_{2}$). For example, the RMSE of the planar force $F_{xy} $ is defined as $\sqrt{{N_{d}^{-1}}{\sum_{i=1}^{N_{d}}\left ( \Delta F_{x_i}^{2}+ \Delta F_{y_i}^{2}\right )}}$ where $N_d$ is the number of data and $\Delta F_{* }$ denotes the force estimation error between the NeuroMHE estimate and the ground truth data along *-axis. The force is expressed in the body frame to facilitate the comparison with NeuroBEM (which is provided in the body frame in the dataset and labeled as the 'predicted force' in~\cite{bauersfeld2021neurobem}). Note that the residual force data provided in the NeuroBEM dataset (columns 36-38 in the file 'predictions.tar.xz') was computed using the initially reported mass of $0.752\ {\rm kg}$ instead of the later revised value of $0.772\ {\rm kg}$. As a result, we refrain from utilizing this data to compute NeuroBEM's RMSE. The rest of the dataset remains unaffected.}
\end{tablenotes}
\end{threeparttable}
\end{table*}

\subsection{Design of Geometric Flight Control}\label{appendix:geometric}

Based on the quadrotor dynamics (\ref{eq:quadrotor model}), the nominal control force and torque of the geometric controller are given by:
\begin{subequations}
\begin{align}
    \bar{\bm F}_{d} &=-{\bm K}_{p}{\bm e}_{p}-{\bm K}_{v}{\bm e}_{v}+mg{\bm e}_3+m\ddot{\bm p}_{d}
    \label{eq:desired nominal control force},\\
    \begin{split}
    \bar{\boldsymbol\tau} _{m}&=-{\bm K}_{R}{\bm e}_{R}-{\bm K}_{\omega }{\bm e}_{\omega }+{\boldsymbol \omega}^{\times }{\bm J}{\boldsymbol \omega}\\
    &\quad -{\boldsymbol J}\left ({\boldsymbol \omega}^{\times }{\bm R}^{T}{\bm R}_{d}{\boldsymbol \omega}_{d}-{\bm R}^{T}{\bm R}_{d}{\dot {\boldsymbol \omega}}_{d} \right ),
    \end{split}
    \label{eq:nominal control torque}
\end{align}
\label{eq:active control appendix}%
\end{subequations}
where ${\bm K}_{p}$, ${\bm K}_{v}$, ${\bm K}_{R}$, ${\bm K}_{\omega } \in \mathbb{R}^{3\times3}$ are positive-definite gain matrices. These gains are manually tuned before training to achieve the best tracking performance in an ideal scenario where the external disturbances are fully known. The tracking errors of the above controller are defined by:
\begin{equation}
    \begin{aligned}
       {\bm e}_{p} &={\bm p}-{\bm p}_{d},& {\bm e}_{v} &={\bm v}-\dot{\bm p}_{d},\\
       {\bm e}_{R} &=\frac{1}{2}\left ( {\bm R}_{d}^{T}{\bm R}-{\bm R}^{T}{\bm R}_{d} \right )^{\vee },& {\bm e}_{\omega }&={\bm \omega} -{\bm R}^{T}{\bm R}_{d}{\bm \omega} _{d},
    \end{aligned}
\nonumber
\end{equation}
where $^{\vee }$ is the \textit{vee} operator: $\mathfrak{so}\left ( 3 \right )\rightarrow  \mathbb{R}^{3}$, and $\boldsymbol{\omega}_d$ is the desired angular rate defined using the method presented in Appendix-F of \cite{lee2010control}. The desired rotation matrix ${\bm R}_d$ is defined by ${\bm R}_{d}=\left [ {\bm b}_{1d},{\bm b}_{2d},{\bm b}_{3d} \right ]$ where ${{\bm b}_{2d}} = {{\left( {{{\bm b}_{3d}} \times {{\bm b}_{{\rm{int}}}}} \right)} \mathord{\left/
 {\vphantom {{\left( {{{\bm b}_{3d}} \times {{\bm b}_{{\rm{int}}}}} \right)} {\left\| {{{\bm b}_{3d}} \times {{\bm b}_{{\rm{int}}}}} \right\|}}} \right.
 \kern-\nulldelimiterspace} {\left\| {{{\bm b}_{3d}} \times {{\bm b}_{{\rm{int}}}}} \right\|}}$, ${\bm b}_{\rm int}=\left [ 1;0;0 \right ]$ for $\psi _{d}=0$, ${\bm b}_{1d}={\bm b}_{2d}\times {\bm b}_{3d}$, and 
 \begin{equation}
     {\bm b}_{3d}={\left\{\begin{matrix}
{\bm F}_{d}/ \left \|{\bm F}_{d}  \right \|& {\rm for}\ {\bm u}_{\rm r}\\ 
\bar{\bm F}_{d}/\left \| \bar{\bm F}_{d} \right \| & {\rm for}\ {\bm u}_{\rm b}
\end{matrix}\right.}.
\label{eq: desired z axis}
 \end{equation}

 \subsection{Design of $\mathcal{L}_1$ Adaptive Flight Control}\label{appendix:l1-ac}
 
In the design of the $\mathcal{L}_1$-AC, the disturbance force and torque are partitioned into matched and unmatched components. Based on this partition, the quadrotor model (\ref{eq:quadrotor model}) can be re-written as:
\begin{equation}
    \dot{\bm z}={\bm f}_v+{\bm B}\left ( {\bm u}+{\bm d}_{\rm m} \right )+{\bm B}^{\perp }{\bm d}_{\rm um},
    \label{eq:dynamics partition}
\end{equation}
where ${\bm z}=\left [ {\bm v};{\bm \omega}  \right ]$ denotes the partial quadrotor state as did in~\cite{wu20221}, ${\bm f}_{v}=\begin{bmatrix}
-g{\bm e}_3\\ 
-{\bm J}^{-1}{\bm \omega} ^{\times }{\bm J}{\bm \omega} 
\end{bmatrix}$, ${\bm B}=\begin{bmatrix}
m^{-1}{\bm R}{\bm e}_{3} & {\bm 0}_{3\times 3}\\ 
{\bm 0}_{3\times 1} & {\bm J}^{-1}
\end{bmatrix}$, ${\bm B}^{\perp }=\begin{bmatrix}
m^{-1}{\bm R}{\bm e}_{1} & m^{-1}{\bm R}{\bm e}_{2}\\ 
{\bm 0}_{3\times 1} & {\bm 0}_{3\times 1}
\end{bmatrix}$, ${\bm e}_{1}=\left [ 1;0;0 \right ]$, and ${\bm e}_{2}=\left [ 0;1;0 \right ]$. The matched disturbance ${\bm d}_{\rm m}$ is composed of the projection of the disturbance force onto the body-$z$ axis and the disturbance torque $\bm d_{\tau}$ (i.e., ${\bm d}_{\rm m}=\left [ {\bm d}_{f}\cdot \left ({\bm R}{\bm e}_{3}  \right );{\bm d}_{\tau } \right ]$), whereas the unmatched disturbance $\bm d_{\rm um}$ is the projection of the disturbance force onto the body-$xy$ plane (i.e., ${\bm d}_{\rm um}=\left [ {\bm d}_{f}\cdot \left ( {\bm R}{\bm e}_{1} \right );{\bm d}_{f}\cdot \left ( {\bm R}{\bm e}_{2} \right ) \right ]$). The design of the $\mathcal{L}_1$-AC follows the methodology presented in~\cite{wu20221}. For clarity, we outline two key components of the $\mathcal{L}_1$-AC below. The first component is the state predictor: 
\begin{equation}
    \dot{\hat{\bm z}}={\bm f}_{v}+{\bm B}\left ({\bm u}_{\rm b}+{\bm u}_{\rm ad} +\hat{\bm d}_{\rm m} \right )+{\bm B}^{\perp }\hat{\bm d}_{\rm um}+{\bm A}_{s}\tilde{\bm z}
    \label{eq:state predictor}
\end{equation}
where ${\bm u}_{\rm ad}$ is the $\mathcal{L}_1$ adaptive control law, $\hat{\bm d}_{\rm m}$ and $\hat{\bm d}_{\rm um}$ are the disturbance estimates in the body frame, $\tilde{\bm z}=\hat{\bm z}-{\bm z}$ is the prediction error, and ${\bm A}_s\in \mathbb{R}^{6 \times 6}$ is a diagonal Hurwitz matrix. The second component is the piecewise-constant $\mathcal{L}_1$ adaptation law that updates the estimate $\hat{\bm d}=\left [ \hat{\bm d}_{\rm m};\hat{\bm d}_{\rm um} \right ]$ by:
\begin{equation}
    \hat{\bm d}\left ( t \right )=\hat{\bm d}\left ( iT_{s} \right )=-\bar{\bm B}\left ( iT_{s} \right )^{-1}{\boldsymbol \Phi} ^{-1}{\boldsymbol \mu} \left ( iT_{s} \right )
    \label{eq:adaptation law}
\end{equation}
where $\bar{\bm B}\left ( iT_{s} \right )=\left [ {\bm B}\left ( {\bm R}\left ( iT_{s} \right ) \right ),{\bm B}^{\perp }\left ( {\bm R}\left ( iT_{s} \right ) \right ) \right ]$, $\boldsymbol\Phi={\bm A}_{s}^{-1}\left ( {\rm exp}\left ( {\bm A}_{s}T_{s} \right )-{\bm I} \right ) $, $\boldsymbol\mu\left ( iT_{s} \right )= {\rm exp}\left ( {\bm A}_{s}T_{s} \right )\tilde{\bm z}\left ( iT_{s} \right )$ for $i\in \mathbb{N}$, and $T_s$ is the time step. The $\mathcal{L}_1$ adaptive control law $\bm u_{\rm ad}$ is designed to compensate only for $\bm d_{\rm m}$ within the bandwidth of a low-pass filter (LPF). In the Laplacian domain, the control law can be expressed as ${\bm u}_{\rm ad}\left ( s \right )=-C\left ( s \right )\hat{\bm d}_{\rm m}\left ( s \right )$ where $C\left ( s \right )$ is the transfer function of the LPF and the bandwidth $\omega_c$ must satisfy the stability conditions~\cite{lakshmanan2020safe}.

\section*{Acknowledgement} \label{section: acknowledgement}

We thank Leonard Bauersfeld for the help in using the flight dataset of NeuroBEM.

\bibliographystyle{IEEEtran}
\bibliography{reference}

\end{document}